\documentclass[10pt,twocolumn,letterpaper]{article}

\PassOptionsToPackage{table}{xcolor}
\usepackage[pagenumbers]{cvpr}

\usepackage{tabularx}
\usepackage{array}
\usepackage{multirow}
\usepackage{pifont}
\usepackage{placeins}

\definecolor{cvprblue}{rgb}{0.21,0.49,0.74}
\usepackage[pagebackref,breaklinks,colorlinks,allcolors=cvprblue]{hyperref}

\usepackage[nointegrals]{wasysym}

\newcommand{\rsvdata}{RSVideo-10K}
\newcommand{\bench}{RSVideo-Bench}
\newcommand{\method}{RSVideo}

\newcommand{\yesmark}{\textcolor{green!50!black}{\ding{51}}}
\newcommand{\nomark}{\textcolor{red!75!black}{\ding{55}}}
\newcolumntype{P}[1]{>{\raggedright\arraybackslash}p{#1}}
\newcolumntype{C}[1]{>{\centering\arraybackslash}p{#1}}
\newcolumntype{Y}{>{\raggedright\arraybackslash}X}

\definecolor{wdcolor}{RGB}{128,0,255}

\definecolor{wdqcolor}{RGB}{255,0,0}

\definecolor{BurntOrange}{RGB}{191,87,0}

\renewcommand{\arraystretch}{1}

\def\confName{CVPR}
\def\confYear{2026}

\title{RSVideo: Are Your Vision-Language Models Ready for Remote Sensing Videos?}

\author{
Hongjie Zhou\textsuperscript{1,2,*} \quad
Shiqin Wang\textsuperscript{1,*} \quad
Haoyang Chen\textsuperscript{1,2,*} \quad
Haonan Guo\textsuperscript{1,2}\\
Di Wang\textsuperscript{1,2,\textdagger} \quad
Juhua Liu\textsuperscript{1} \quad
Fu Lin\textsuperscript{1} \quad
Yong Luo\textsuperscript{1,\textdagger}\\[3pt]
\textsuperscript{1}Wuhan University \quad
\textsuperscript{2}Zhongguancun Academy\\
Project: \url{https://HongjieZhou0329.github.io/RSVideo}
}

\begin{document}
\maketitle

{
\renewcommand{\thefootnote}{}
\footnotetext{\textsuperscript{*}Equal contribution}
\footnotetext{\textsuperscript{\textdagger}Corresponding authors (\{d\_wang,luoyong\}@whu.edu.cn)}
}

\begin{abstract}
Remote-sensing videos enable real-time observation of changes in target attributes, short-term activities, and scene evolution. They record motion, actions, interactions, and scene changes that cannot be captured by isolated images. Existing models primarily target single images or discrete temporal observations spanning a long time range. However, a unified evaluation setting for assessing vision-language models on continuous remote-sensing video understanding remains lacking. We introduce \textbf{\rsvdata{}}, a remote-sensing video dataset comprising 10,773 instances, 1.47 million frames, and 17.02 hours of footage, containing both unmanned aerial vehicles and satellite platforms. Its fixed evaluation benchmark, \textbf{\bench{}}, contains 2,731 test instances and evaluates two complementary aspects of remote-sensing video understanding: L1 Perception and L2 Reasoning, spanning seven capability groups and 17 tasks. Evaluations show that current vision-language models still struggle to recover small local evidence, track short-lived states, and use scene-constrained spatial relations. Based on this analysis, we further propose \textbf{\method{}}, a reinforcement learning framework for small-target spatiotemporal focusing that selects question-relevant regions across frames and suppresses redundant background tokens. \method{} achieves a maximum absolute improvement of 9.01\% with InternVL3.5-14B and attains the highest accuracy of 40.63\% with Qwen3.6-27B across 26 open-source vision-language backbones. Codes will be available at https://github.com/HongjieZhou0329/RSVideo.

\end{abstract}


\section{Introduction}

Remote-sensing videos are becoming an important source for traffic monitoring, public-safety surveillance, moving-target analysis, disaster response, and environmental observation \citep{bozcan2020auair,barekatain2017okutama,robicquet2016sdd,mou2020era,zhao2022satsot, rvsa}. Unlike a single remote-sensing image, a video continuously observes the scene and captures transient changes. This enables real-time observation of target attribute changes, short-term activities, and scene evolution. Recent vision-language models improve the automation of video analysis and achieved strong performance on natural-video tasks \citep{li2024llavaonevision,zhang2025videollama3,bai2025qwen25vl}. 

In remote-sensing videos, targets may occupy only a few pixels, while objects with similar appearances are difficult to distinguish. Meanwhile, critical actions and state changes can be subtle and short-lived, and complex backgrounds introduce substantial redundancy. These challenges lead to a fundamental question: \emph{How well do current vision-language models understand remote-sensing videos?}

\begin{figure}[t]
	\centering
	\includegraphics[width=1\linewidth]{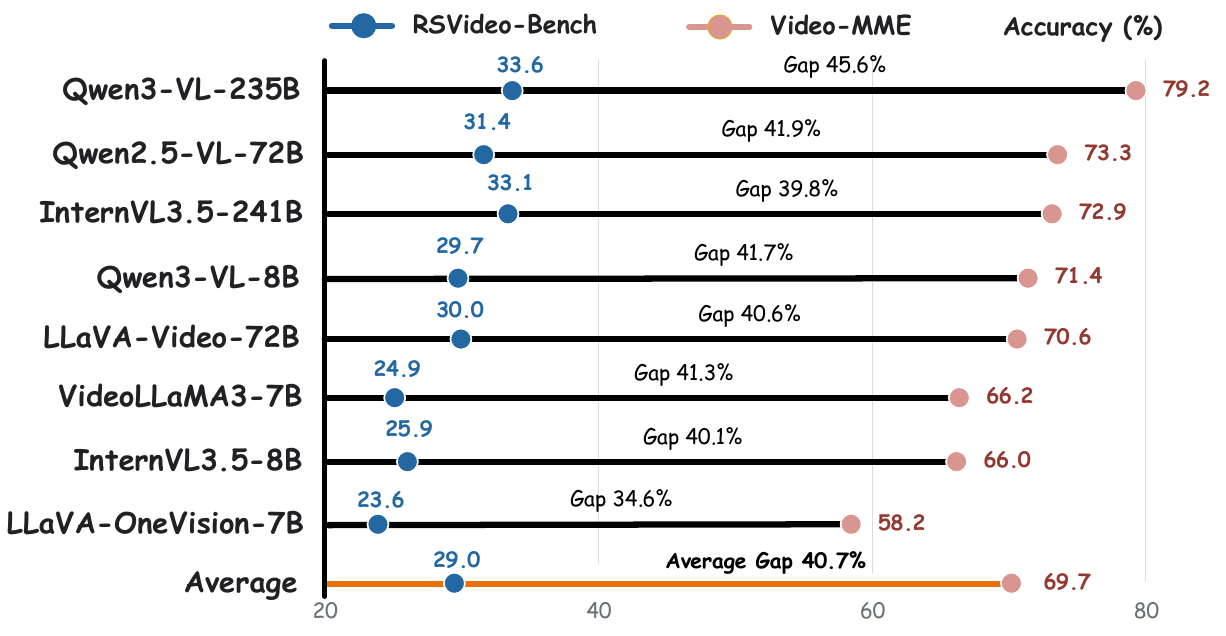}\\
	\caption{Accuracy of representative vision-language models on RSVideo-Bench (ours) and Video-MME~\citep{fu2025videomme}. Existing multimodal large language models achieve an average accuracy of 69.7\% on natural videos, but only 29.0\% on remote-sensing videos, resulting in an average performance gap of 40.7\%.}
	\label{fig:benchmark-gap}
\end{figure}
Answering this question using existing remote-sensing benchmarks is difficult.
Static remote-sensing vision-language datasets support scene understanding, visual question answering, captioning and visual grounding \citep{lobry2020rsvqa,wang2024earthvqa,wang2024skyscript,li2024vrsbench,kuckreja2024geochat,ni2026uhrmicro}. Their basic unit is a single image or region. They cannot capture continuous motion or short-lived states. Multi-temporal remote-sensing datasets \citep{toker2022dynamicearthnet,yuan2024fusu,xuan2025dynamicvl,li2024unirs} support long-term change analysis. However, their observations are usually separated by intervals of days, months, or years. This distinguishes them from videos, which capture movements and state changes in seconds. Existing UAV datasets mainly target low-altitude perception, grounding, or task-specific reasoning
\citep{bozcan2020auair,barekatain2017okutama,mou2020era,bashmal2023capera,sun2025refdrone,zhan2025uavsvg,zhan2026uavbench,sun2026uavreason}.
They are insufficient for evaluating continuous remote-sensing video understanding across UAV and satellite platforms, where target scales and visual evidence evolve over time.

To support both controlled evaluation and model development, we introduce \textbf{\rsvdata{}}. To our knowledge, this is the first large-scale remote-sensing video dataset designed for video question answering at the 10K-instance scale. It comprises \textbf{RSVideo-Instruct}, which provides 8,042 training and validation instances, and \textbf{\bench{}}, a fixed test set containing 2,731 instances. \rsvdata{} contains continuous videos from eight public sources spanning UAV and satellite observations, with a total duration of approximately 17.02 hours and 1,473,150 decoded frames. All instances are organized under a unified taxonomy covering two capability dimensions and 17 tasks. 

We use \bench{} to evaluate representative open-weight and proprietary vision-language models. It contains 653 L1 Perception instances and 2,078 L2 Reasoning instances.
For comparison, we perform evaluations on the natural video dataset Video-MME~\citep{fu2025videomme}. All models are evaluated using the same video input and five-choice protocol.
As shown in Figure~\ref{fig:benchmark-gap}, many vision-language models achieve strong performance on Video-MME. However, this performance gain does not directly transfer to remote-sensing videos.
Furthermore, we found that their errors concentrate on small targets, visually similar objects, weak or short-lived state changes, and spatial relations defined by roads, buildings, region boundaries, or nearby objects. This pattern points to a specific research question: \emph{How can vision-language models recover question-relevant spatiotemporal evidence and target states from continuous remote-sensing videos?}

RSVideo-Instruct makes these challenges trainable by providing answer supervision together with temporal and spatial evidence annotations, enabling evidence-aware model initialization. Based on this supervision, we propose \textbf{\method{}}, a two-stage training framework for spatiotemporal evidence focusing. The first stage performs evidence-aware initialization, teaching the model to associate predictions with relevant spatiotemporal evidence. The second stage further optimizes this evidence focusing capability through reinforcement learning, enabling adaptive spatiotemporal evidence selection and redundant context compression under a fixed visual token budget. To guide these behaviors, we design evidence-aware rewards that jointly consider answer correctness, spatiotemporal evidence alignment, and background compression. \method{} achieves consistent performance improvements across multiple open-weight vision-language backbones on \bench{}. Extensive experiments verify that \method{} improves remote-sensing video understanding performance across 26 vision-language backbones ranging from 1B to 241B parameters.

The contributions of this work are summarized as follows:
\begin{itemize}

\item We introduce \textbf{\rsvdata{}}, the first unified dataset for perception and reasoning on continuous UAV and satellite videos. It contains 10,773 instances, including 8,042 training and validation instances from \textbf{RSVideo-Instruct} and 2,731 test instances in \textbf{\bench{}}. All instances follow a unified taxonomy spanning two capability dimensions, seven capability groups, and 17 tasks, with expert annotation and review to ensure quality.

\item We propose \textbf{\method{}}, a two-stage training framework for spatiotemporal evidence focusing. It leverages temporal and spatial evidence annotations in RSVideo-Instruct for evidence-aware initialization and further optimizes evidence selection through reinforcement learning. Under a fixed visual token budget, \method{} adaptively preserves informative evidence tokens and compresses redundant background context.

\item Extensive experiments demonstrate that existing vision-language models still struggle with evidence-grounded remote-sensing video understanding, particularly for small targets, short-lived states, and scene-constrained spatial relations. \method{} consistently improves performance across 26 evaluated vision-language backbones, achieving a maximum absolute gain of 9.01\% with InternVL3.5-14B.

\end{itemize}

\section{Related Work}

\begin{table*}[t]
\centering
\scriptsize
\setlength{\tabcolsep}{1.7pt}
\renewcommand{\arraystretch}{1.06}
\resizebox{\textwidth}{!}{%
\begin{tabular}{P{5cm} C{1.05cm} C{0.82cm} C{0.72cm} C{1.33cm} C{1.27cm} C{0.72cm} C{0.70cm} C{1.00cm} C{0.70cm} C{0.58cm} C{0.58cm} C{0.70cm} C{1.02cm}}
\toprule
\multirow{2}{*}{Dataset}
& \multicolumn{3}{c}{Video Scale}
& \multicolumn{3}{c}{Task Taxonomy}
& \multicolumn{2}{c}{\shortstack[c]{Evidence}}
& \multicolumn{3}{c}{Splits}
& \multicolumn{2}{c}{Video Coverage} \\
\cmidrule(lr){2-4}
\cmidrule(lr){5-7}
\cmidrule(lr){8-9}
\cmidrule(lr){10-12}
\cmidrule(lr){13-14}
& Queries & Clips & \shortstack[c]{Hours}
& Perception & Reasoning & Tasks
& Time & Position
& Train & Val. & Test
& UAV & Satellite \\
\midrule

\rowcolor{black!6}
\multicolumn{14}{l}{\textit{Static Remote-Sensing Vision-Language Datasets}} \\
RSVQA \citep{lobry2020rsvqa}
& 0 & 0 & 0
& \yesmark & \nomark & 5
& \nomark & \nomark
& \yesmark & \yesmark & \yesmark
& \nomark & \nomark \\
EarthVQA \citep{wang2024earthvqa}
& 0 & 0 & 0
& \yesmark & \yesmark & 6
& \nomark & \nomark
& \yesmark & \yesmark & \yesmark
& \nomark & \nomark \\
VRSBench \citep{li2024vrsbench}
& 0 & 0 & 0
& \yesmark & \yesmark & 3
& \nomark & \yesmark
& \yesmark & \nomark & \yesmark
& \nomark & \nomark \\
XLRS-Bench \citep{wang2025xlrsbench}
& 0 & 0 & 0
& \yesmark & \yesmark & 16
& \nomark & \yesmark
& \nomark & \nomark & \yesmark
& \nomark & \nomark \\

\midrule
\rowcolor{black!6}
\multicolumn{14}{l}{\textit{Multi-Temporal Remote-Sensing Vision-Language Datasets}} \\
TEOChat / TEOChatlas \citep{irvin2025teochat}
& 0 & 0 & 0
& \yesmark & \yesmark & 0
& \yesmark & \nomark
& \yesmark & \nomark & \nomark
& \nomark & \nomark \\
DynamicVL / DVL-Suite \citep{xuan2025dynamicvl}
& 0 & 0 & 0
& \yesmark & \yesmark & 6
& \yesmark & \yesmark
& \yesmark & \nomark & \yesmark
& \nomark & \nomark \\
VLRS-Bench \citep{luo2026vlrsbench}
& 0 & 0 & 0
& \yesmark & \yesmark & 14
& \yesmark & \nomark
& \nomark & \nomark & \yesmark
& \nomark & \nomark \\

\midrule
\rowcolor{black!6}
\multicolumn{14}{l}{\textit{General Vision-Language Datasets}} \\
VSI-Bench \citep{yang2025thinking}
& 5,130 & 288 & 7.73
& \yesmark & \yesmark & 8
& \nomark & \yesmark
& \nomark & \nomark & \yesmark
& \nomark & \nomark \\
TempCompass \citep{liu2024tempcompass}
& 7,540 & 410 & 1.30
& \yesmark & \nomark & 4
& \nomark & \nomark
& \nomark & \nomark & \yesmark
& \nomark & \nomark \\
MVBench \citep{li2024mvbench}
& 4,000 & 3,641 & 13.22
& \yesmark & \yesmark & 20
& \yesmark & \nomark
& \nomark & \nomark & \yesmark
& \nomark & \nomark \\
Video-MME \citep{fu2025videomme}
& 2,700 & 900 & 255.32
& \yesmark & \yesmark & 12
& \nomark & \nomark
& \nomark & \nomark & \yesmark
& \nomark & \nomark \\

\midrule
\rowcolor{black!6}
\multicolumn{14}{l}{\textit{UAV-Centric Multimodal Datasets}} \\
UAVBench / UAVIT-1M \citep{zhan2026uavbench}
& 0 & 0 & 0
& \yesmark & \yesmark & 10/11
& \nomark & \yesmark
& \yesmark & \nomark & \yesmark
& \nomark & \nomark \\
UAVReason \citep{sun2026uavreason}
& 0 & 0 & 0
& \yesmark & \yesmark & 22
& \yesmark & \yesmark
& \yesmark & \nomark & \yesmark
& \nomark & \nomark \\
UrbanVideo-Bench \citep{zhao2025urbanvideo}
& 5,355 & 1,393 & 48.71
& \yesmark & \yesmark & 16
& \nomark & \nomark
& \yesmark & \nomark & \yesmark
& \yesmark & \nomark \\
SIS-Bench \citep{zou2026sisbench}
& 4,856 & 1,646 & 14.06
& \yesmark & \yesmark & 13
& \nomark & \nomark
& \nomark & \nomark & \yesmark
& \yesmark & \nomark \\
\rsvdata{} (ours)
& 10,773 & 4,629 & 17.02
& \yesmark & \yesmark & 17
& \yesmark & \yesmark
& \yesmark & \yesmark & \yesmark
& \yesmark & \yesmark \\

\bottomrule
\end{tabular}%
}
\caption{
Comparison with representative vision-language datasets. Queries and Clips denote the number of continuous-video QA instances and unique video inputs, respectively. Hours indicate the total duration of deduplicated video segments used for question construction; each segment is counted once regardless of the number of associated queries. Task Taxonomy denotes explicit capability dimensions and task numbers. Evidence indicates whether temporal or spatial evidence annotations are provided. Splits follow the corresponding papers or official releases. Video Coverage indicates the source platforms of the videos. ``0'' in Video Scale denotes image-only datasets.
}
\label{tab:benchmark-comparison}
\end{table*}

\subsection{Remote-Sensing Vision-Language Data}

Remote-sensing vision-language datasets support question answering, captioning, and visual grounding over satellite and aerial images \citep{lobry2020rsvqa,wang2024earthvqa,wang2024skyscript,li2024vrsbench,kuckreja2024geochat}. XLRS-Bench and UHR-Micro study multi-scale understanding and local evidence extraction from ultra-high-resolution imagery \citep{wang2025xlrsbench,ni2026uhrmicro}. These datasets provide broad spatial coverage but operate mainly on static images. Multi-temporal datasets instead model changes between observations acquired at different times. DynamicEarthNet and FUSU support land-cover and semantic change analysis, while DynamicVL introduces language-based reasoning over long-term urban changes \citep{toker2022dynamicearthnet,yuan2024fusu,xuan2025dynamicvl}. However, these observations are typically separated by long intervals and cannot capture continuous motion and short-lived state changes. 

\subsection{General Video-Language Understanding} 

Vision-language datasets have been developed for action recognition, event understanding, temporal reasoning, and long-form video comprehension \citep{mangalam2023egoschema,li2024mvbench,fu2025videomme}. Recent methods improve video reasoning through hierarchical memory, adaptive frame selection, active perception, and evidence retrieval \citep{wang2025videotree,arnab2025temporal,yang2025vca,zou2026air,ma2026vap,gao2026videotir,yin2026videoarm}. These studies establish strong foundations for video-level reasoning, but are mainly developed for natural videos with relatively salient targets and temporal changes. In contrast, remote-sensing videos contain small targets, large repetitive backgrounds, subtle state changes, and spatial relations constrained by scene structures such as roads, buildings, and region boundaries. These characteristics require models to recover weak and sparse spatiotemporal evidence before performing reliable video reasoning.

\subsection{Aerial and Remote-Sensing Video Data}

Existing aerial video datasets have primarily focused on task-specific problems, such as object detection, tracking, action recognition, and event analysis \citep{zhao2022satsot,chen2024ootb,yin2021viso}. Recent efforts extend aerial video understanding toward vision-language tasks. For example, UAVBench and UAVIT-1M explore UAV vision-language understanding at different granularities, while UAVReason investigates geometric and topological reasoning in simulated nadir-view scenarios \citep{zhan2026uavbench,sun2026uavreason}. Other benchmarks, such as UrbanVideo-Bench and SIS-Bench, evaluate broader UAV video understanding beyond perception tasks \citep{zhao2025urbanvideo,zou2026sisbench}. However, existing aerial video resources remain fragmented across platforms and tasks, lacking comprehensive evaluation of continuous remote-sensing video understanding. In contrast, \rsvdata{} provides a unified framework that integrates continuous UAV and satellite videos with fine-grained perception and reasoning tasks under a shared taxonomy.

\section{Remote Sensing Video Understanding Dataset \rsvdata{}}

\begin{figure*}[t]
	\centering
	\includegraphics[width=\linewidth]{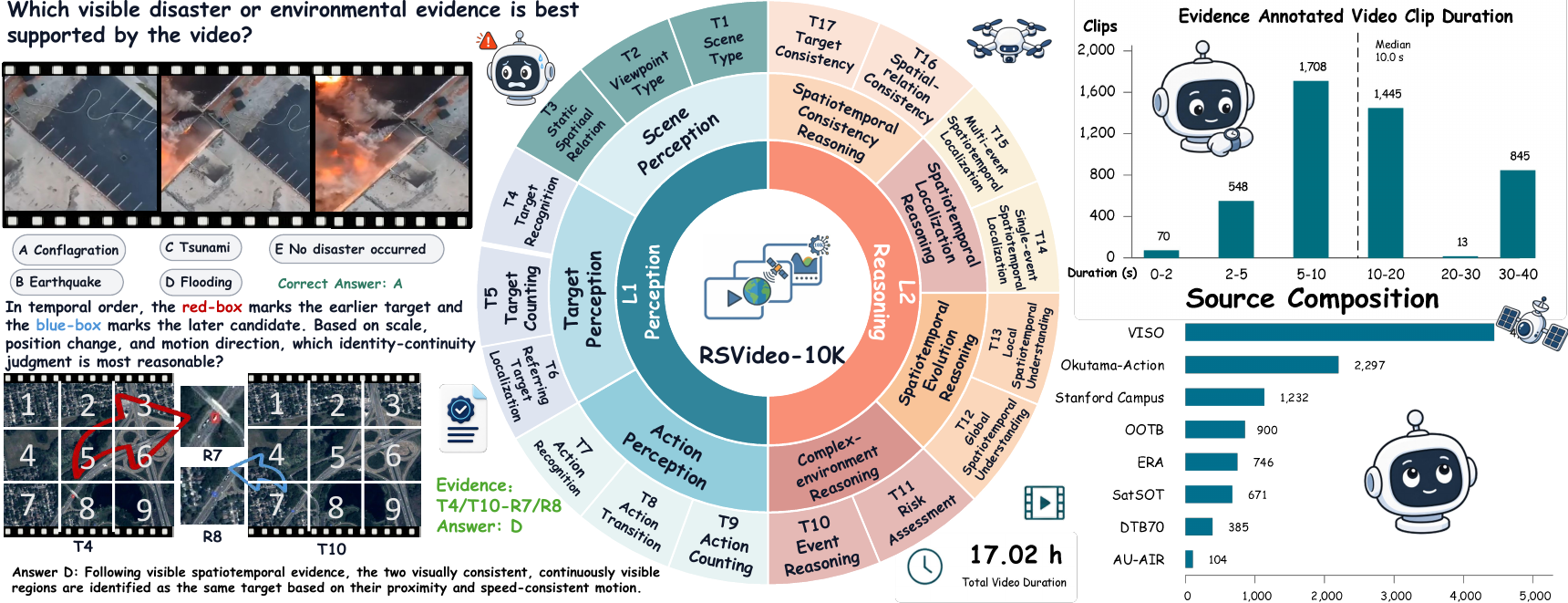}\\
	\caption{Overview of \rsvdata{}. The dataset contains 10,773 five-choice QA instances grounded in 4,629 audited evidence clips from eight public video sources. It covers 17.02 hours of continuous remote-sensing video with 1,473,150 decoded frames. \rsvdata{} comprises RSVideo-Instruct for training and validation and \bench{} for fixed evaluation. The figure summarizes the source composition, clip duration of evidence annotated videos, and the L1/L2 capability distribution.}
	\label{fig:rsv10k-overview}
\end{figure*}

\rsvdata{} contains 10,773 QA instances derived from 4,629 manually audited video clips, covering 17.02 hours of continuous video and 1.47M decoded frames. Each instance contains a video, a question, and five answer choices, with all questions formulated as single-choice questions. Notably, 4,638 instances of \rsvdata{} were additionally annotated temporal and spatial evidence, indicating where and when the question-relevant target or event is located. \rsvdata{} consists of 8,042 \textbf{RSVideo-Instruct} instances and 2,731 \textbf{\bench{}} instances under a shared taxonomy and annotation protocol. RSVideo-Instruct provides training and validation data for model development. \bench{} is a locked test set for evaluating remote-sensing video perception and reasoning. Table~\ref{tab:benchmark-comparison} compares \rsvdata{} with representative vision-language and remote-sensing datasets.

\subsection{Task Taxonomy}

We organize \rsvdata{} with a four-level taxonomy: two capability dimensions, seven capability groups, 17 tasks, and 26 fine-grained leaves. The leaves specify distinct evidence patterns within each task. Their definitions are provided in the Appendix A.2. Figure~\ref{fig:rsv10k-overview} shows the specific details.

\textbf{L1 Perception} includes Scene Perception (SP), Target Perception (TP), and Action Perception (AP). SP covers scene type, viewpoint type, and static spatial relation. TP contains target recognition, target counting, and referring spatial localization. AP involves action recognition, action transition, and action counting.

\textbf{L2 Reasoning} contains Complex-environment Reasoning (CER), Spatiotemporal Evolution Reasoning (SER), Spatiotemporal Localization Reasoning (SLR), and Spatiotemporal Consistency Reasoning (SCR). CER covers event reasoning and risk assessment. SER covers spatial relation understanding. SLR covers local-event, single-target, and multi-event spatiotemporal localization. SCR covers spatial relation consistency and target consistency. These seven capability groups also serve as group-level evaluation metrics.

\subsection{Dataset Construction}
\rsvdata{} is built from eight public sources spanning UAV and satellite videos. To construct this datset, we first decode source videos and identify continuous clips with sufficient visual evidence. Selected clips contain observable targets, actions, scene states, spatial relations, temporal changes, or cross-frame consistency cues. We exclude clips with severe blur, invalid viewpoints, weak temporal continuity, heavy occlusion, or insufficient evidence.

For each task, we design an automated pipeline guided by ground-truth annotations and large-model assistance. Candidate answers are constructed from visually similar categories, neighboring regions, related actions, or easily confused spatial and temporal relations. For cases where the available visual evidence is insufficient for a reliable judgment, we add an ``insufficient evidence'' answer option. This design avoids forced guessing and evaluates whether models can identify when the available evidence is insufficient for answering the question. The detailed pipeline is provided in Appendix B.1.

\subsection{Data Quality and Release}

We construct \rsvdata{} through a multi-stage quality control process, including clip screening, question generation, evidence verification, and multi-round review. Three senior domain experts conducted annotation, independent verification, and final adjudication to ensure the quality of videos, questions, answer choices, evidence annotations, and task taxonomy. Qwen-based models were used only for auxiliary consistency checking, while final decisions were made by human annotators. The complete construction procedure is described in Appendix B.1.

All source videos are collected from publicly available resources. We carefully review their licenses and release \rsvdata{} following a license-compliant policy. For redistributable sources, processed clips are released under their corresponding licenses. Source-level license information and release details are provided in Appendix A.1.

\begin{figure}[t]
	\centering
	\includegraphics[width=\linewidth]{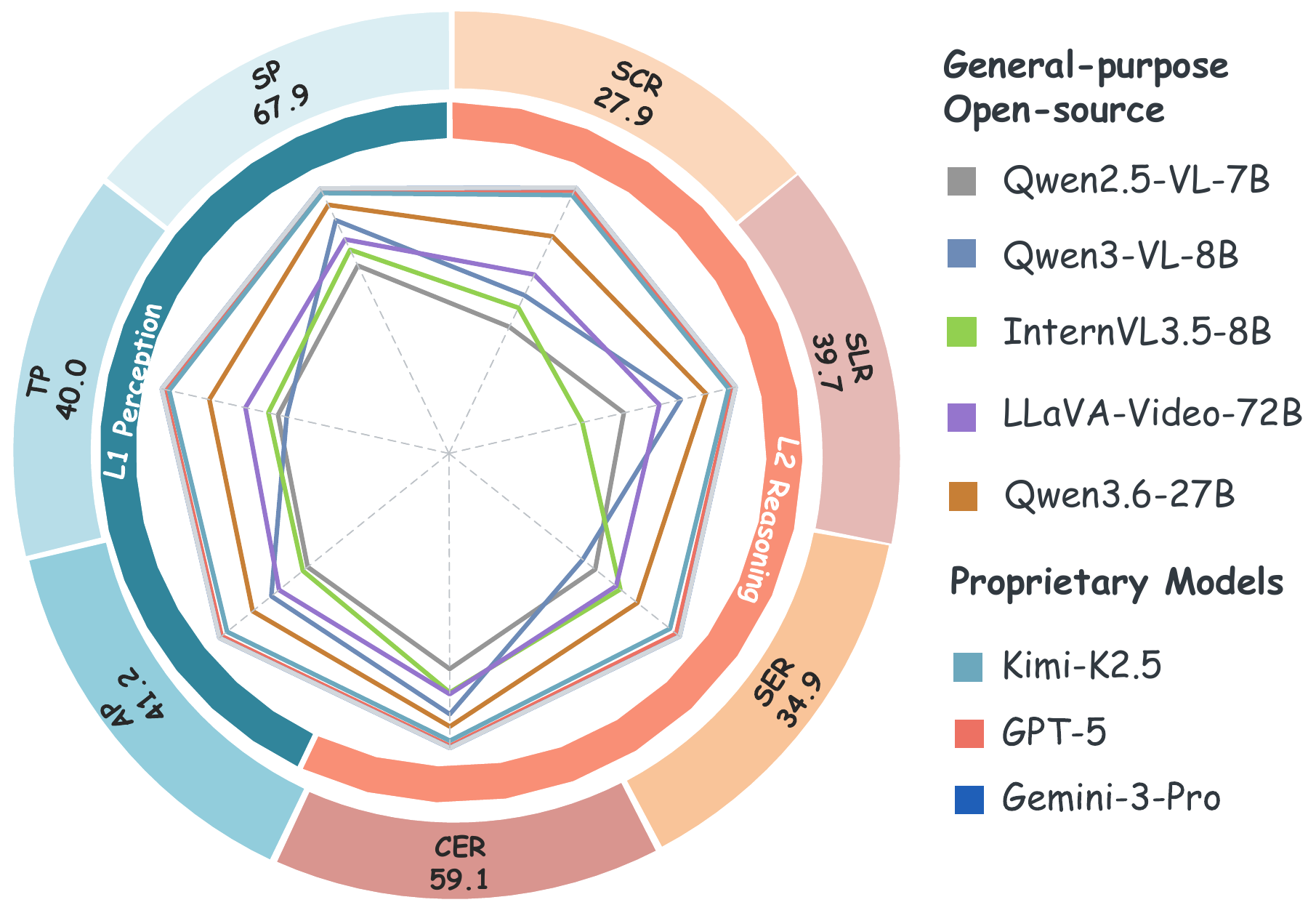}\\
	\caption{Performance of representative vision-language models on \bench{} across seven capability groups. The values on the outermost ring represent the results of the proprietary model Gemini-3-Pro on \bench{}.}
	\label{fig:pilot-radar}
\end{figure}

\begin{figure*}[t]
  \centering
   \includegraphics[width=\linewidth]{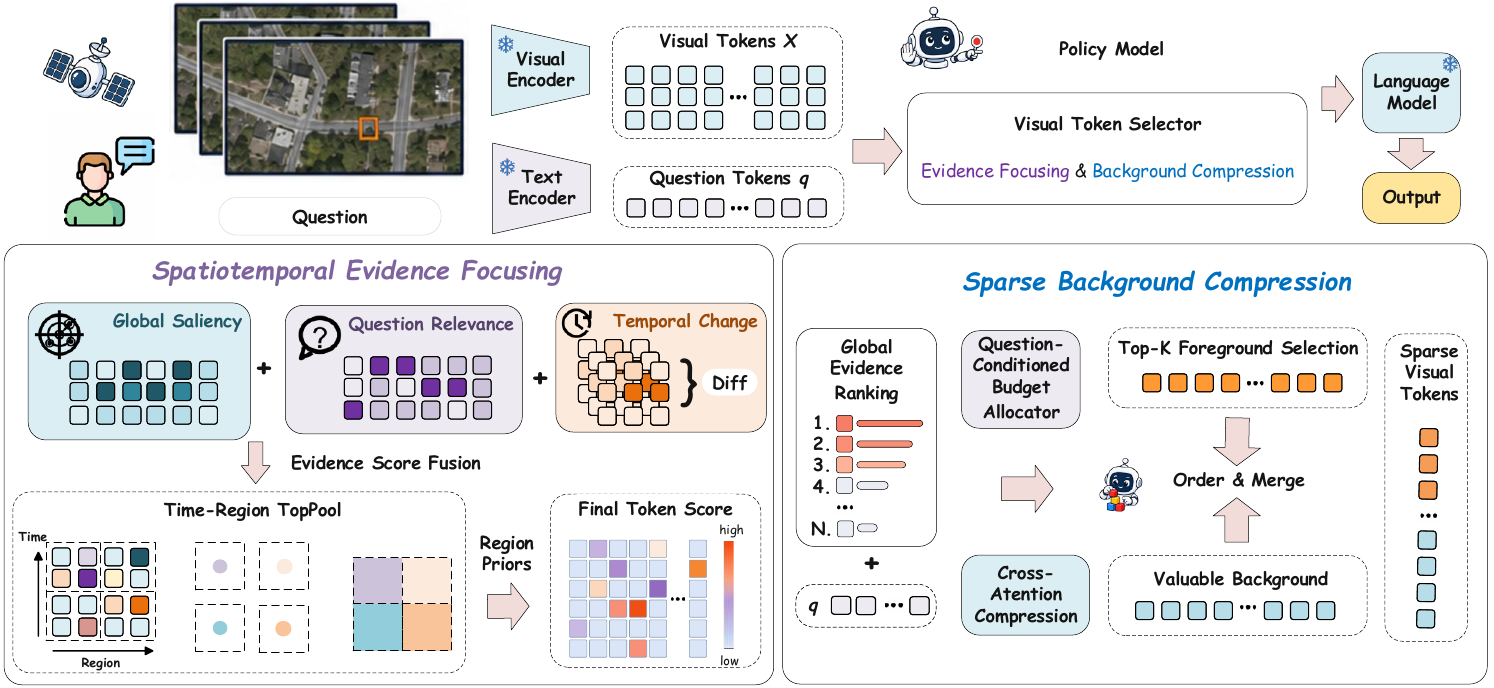}
   \caption{Overview of \method{}. The visual encoder and text encoder first generate visual tokens and question tokens. \method{} combines global saliency, question relevance, and temporal change for spatiotemporal evidence focusing, integrated with spatiotemporal region priors to obtain final evidence scores for visual tokens. Then \method{}  performs sparse background compression. Based on the evidence scores, \method{} selects high-value spatiotemporal tokens through global ranking and compresses low-value background tokens into question-related contextual slots. It finally orders and merges the selected foreground tokens and compressed background slots into a sparse visual sequence for the vision-language model to generate the answer.}
   \label{fig:framework}
\end{figure*}

\section{Findings and Method}
\label{sec:method}

We conduct extensive evaluations of existing open-source and proprietary models on \bench{}. Additional benchmark results and diagnostic analyses are reported in Appendix~\ref{app:additional_analyses}. In this section, we first summarize the empirical findings and then introduce \method{} in detail.

\subsection{Benchmark Findings}
Figure~\ref{fig:pilot-radar} presents the performance of representative open-source and proprietary vision-language models across seven \bench{} capability dimensions. The models perform relatively well on SP and CER, but drop clearly on AP, TP, SER, SLR, and SCR, leading to following findings:

\textbf{Sparse evidence is easily missed.}
Errors often concentrate on small targets, subtle actions, and short-lived states. The decisive evidence is frequently sparse in both space and time, occupying only a few visual tokens across a limited number of frames. Once these local cues are overlooked, simply increasing model capacity or extending the video context provides limited improvement.

\textbf{Scene context is necessary but highly redundant.}
Large remote-sensing scenes contain extensive background regions, such as fields, water bodies, and vegetation, most of which are irrelevant to a given question. However, informative structures, including roads, buildings, region boundaries, and nearby objects, often provide essential spatial cues for reasoning. Therefore, the model needs to preserve useful scene context while avoiding the cost of retaining the entire background at full resolution.

\textbf{Frame-level recognition is insufficient for spatiotemporal reasoning.}
Low performance on SER, SLR, and SCR indicates that recognizing individual visual cues is insufficient for coherent video understanding. These tasks require models to associate sparse evidence with the correct target, temporal stage, and spatial context across frames. Missing or misaligned evidence can therefore lead to incorrect reasoning outcomes.

\textbf{Correct answers do not guarantee grounded decisions.}
Achieving the correct answer does not necessarily indicate that the model identifies the underlying visual evidence. A model may select the correct option through spurious correlations without associating its prediction with the relevant frames and regions. Reliable video understanding therefore requires not only answer accuracy but also evidence-grounded decision making. These observations motivate the need for explicit supervision that connects model predictions with their supporting spatiotemporal evidence.

\subsection{\method{}}
\label{sec:MethodRSVideo}

Motivated by the above findings, \method{} treats evidence as a compact spatiotemporal support set that explains an answer, encoded by temporal positions and spatial cells, e.g., \texttt{Evidence: T08-R05/R06; Answer: C}. Training follows a two-stage design. We first perform evidence-aware supervised fine-tuning (E-SFT) on the full visual sequence to establish the evidence format and a reliable policy prior. We then enable sparse visual selection and optimize the token scorer, background allocator, and answer/evidence generation with reinforcement learning. As shown in Figure~\ref{fig:framework}, the RL-stage method contains three parts: spatiotemporal evidence focusing, sparse background compression, and evidence-aware reward optimization.

\paragraph{Spatiotemporal Evidence Focusing.}
Given a video \(V\) and a question \(Q\), we sample \(T\) frames, extract patch-level visual features, project them into the language-model embedding space, and flatten all frame tokens into a unified sequence. The question is encoded into a pooled semantic representation:
\begin{equation}
\begin{aligned}
\mathbf{H}
&=
[\mathbf{v}_1;\ldots;\mathbf{v}_L]
=
\operatorname{Flatten}
\left(
\operatorname{Proj}(f_{\mathrm{vis}}(V))
\right),\\
\mathbf{q}
&=
\operatorname{Pool}
\left(
f_{\mathrm{txt}}(Q)
\right),
\end{aligned}
\label{eq:visual-question-encoding}
\end{equation}
where \(\mathbf{H}\in\mathbb{R}^{L\times d}\), \(L=T\times N_p\), and \(N_p\) is the number of patch tokens per frame.

To rank visual evidence, each token \(\mathbf{v}_i\) receives three complementary signals: global saliency \(s_i^{\mathrm{sal}}\) from visual self-attention, question relevance \(s_i^{\mathrm{rel}}\) from cross-modal similarity, and temporal change \(s_i^{\mathrm{chg}}\) from local spatiotemporal inconsistency. After per-video normalization, we fuse them as
\begin{equation}
u_i
=
\alpha_{\mathrm{sal}}\bar{s}_i^{\mathrm{sal}}
+
\alpha_{\mathrm{rel}}\bar{s}_i^{\mathrm{rel}}
+
\alpha_{\mathrm{chg}}\bar{s}_i^{\mathrm{chg}}.
\label{eq:token-evidence-score}
\end{equation}
Since small remote-sensing targets may be weak at single-token level, we further aggregate responses within coarse time--region cells and assign the normalized cell prior back to each token:
\begin{equation}
\ell_i
=
u_i
+
\alpha_{\mathrm{cell}}
\bar{c}_{\tau_i,g_i},
\label{eq:final-token-score}
\end{equation}
where \((\tau_i,g_i)\) denotes the temporal index and coarse spatial region of token \(i\). The final score \(\ell_i\) therefore combines token-level evidence and local region-level support. The definitions of \(s_i^{\mathrm{sal}}\), \(s_i^{\mathrm{rel}}\), \(s_i^{\mathrm{chg}}\), and \(c_{\tau,g}\) are given in Appendix D.1.

\paragraph{Sparse Background Compression and Allocation.}
Directly dropping all low-score tokens may remove useful context such as roads or buildings. Therefore, \method{} keeps the strongest evidence tokens and compresses the remaining tokens into a small set of question-conditioned background representations.

Given \(\boldsymbol{\ell}=(\ell_1,\ldots,\ell_L)\), we keep a fixed visual budget \(B=\max(2,\operatorname{round}(\rho L))\), with \(\rho=0.40\) by default. A lightweight allocator conditioned on \(\mathbf{q}\) and the score distribution predicts the background ratio \(\gamma\). During training, \(\gamma\) is sampled from a predicted Beta distribution to explore different evidence/background trade-offs; during inference, its distribution mean is used. The number of compressed background tokens and retained evidence tokens is
\begin{equation}
M
=
\operatorname{clip}
\left(
\operatorname{round}(\gamma B),
1,
B-1
\right),
\qquad
K=B-M .
\label{eq:visual-budget}
\end{equation}
The \(K\) highest-scoring tokens form \(\mathbf{H}_{\mathrm{loc}}\), while the remaining tokens form \(\mathbf{H}_{\mathrm{rem}}\), both kept in their original spatiotemporal order.

To preserve compact scene context, we construct \(M\) question-conditioned background queries and let them attend to \(\mathbf{H}_{\mathrm{rem}}\):
\begin{equation}
\begin{aligned}
\mathbf{q}^{\mathrm{bg}}_m
&=
\mathbf{e}_m
+
\mathbf{W}_{\mathrm{bg}}\mathbf{q},
\quad
m=1,\ldots,M,\\
\mathbf{Z}_{\mathrm{bg}}
&=
\operatorname{Attn}
\left(
\mathbf{Q}_{\mathrm{bg}},
\mathbf{H}_{\mathrm{rem}},
\mathbf{H}_{\mathrm{rem}}
\right),\\
\widehat{\mathbf{H}}
&=
\operatorname{Concat}
\left(
\mathbf{H}_{\mathrm{loc}},
\mathbf{Z}_{\mathrm{bg}}
\right),
\end{aligned}
\label{eq:background-compression}
\end{equation}
where \(\mathbf{Q}_{\mathrm{bg}}=[\mathbf{q}^{\mathrm{bg}}_1;\ldots;\mathbf{q}^{\mathrm{bg}}_M]\). The final sequence \(\widehat{\mathbf{H}}\) has fixed length \(K+M=B\) and is concatenated with the question embeddings before being fed into the language model. The score summary, Beta parameterization, and deterministic inference rule of this allocator are provided in Appendix D.2.

\paragraph{Evidence-Aware Reward and Policy Optimization.}

After E-SFT, we enable sparse visual selection and optimize the policy with evidence-aware rewards. For each sampled trajectory, the model predicts an answer \(A\) and an evidence tag \(E\). The first key reward aligns the predicted evidence with annotated temporal positions and spatial cells. Let \(\widehat{\mathcal{T}}(E)\) and \(\widehat{\mathcal{G}}(E)\) be the temporal indices and spatial cells parsed from \(E\), and let \(\mathcal{T}^{*}\) and \(\mathcal{G}^{*}\) denote the corresponding annotations. We compute
\begin{equation}
\begin{aligned}
R_{\mathrm{T}}
&=
\frac{|\widehat{\mathcal{T}}(E)\cap\mathcal{T}^{*}|}{|\mathcal{T}^{*}|},
\qquad
R_{\mathrm{G}}
=
\frac{|\widehat{\mathcal{G}}(E)\cap\mathcal{G}^{*}|}{|\mathcal{G}^{*}|},\\
R_{\mathrm{st}}
&=
\begin{cases}
(R_{\mathrm{T}}+R_{\mathrm{G}})/2, & \mathcal{T}^{*},\mathcal{G}^{*}\ \text{available},\\
R_{\mathrm{T}}, & \mathcal{T}^{*}\ \text{only},\\
R_{\mathrm{G}}, & \mathcal{G}^{*}\ \text{only}.
\end{cases}
\end{aligned}
\label{eq:evidence-alignment-reward}
\end{equation}

Evidence alignment specifies which spatiotemporal regions should support the answer, but it does not by itself enforce efficient visual routing. We therefore map target annotations to visual-token indices \(\mathcal{I}^{*}\subseteq[L]\). Let \(\mathcal{I}_{K}\) be the Top-\(K\) retained indices, \(\mathcal{I}_{\mathrm{rem}}=[L]\setminus\mathcal{I}_{K}\) the residual tokens sent to background compression, and \(\mathcal{I}_{\mathrm{bg}}^{*}=[L]\setminus\mathcal{I}^{*}\) the non-target tokens. The background compression reward is
\begin{equation}
R_{\mathrm{bg}}
=
\frac{|\mathcal{I}_{\mathrm{rem}}\cap\mathcal{I}_{\mathrm{bg}}^{*}|}{|\mathcal{I}_{\mathrm{bg}}^{*}|}
+
\frac{|\mathcal{I}_{\mathrm{rem}}|}{L\,M}.
\label{eq:background-compression-reward}
\end{equation}
The first term rewards routing redundant non-target tokens to the compression branch, while the second favors summarizing more residual tokens with fewer background slots.

The overall return combines answer correctness, evidence alignment, background compression, and a cost penalty for overly broad evidence tags:
\begin{equation}
\mathcal{R}
=
\lambda_{\mathrm{ans}}R_{\mathrm{ans}}
+
g_{\mathrm{ans}}
\left(
\lambda_{\mathrm{st}}R_{\mathrm{st}}
+
\lambda_{\mathrm{bg}}R_{\mathrm{bg}}
-
\lambda_{\mathrm{cost}}C_{\mathrm{cost}}
\right),
\label{eq:evidence-aware-reward}
\end{equation}
where \(R_{\mathrm{ans}}=\mathbb{I}[A=A^{*}]\), and \(g_{\mathrm{ans}}=\mathbb{I}[A=A^{*}]\mathbb{I}[E\ \text{is valid}]\) gates the evidence rewards to trajectories with a correct answer and a parseable evidence tag. We optimize the policy using Group Relative Policy Optimization (GRPO)~\citep{shao2024deepseekmath}. More details are provided in the Appendix D.5.

\section{Experiments}
\label{sec:experiments}

\subsection{Experimental Setup}

Training follows the two-stage protocol: evidence-aware SFT on the full visual sequence, followed by GRPO with sparse evidence focusing and background compression enabled. All training is conducted on 8$\times$ NVIDIA 80\,GB A100 GPUs. Unless otherwise specified, we use LoRA fine-tuning with BF16 precision, a maximum sequence length of 2048, gradient checkpointing, per-device batch size 1, gradient accumulation 8, and a cosine learning-rate schedule. The base learning rate is \(1\times10^{-5}\), with warmup ratio 0.03 and weight decay 0.01. LoRA uses rank 32, alpha 64, and dropout 0.05. Complete optimization settings and preprocessing details are provided in Appendix D.7. Accuracy is the proportion of correctly answered questions among all evaluated questions. SP, TP, AP, CER, SER, SLR, and SCR denote the accuracies of the seven capability groups. T-Hit (TH) and R-Hit (RH) measure whether the predicted temporal index \(T\) overlaps the annotated key frame or temporal window, and whether the predicted spatial grid index \(R\) overlaps the annotated target region, respectively.

\subsection{Comprehensive Evaluation}
\label{sec:effectiveness}

We conduct a comprehensive evaluation using a diverse set of advanced vision-language backbones spanning parameter scales from 1B to over 200B. For each backbone, we compare our \method{} with several representative training strategies. All trainable methods train on RSVideo-Instruct and report results on \bench{}. To ensure a fair comparison, we adopt consistent experimental settings for all methods.

\begin{table}[t]
\centering
\begingroup
\fontsize{6.8pt}{7.6pt}\selectfont
\setlength{\tabcolsep}{1.25pt}
\renewcommand{\arraystretch}{0.94}

\def\tblbest#1{\textbf{#1}}
\def\tblsecond#1{\underline{#1}}

\begin{tabular*}{\columnwidth}
{@{\extracolsep{\fill}}lccccccc@{}}
\toprule
Model & Direct & SFT & E-SFT & OG & TG & GSPO & \textbf{Ours} \\
\midrule

\rowcolor{black!6}
\multicolumn{8}{l}{\textit{Open-source VLMs: $<7$B}} \\

InternVL3-1B
& 20.13
& 21.08
& 22.32
& 23.08
& \tblsecond{24.37}
& 24.01
& \tblbest{25.73} \\

InternVL3.5-1B
& 20.46
& 21.55
& 22.91
& 23.62
& 24.18
& \tblsecond{24.67}
& \tblbest{26.23} \\

InternVL3-2B
& 21.92
& 23.10
& 24.54
& 25.11
& 25.96
& \tblsecond{26.42}
& \tblbest{27.87} \\

InternVL3.5-2B
& 22.34
& 23.77
& 25.16
& 25.79
& 26.38
& \tblsecond{26.91}
& \tblbest{28.62} \\

Qwen3-VL-2B
& 22.91
& 24.46
& 26.08
& 26.64
& \tblsecond{27.79}
& 27.38
& \tblbest{29.61} \\

Qwen2.5-VL-3B
& 21.68
& 22.97
& 24.21
& 25.02
& \tblsecond{26.06}
& 25.49
& \tblbest{27.68} \\

Qwen3-VL-4B
& 25.46
& 27.53
& 29.86
& 30.47
& 31.34
& \tblsecond{31.82}
& \tblbest{33.56} \\

\rowcolor{black!6}
\multicolumn{8}{l}{\textit{Open-source VLMs: 7--9B}} \\

Qwen2.5-VL-7B
& 24.88
& 26.24
& 28.41
& 29.02
& 29.73
& \tblsecond{30.16}
& \tblbest{31.84} \\

LLaVA-OneVision-7B
& 23.62
& 24.72
& 26.37
& 27.18
& 27.84
& \tblsecond{28.31}
& \tblbest{29.89} \\

InternVL3-8B
& 26.78
& 28.68
& 30.62
& 31.23
& \tblsecond{32.57}
& 32.11
& \tblbest{34.36} \\

InternVL3.5-8B
& 25.87
& 27.46
& 29.78
& 30.37
& \tblsecond{31.72}
& 31.16
& \tblbest{33.59} \\

Qwen3-VL-8B
& 29.74
& 31.71
& 33.68
& 34.22
& 35.18
& \tblsecond{35.61}
& \tblbest{37.27} \\

GLM-4.6V-Flash-9B
& 30.48
& 32.05
& 34.07
& 34.76
& 35.38
& \tblsecond{36.09}
& \tblbest{37.83} \\

\rowcolor{black!6}
\multicolumn{8}{l}{\textit{Open-source VLMs: 10--30B}} \\

InternVL3-14B
& 29.12
& 31.25
& 33.62
& 34.31
& 35.24
& \tblsecond{35.81}
& \tblbest{37.21} \\

InternVL3.5-14B
& 28.36
& 30.52
& 33.31
& 33.88
& \tblsecond{35.27}
& 34.79
& \tblbest{37.37} \\

Qwen3.6-27B
& 35.30
& 35.92
& 36.78
& 37.62
& 37.99
& \tblsecond{38.47}
& \tblbest{40.63} \\

\rowcolor{black!6}
\multicolumn{8}{l}{\textit{Open-source VLMs: 30--70B}} \\

Qwen2.5-VL-32B
& 30.92
& 32.77
& 34.82
& 35.47
& 36.31
& \tblsecond{36.78}
& \tblbest{38.29} \\

Qwen3-VL-32B
& 32.74
& 34.44
& 36.75
& 37.56
& \tblsecond{38.77}
& 38.31
& \tblbest{40.38} \\

InternVL3-38B
& 31.54
& 33.72
& 35.83
& 36.42
& 37.31
& \tblsecond{37.78}
& \tblbest{39.61} \\

InternVL3.5-38B
& 31.08
& 33.22
& 35.12
& 35.81
& \tblsecond{36.89}
& 36.47
& \tblbest{38.79} \\

\rowcolor{black!6}
\multicolumn{8}{l}{\textit{Open-source VLMs: 70--80B}} \\

Qwen2.5-VL-72B
& 31.36
& 33.38
& 35.24
& 35.82
& 36.41
& \tblsecond{37.03}
& \tblbest{38.68} \\

LLaVA-OneVision-72B
& 28.74
& 30.21
& 32.11
& 32.68
& \tblsecond{33.95}
& 33.52
& \tblbest{35.47} \\

LLaVA-Video-72B
& 29.95
& 31.72
& 33.89
& 34.52
& 35.11
& \tblsecond{35.68}
& \tblbest{37.42} \\

InternVL3-78B
& 33.02
& 35.06
& 37.16
& 37.73
& \tblsecond{38.66}
& 38.21
& \tblbest{40.36} \\

\rowcolor{black!6}
\multicolumn{8}{l}{\textit{Open-source VLMs: $>200$B}} \\

Qwen3-VL-235B
& 33.62
& 35.67
& 37.21
& 37.92
& 38.58
& \tblsecond{39.07}
& \tblbest{40.49} \\

InternVL3.5-241B
& 33.14
& 35.06
& 36.98
& 37.53
& \tblsecond{38.84}
& 38.36
& \tblbest{40.31} \\

\bottomrule
\end{tabular*}
\endgroup

\caption{
Comparison of different training strategies across various vision-language backbones on \bench{}. We compare our \method{} with Direct Inference, supervised fine-tuning (SFT), evidence-aware SFT (E-SFT), and three policy optimization algorithms: outcome-only GRPO (OG)~\citep{shao2024deepseekmath}, T-GRPO(TG)~\citep{feng2025videor1}, and GSPO~\citep{zheng2025gspo}. The best and second-best results for each backbone are highlighted in bold and underline, respectively.
}
\label{tab:rsv-effectiveness}
\end{table}

Table~\ref{tab:rsv-effectiveness} reports the evaluation results on \bench{}. The gains of E-SFT over Direct Inference confirm that RSVideo-Instruct provides effective supervision for remote-sensing video understanding. Building on E-SFT, RL-based approaches further improve performance, while \method{} ranks first for every evaluated backbone. Specifically, compared with Direct Inference, \method{} achieves a maximum absolute improvement of 9.01\% with InternVL3.5-14B and attains the highest accuracy of 40.63\% with Qwen3.6-27B, demonstrating strong generalization across backbone architectures.

\subsection{Ablation Study}

\begin{table}[t]
\centering
\fontsize{6.0pt}{6.8pt}\selectfont
\setlength{\tabcolsep}{0.45pt}
\renewcommand{\arraystretch}{0.94}
\begin{tabular*}{\columnwidth}{@{\extracolsep{\fill}}ccccccccccc@{}}
\toprule
E-SFT & $R_{\mathrm{ans}}$ & $R_{\mathrm{st}}$ & $R_{\mathrm{bg}}$ &
Accuracy $\uparrow$ & AP $\uparrow$ & SER $\uparrow$ &
SLR $\uparrow$ & SCR $\uparrow$ & TH $\uparrow$ & RH $\uparrow$ \\
\midrule

\yesmark & \nomark & \nomark & \nomark
& 36.78 & 35.96 & 31.41 & 36.48 & 26.67 & 50.4 & 47.4 \\

\yesmark & \yesmark & \nomark & \nomark
& 36.90 & 36.86 & 32.12 & 36.74 & 26.91 & 51.2 & 48.3 \\

\yesmark & \yesmark & \yesmark & \nomark
& 38.85 & 39.12 & 33.90 & 37.86 & 27.76 & 55.4 & 52.8 \\

\yesmark & \yesmark & \nomark & \yesmark
& 39.10 & 39.96 & 31.42 & 37.61 & 27.48 & 52.1 & 53.9 \\

\yesmark & \yesmark & \yesmark & \yesmark
& \textbf{40.63} & \textbf{41.73} & \textbf{35.79}
& \textbf{39.43} & \textbf{29.18}
& \textbf{57.4} & \textbf{54.8} \\
\bottomrule
\end{tabular*}
\caption{
Ablation of the evidence-aware rewards.
}
\label{tab:rsv-reward}
\end{table}

To evaluate the contributions of different reinforcement learning rewards, we use Qwen3.6-27B as the backbone and conduct ablation studies under identical experimental settings. Starting from E-SFT, incorporating only the answer-correctness reward (\(R_{\mathrm{ans}}\)) provides a further but limited improvement, achieving 36.90\% accuracy. Adding the spatiotemporal evidence alignment reward (\(R_{\mathrm{st}}\)) brings the major performance gain, improving accuracy by 1.95 percentage points and increasing TH/RH by 4.2/4.5 points. The improvements are also reflected in AP and SER, which increase from 36.86\% and 32.12\% to 39.12\% and 33.90\%, demonstrating the importance of evidence grounding. Adding only the background compression reward (\(R_{\mathrm{bg}}\)) instead improves accuracy by 2.20 percentage points and RH by 5.6 points, but yields only a 0.9-point increase in TH and a 0.70-point decrease in SER, indicating that it primarily promotes region-level focusing rather than temporally consistent evidence grounding. Combining all rewards achieves the best performance, improving accuracy from 36.90\% to 40.63\% and TH/RH from 51.2\%/48.3\% to 57.4\%/54.8\%. These rewards are complementary: spatiotemporal alignment improves temporal grounding, while background compression enhances regional focus by reducing redundant context.

\subsection{Transfer Evaluation}
\label{sec:transfer-evaluation}

The in-domain results on \bench{} do not by themselves establish whether the learned evidence-focusing policy generalizes beyond the training and evaluation distribution. We therefore test each vision-language backbone and its corresponding checkpoint trained with \method{} on RSVideo-Instruct across two general video benchmarks, MVBench and Video-MME, and two aerial or remote-sensing video benchmarks, UrbanVideo-Bench and SIS-Bench~\citep{li2024mvbench,fu2025videomme,zhao2025urbanvideo,zou2026sisbench}. The paired rows in Table~\ref{tab:transfer-evaluation} keep the backbone and evaluation protocol fixed and changes only whether \method{} training is applied, thereby isolating the effect of the proposed training strategy. Video-MME scores use the without-subtitle setting throughout.

\begin{table}[t]
\centering
\small
\setlength{\tabcolsep}{2pt}
\renewcommand{\arraystretch}{1.03}
\resizebox{\columnwidth}{!}{%
\begin{tabular}{@{}lcccc@{}}
\toprule
Model / Setting & MVBench $\uparrow$ & Video-MME $\uparrow$ & UrbanVideo-Bench $\uparrow$ & SIS-Bench $\uparrow$ \\
\midrule
InternVL3.5-2B & 58.92 & 67.34 & 34.76 & 46.21 \\
InternVL3.5-2B + \method{} & 59.68 & 67.81 & 36.19 & 48.04 \\
Qwen3-VL-4B & 62.44 & 70.86 & 37.62 & 50.33 \\
Qwen3-VL-4B + \method{} & 63.31 & 71.29 & 39.08 & 52.11 \\
Qwen2.5-VL-7B & 64.78 & 73.52 & 39.15 & 53.46 \\
Qwen2.5-VL-7B + \method{} & 65.42 & 73.96 & 40.77 & 55.02 \\
LLaVA-OV-7B & 61.35 & 72.14 & 36.83 & 49.27 \\
LLaVA-OV-7B + \method{} & 62.09 & 72.48 & 38.31 & 50.96 \\
InternVL3-8B & 66.91 & 75.63 & 41.28 & 55.19 \\
InternVL3-8B + \method{} & 67.56 & 76.02 & 42.91 & 56.88 \\
GLM-4.6V-Flash-9B & 69.47 & 78.38 & 43.52 & 58.64 \\
GLM-4.6V-Flash-9B + \method{} & 70.16 & 78.74 & 45.09 & 60.31 \\
Qwen3.6-27B & 75.50 & 87.70 & 49.71 & 68.83 \\
Qwen3.6-27B + \method{} & 76.08 & 88.04 & 51.58 & 70.91 \\
Qwen2.5-VL-72B & 72.34 & 82.16 & 46.27 & 63.58 \\
Qwen2.5-VL-72B + \method{} & 72.91 & 82.49 & 47.72 & 65.03 \\
LLaVA-OV-72B & 70.82 & 80.54 & 43.68 & 60.42 \\
LLaVA-OV-72B + \method{} & 71.36 & 80.87 & 45.01 & 61.93 \\
InternVL3-78B & 73.18 & 83.41 & 47.84 & 65.26 \\
InternVL3-78B + \method{} & 73.79 & 83.76 & 49.28 & 66.71 \\
Qwen3-VL-235B & 74.06 & 85.12 & 48.63 & 66.17 \\
Qwen3-VL-235B + \method{} & 74.62 & 85.38 & 50.02 & 67.54 \\
\bottomrule
\end{tabular}}
\caption{Transfer evaluation on external video benchmarks after training on RSVideo-Instruct. Values are overall accuracies (\%), where higher is better. Each pair uses the same backbone and evaluation protocol; no external benchmark data are used for training or model selection. OV denotes OneVision.}
\label{tab:transfer-evaluation}
\end{table}
\begin{figure}[tbp]
\centering
\includegraphics[width=\columnwidth]{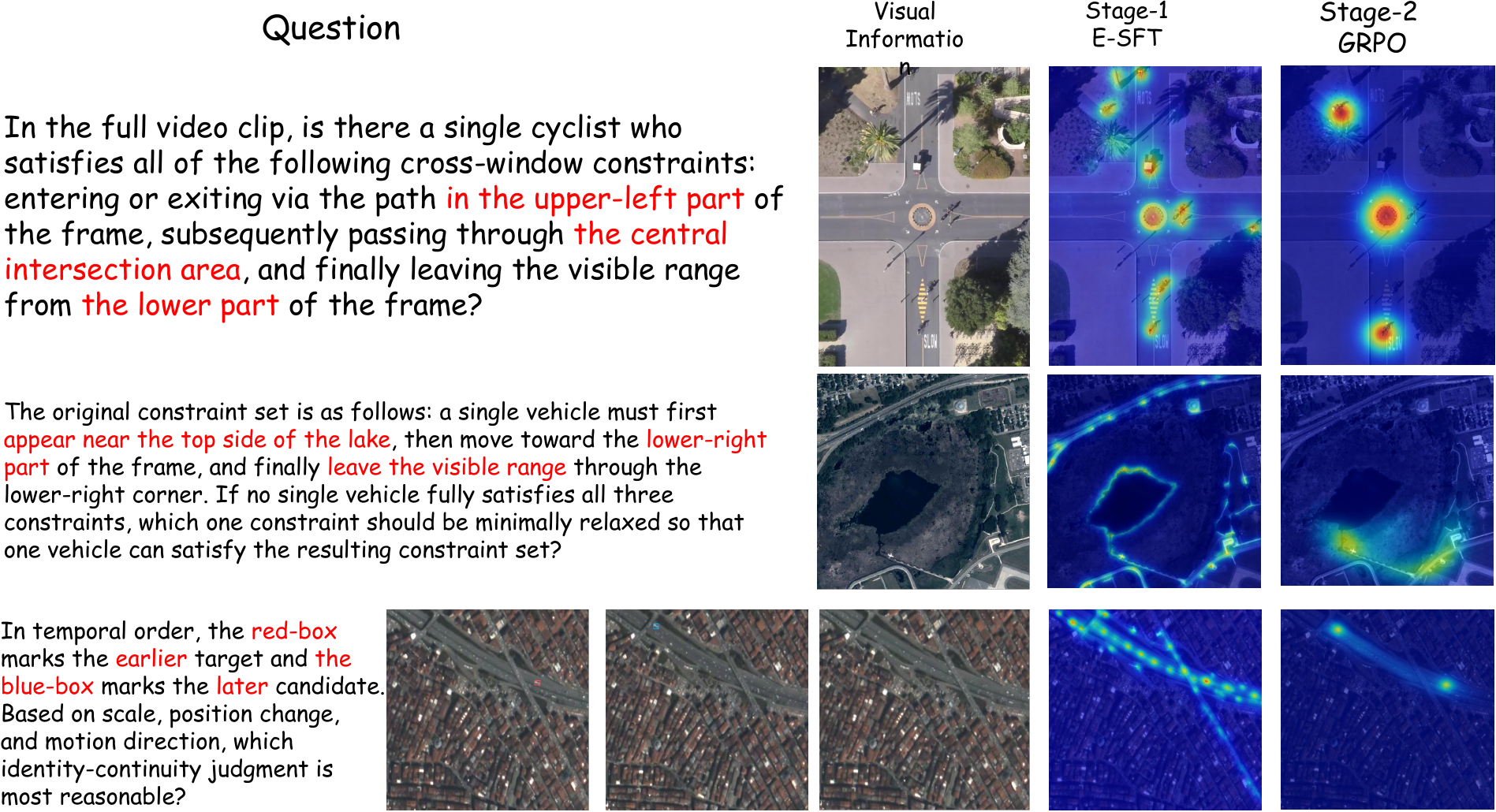}
\caption{Qualitative comparison of question-conditioned visual evidence selection before and after reinforcement learning. From left to right, each row presents the question, the relevant video frame(s), the attention map of the Stage-1 E-SFT model, and that of the Stage-2 GRPO model. The three examples respectively require cross-window trajectory reasoning, minimal relaxation of a spatiotemporal constraint set, and temporal identity-continuity judgment. After GRPO, the attention maps are more concentrated on task-relevant targets, regions, and transition frames.}
\label{fig:qualitative-evidence-selection}
\end{figure}

Table~\ref{tab:transfer-evaluation} shows that all evaluated backbones improve on external benchmarks after \method{} training. Averaged over paired backbones, the gains are 0.66, 0.37, 1.52, and 1.64 percentage points on MVBench, Video-MME, UrbanVideo-Bench, and SIS-Bench, respectively. The consistent improvements on the two general video benchmarks indicate that the gains are not confined to \bench{}, while the larger average improvements on UrbanVideo-Bench and SIS-Bench suggest that the learned evidence-focusing policy transfers particularly well to aerial and remote-sensing video settings.

\subsection{Qualitative Evidence Selection Comparison}
\label{sec:qualitative-evidence-selection}

Higher answer accuracy alone does not reveal whether the policy identifies the visual evidence that supports its predictions. We therefore compare the question-conditioned evidence maps before and after reinforcement learning to examine how the learned policy changes its spatiotemporal focus. Figure~\ref{fig:qualitative-evidence-selection} contrasts the Stage-1 E-SFT model with the Stage-2 GRPO model on three examples requiring cross-window trajectory reasoning, minimal spatiotemporal-constraint relaxation, and temporal identity-continuity judgment.

Compared with Stage-1 E-SFT, Stage-2 GRPO produces more concentrated evidence maps on the question-relevant targets, spatial regions, and transition frames, while suppressing responses from redundant background areas. This pattern is consistent across the three illustrated reasoning cases and provides qualitative evidence that reinforcement learning improves spatiotemporal evidence grounding rather than merely changing the final answer distribution.

\section{Conclusion}
In this paper, we investigate continuous remote-sensing video understanding and reveal that effective reasoning requires accurately identifying and leveraging sparse spatiotemporal evidence. To facilitate systematic evaluation, we introduce \rsvdata{}, including RSVideo-Instruct for training and validation and \bench{} for testing. Comprehensive evaluations under a unified five-choice protocol demonstrate that current models often fail to ground their predictions on relevant targets, frames, and scene contexts. Based on these findings, we propose \method{}, a reinforcement learning framework for evidence-aware spatiotemporal focusing, which adaptively preserves informative evidence tokens, compresses redundant background context, and optimizes evidence-aware rewards under a fixed visual token budget. Extensive experiments demonstrate that \method{} consistently outperforms existing training methods across diverse vision-language backbones.


{
  \small
  \bibliographystyle{ieeenat_fullname}
  \bibliography{rsvb_aaai2027}
}

\clearpage
\appendix
\twocolumn[{%
\begin{minipage}{\textwidth}

\section*{Appendix Contents}

\begingroup
\small
\setlength{\parindent}{0pt}
\setlength{\parskip}{2pt}

\newcommand{\appcontentsline}[2]{%
  \noindent
  \textbf{Appendix~\ref{#1}. #2}%
  \nobreak\dotfill\mbox{\pageref{#1}}\par
}

\newcommand{\appcontentsubline}[2]{%
  \noindent
  \hangindent=1.5em
  \hangafter=1
  \hspace*{1.5em}\ref{#1}\enspace #2%
  \nobreak\dotfill\mbox{\pageref{#1}}\par
}

\appcontentsline{app:dataset_details}
  {Dataset Construction Details}
\appcontentsubline{app:source-license}
  {Video Sources, Licensing, and Filtering}
\appcontentsubline{app:taxonomy-leaves}
  {Task Taxonomy, Operational Leaves, and Question Schema}
\appcontentsubline{app:split-statistics}
  {Split Construction and Data Statistics}

\medskip

\appcontentsline{app:data_engine}
  {Data Engine and Annotation Protocol}
\appcontentsubline{app:construction-pipeline}
  {Complete Construction and Annotation Pipeline}
\appcontentsubline{app:leaf-guided-construction}
  {Leaf-guided Question, Answer, and Candidate Construction}
\appcontentsubline{app:expert-review}
  {Expert Annotation, Multi-round Review, and Adjudication}
\appcontentsubline{app:quality-release}
  {Quality Control, Label Correction, and Release Audit}

\medskip

\appcontentsline{app:metrics}
  {Evaluation Protocol and Metrics}
\appcontentsubline{app:video-input}
  {Video Input Protocol}
\appcontentsubline{app:answer-parsing}
  {Answer Parsing and Accuracy}
\appcontentsubline{app:evidence-localization}
  {Evidence Localization Metrics}
\appcontentsubline{app:score-aggregation}
  {Score Aggregation}
\appcontentsubline{app:validity-audits}
  {Validity Audits}

\medskip

\appcontentsline{app:method-details}
  {Additional Details of \method{}}
\appcontentsubline{app:evidence-scores}
  {Evidence Scores}
\appcontentsubline{app:budget-allocator}
  {Budget Allocator}
\appcontentsubline{app:evidence-aware-sft}
  {Evidence-Aware SFT}
\appcontentsubline{app:cost-penalty}
  {Cost Penalty}
\appcontentsubline{app:grpo}
  {Group Relative Policy Optimization}
\appcontentsubline{app:evidence-parser}
  {Evidence Mapping, Tag Parsing, and Run Records}
\appcontentsubline{app:training-preprocessing}
  {Training Configuration, Video Preprocessing, and Reproducibility}

\medskip

\appcontentsline{app:additional_analyses}
  {Additional Diagnostic and Robustness Analyses}
\appcontentsubline{app:experiment-scope}
  {Complete Benchmark Results of Existing VLMs}
\appcontentsubline{app:video-evidence-audit}
  {Validity Audit: Dependence on Video Evidence}
\appcontentsubline{app:eta-sensitivity}
{Time--Region Cell Pooling Ratio}
\appcontentsubline{app:budget-sensitivity}
  {Visual-token Budget and Background Context}
\appcontentsubline{app:inference-efficiency}
  {Inference Efficiency}
\appcontentsubline{app:sparse-integration}
  {Sparse-token Integration}
\appcontentsubline{app:rl-sensitivity}
  {RL Optimization Sensitivity}
\appcontentsubline{app:reward-weight}
  {Reward-weight Sensitivity}

\medskip

\appcontentsline{app:qualitative_examples}
  {Leaf-complete Qualitative Examples}

\medskip

\appcontentsline{app:datasheets}
  {Datasheets}
\appcontentsubline{app:datasheets-motivation}
  {Motivation}
\appcontentsubline{app:datasheets-composition}
  {Composition}
\appcontentsubline{app:datasheets-collection}
  {Collection Process}
\appcontentsubline{app:datasheets-preprocessing}
  {Preprocessing, Cleaning, and Labeling}
\appcontentsubline{app:datasheets-uses}
  {Uses}
\appcontentsubline{app:datasheets-distribution}
  {Distribution}
\appcontentsubline{app:datasheets-maintenance}
  {Maintenance}

\endgroup

\end{minipage}
}]
\clearpage

\section{Dataset Construction Details}
\label{app:dataset_details}

\subsection{Video Sources, Licensing, and Filtering}
\label{app:source-license}
\rsvdata{} is constructed from eight publicly available remote-sensing video sources spanning UAV, aerial, overhead, and satellite observations.  The sources were selected to expose large scene extent, small or crowded targets, camera motion, temporally sparse events, and cross-platform appearance changes.  Source membership alone is not an acceptance criterion: every retained segment passes the clip-level audit below.

\textbf{AU-AIR}~\citep{bozcan2020auair}.
AU-AIR records low-altitude UAV traffic scenes with varied viewpoints and visible road users.  Its vehicle-rich sequences support target identification, counting, and spatial-relation questions under camera motion.

\textbf{DTB70}~\citep{li2017dtb70}.
DTB70 contains UAV tracking sequences with scale variation, occlusion, deformation, and rapid viewpoint change.  We use these temporal patterns to construct moving-target localization, trajectory-continuity, and occlusion-recovery tasks.

\textbf{ERA}~\citep{mou2020era}.
ERA provides aerial videos of human activities and events observed from overhead platforms.  Its event-centered clips support action recognition, event-state analysis, abnormal-evidence monitoring, and action--event coupling.

\textbf{OOTB}~\citep{chen2024ootb}.
OOTB contributes satellite video sequences in which targets occupy only a small fraction of the frame and their appearance changes across time.  These properties support small-target motion analysis and state recovery across temporal windows.

\textbf{Okutama-Action}~\citep{barekatain2017okutama}.
Okutama-Action captures concurrent human activities from UAV viewpoints in outdoor scenes.  Its temporally localized action annotations support questions about action transitions, duration, and intervals in which multiple targets remain jointly visible.

\textbf{SatSOT}~\citep{zhao2022satsot}.
SatSOT contains satellite videos for tracking tiny moving targets under scale change, background interference, and intermittent visibility.  We use it for target continuity, small-object localization, and recovery after temporary disappearance.

\textbf{Stanford Drone Dataset}~\citep{robicquet2016sdd}.
The Stanford Drone Dataset provides overhead trajectories of pedestrians, bicycles, and vehicles in shared outdoor spaces.  Its multi-agent scenes support target identification, co-visibility, trajectory interaction, and spatial-relation reasoning.

\begin{table}[t]
\centering
\small
\setlength{\tabcolsep}{3pt}
\renewcommand{\arraystretch}{1.08}
\begin{tabularx}{\columnwidth}{P{0.30\columnwidth}P{0.16\columnwidth}Y}
\toprule
Group & Task IDs & Task scope \\
\midrule
G1 Scene Perception & T1--T3 & scene type; viewpoint type; static spatial relation \\
G2 Target Perception & T4--T6 & target recognition; target counting; reference-image target localization \\
G3 Action Perception & T7--T9 & action recognition; action transition; action counting \\
G4 Complex-Environment Reasoning & T10--T11 & event reasoning; risk assessment \\
G5 Spatiotemporal Evolution Reasoning & T12--T13 & global spatiotemporal understanding; long-term understanding \\
G6 Spatiotemporal Localization Reasoning & T14--T15 & static-spatial dynamization; motion-event matching verification \\
G7 Spatiotemporal Consistency Reasoning & T16--T17 & spatial relation consistency; target consistency \\
\bottomrule
\end{tabularx}
\caption{The 17 tasks grouped into the seven capability groups used by the main-paper evaluation.}
\label{tab:app-taxonomy}
\end{table}

\textbf{VISO}~\citep{yin2021viso}.
VISO provides satellite videos with sparse, small moving objects against wide-area backgrounds.  It supports candidate localization, motion reasoning, and negative-evidence questions in which a proposed target or relation is absent from the observed interval.

For every candidate segment, annotators inspect decodability, temporal continuity, observable target/action evidence, and the absence of unusable corruption or watermarks.  They then determine whether the segment supports at least one well-posed task in the taxonomy below.  We retain clips only when the correct answer can be established from the released visual input and when a concrete temporal window can be identified.  Segments with accidental missing frames, ambiguous targets, severe corruption, duplicated content, or answer cues available only outside the released input are rejected.  This procedure produces \textbf{4,629 audited evidence clips}, covering \textbf{17.02 hours} and \textbf{1,473,150 decoded frames}.

An item whose intended answer is \emph{insufficient evidence} is treated differently from a defective clip.  Such an item is retained only when insufficiency itself is a well-posed visual conclusion---for example, all five candidate claims require an unobserved continuation---and independent reviewers verify that no released frame resolves the alternatives.  A clip that is merely corrupted or accidentally truncated is rejected rather than converted into an insufficiency item.

\begin{table*}[t]
\centering
\scriptsize
\setlength{\tabcolsep}{3pt}
\renewcommand{\arraystretch}{1.06}
\begin{tabularx}{\textwidth}{C{0.045\textwidth}P{0.21\textwidth}P{0.31\textwidth}Y}
\toprule
Leaf & Parent task & Leaf capability & Required observable evidence \\
\midrule
01 & T1 Scene Type & Scene type & Global scene appearance. \\
02 & T2 Viewpoint Type & Viewpoint type & View geometry and perspective. \\
03 & T3 Static Spatial Relation & Spatial relation between static land objects & Spatial relation in one or more frames. \\
04 & T4 Target Recognition & Target identification & Localized target appearance. \\
05 & T5 Target Counting & Target counting & All qualifying targets in the stated region. \\
06 & T6 Reference-Image Target Localization & Reference-image target localization & Reference cue and target region. \\
07 & T7 Action Recognition & Action recognition & Motion/state evidence across frames. \\
08 & T8 Action Transition & Action transition & Before/after action states. \\
09 & T9 Action Counting & Action counting & Event boundaries for each instance. \\
10 & T10 Event Reasoning & Environmental constraint reasoning & Event plus relevant scene context. \\
11 & T10 Event Reasoning & Target state inference & Target/event interaction evidence. \\
12 & T10 Event Reasoning & Main visible activity recognition & Dominant visible activity over the evidence span. \\
13 & T11 Risk Assessment & Disaster evidence monitoring & Hazard cue and affected region/target. \\
14 & T11 Risk Assessment & Abnormal behavior & Contextual norm and observed deviation. \\
15 & T12 Global Spatiotemporal Understanding & Long-term trajectory summary for small moving targets & Target trajectory over a long temporal window. \\
16 & T12 Global Spatiotemporal Understanding & Displacement and speed estimation & At least two temporal positions per moving target. \\
17 & T13 Long-Term Understanding & Temporal ordering & Separate evidence spans for ordered events. \\
18 & T13 Long-Term Understanding & Action-duration comparison & Comparable action intervals or state durations. \\
19 & T13 Long-Term Understanding & Concurrent visibility interval for multiple targets & Overlapping visibility intervals for queried targets. \\
20 & T14 Static-Spatial Dynamization & Single-event spatiotemporal localization & Event interval and local region. \\
21 & T15 Motion-Event Matching Verification & Regional trajectory co-occurrence & Trajectory, region, and event co-occurrence evidence. \\
22 & T16 Spatial Relation Consistency & Relation-constrained candidate localization & Relation checks across the relevant frames. \\
23 & T16 Spatial Relation Consistency & Reasoned counting of small moving targets & Constraint region plus all qualifying targets. \\
24 & T17 Target Consistency & Cross-window constraint satisfiability & Identity attributes across windows. \\
25 & T17 Target Consistency & Target re-identification after occlusion & Pre/post gap identity evidence. \\
26 & T17 Target Consistency & Minimal constraint relaxation & Candidate comparison under explicit constraints. \\
\bottomrule
\end{tabularx}
\caption{\textbf{Operational definitions of the 26 leaves in \rsvdata{}}.  Every item is associated with exactly one task ID and one active leaf ID.  The required observable evidence specifies the visual or temporal evidence used to validate an item.}
\label{tab:app-leaves}
\end{table*}

\textbf{License-aware release.}
The public release distributes derived metadata, questions, answer options, split identifiers, and evaluation code while retaining the original providers' terms.  By default, it provides source provenance, temporal boundaries, checksums, and reconstruction instructions rather than repackaging third-party pixels.  A source video is redistributed only when its verified source manifest explicitly permits that use; the frozen manifest records the corresponding license identifier.  The release does not claim ownership of the underlying videos, and users remain responsible for the original providers' terms.

\subsection{Task Taxonomy, Operational Leaves, and Question Schema}
\label{app:taxonomy-leaves}
The taxonomy has two capability levels, seven capability groups, 17 tasks, and 26 operational leaves.  L1 \emph{Perception} covers scene, target, and action understanding.  L2 \emph{Reasoning} covers complex-environment reasoning, spatiotemporal evolution reasoning, spatiotemporal localization reasoning, and spatiotemporal consistency reasoning.  Table~\ref{tab:app-taxonomy} summarizes the operational organization used here.

Table~\ref{tab:app-leaves} enumerates every released operational leaf.  The stable schema uses consecutive identifiers 01--26.

Each item is a fixed five-choice multiple-choice question.  Its released record contains a video/evidence-clip identifier, the question, five shuffled options, the gold option, capability-level/group/task/leaf labels, and, when applicable, a visual mark and temporal evidence annotation.  Questions require visual inspection rather than a source name or memorized fact.  Distractors are drawn from nearby categories, confusable actions, reversed temporal order, or scene-incompatible relations.  The answer is one canonical option label.

\subsection{Split Construction and Data Statistics}
\label{app:split-statistics}
Table~\ref{tab:app-splits} summarizes the released training, validation, and test partitions and their roles in model development and final evaluation.  Splits are created at the evidence-clip level, so all questions tied to the same released segment remain in one partition.  We audit source identifiers, near-duplicate clips, and overlapping temporal ranges to avoid cross-split leakage.  RSVideo-Instruct contains 8,042 train/validation instances, and \bench{} contains 2,731 fixed test instances.  The test set includes 653 L1 and 2,078 L2 items.  All aggregate scores are computed over this fixed item set; they are not unweighted averages over tasks or sources.

\begin{figure*}[t]
\centering
\includegraphics[width=\textwidth]{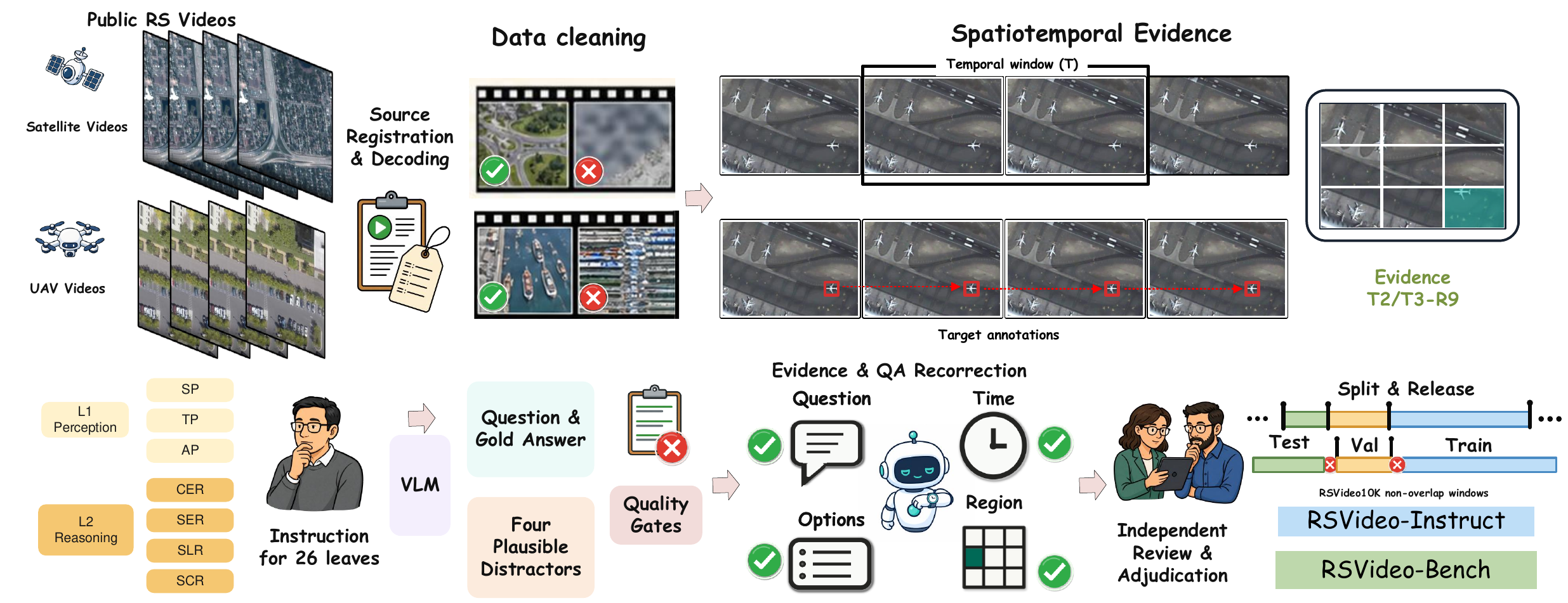}
\caption{Complete RSVideo data-construction pipeline.  Public videos first pass source registration, decoding, quality screening, and task/leaf binding before question construction.  Each candidate item is then grounded by temporal and, when needed, spatial evidence; paired with one gold answer and four plausible distractors; checked through joint evidence--label correction, option permutation, independent review, and expert adjudication; and finally frozen after clip-level split isolation and release audit.  Human verification determines the final labels and evidence fields.}
\label{fig:app-pipeline}
\end{figure*}
\section{Data Engine and Annotation Protocol}
\label{app:data_engine}

\subsection{Complete Construction and Annotation Pipeline}
\label{app:construction-pipeline}
The construction procedure converts each accepted source video into an auditable evidence record through the following ordered stages:
\begin{enumerate}[leftmargin=*,label=(\arabic*),itemsep=1pt,topsep=2pt]
\item \textbf{Source registration and decoding.}  We record each video's provenance, usage terms, and metadata, decode it into candidate frames, and discard material that is unreadable, temporally unstable, or below the required visual-quality threshold.
\item \textbf{Clip screening.}  We locate temporal spans with valid observable content and retain a span only if its released frames support at least one well-posed task.
\item \textbf{Task and leaf binding.}  Before question writing, each retained sample is assigned exactly one task ID and one active leaf ID, enforcing a strict correspondence with the predefined taxonomy.
\item \textbf{Spatiotemporal evidence construction.}  We annotate the shortest temporal interval sufficient to determine the answer and, for target-centric items, the key spatial region.  The interval is expanded only when the surrounding context is necessary to interpret the event or relation.
\item \textbf{Question and answer drafting.}  Writers formulate a question around the recorded visible evidence and specify one canonical gold answer, avoiding dependence on source-side context or video content that is not released.
\item \textbf{Candidate construction.}  We pair the gold answer with four plausible distractors: categorical items use visually similar classes; localization items use neighboring spatial regions or temporal positions; and action or temporal items use related actions, reversed orders, or confusable state changes.  An \emph{insufficient evidence} option is admitted only under the verified rule in Appendix~\ref{app:source-license}.
\item \textbf{Joint evidence and label correction.}  We jointly verify the question, gold answer, task/leaf labels, temporal boundary, spatial mark, and candidate set.  If the released video cannot uniquely support a field, the field is corrected when possible; otherwise, the record is discarded.
\item \textbf{Option permutation.}  After all semantic fields are frozen, we randomly permute the five answer positions and regenerate the answer key to avoid position bias.
\item \textbf{Independent review and adjudication.}  A second expert independently checks answerability, answer uniqueness, evidence sufficiency, and label consistency without seeing the initial annotator's answer.  Disagreements enter a third-expert adjudication round, and unresolved records are removed.
\item \textbf{Split and release audit.}  We enforce clip-level isolation across training, validation, and test partitions; remove duplicates, near-duplicates, and temporally overlapping material; validate the data schema, stable IDs, option permutations, source manifest, and deterministic evaluation scripts; and finally freeze the release version and checksums.
\end{enumerate}

\begin{table}[t]
\centering
\scriptsize
\setlength{\tabcolsep}{3pt}
\renewcommand{\arraystretch}{1.22}
\begin{tabular*}{\columnwidth}{@{\extracolsep{\fill}}lcll@{}}
\toprule
Split & QA & Role & Access \\
\midrule
Train & 5,651 & supervised training & permitted \\
Validation & 2,391 & selection/diagnosis & permitted \\
Test & 2,731 & fixed final evaluation & locked \\
\midrule
Total & 10,773 & complete release & --- \\
\bottomrule
\end{tabular*}
\caption{\textbf{Released split statistics for \rsvdata{}.}  The test partition is \bench{}; training and validation form RSVideo-Instruct.}
\label{tab:app-splits}
\end{table}

\noindent Figure~\ref{fig:app-pipeline} summarizes the same pipeline as an end-to-end data engine.  The upper flow tracks how public remote-sensing videos are registered, decoded, screened, and bound to task-specific spatiotemporal evidence.  The lower flow records the annotation, verification, adjudication, split isolation, and release-audit stages that convert the evidence into RSVideo-Instruct and \bench{} records.

The following subsections describe question construction, expert review, adjudication, quality control, and release auditing in greater detail.

\subsection{Leaf-guided Question, Answer, and Candidate Construction}
\label{app:leaf-guided-construction}
Question writers select a task and operational leaf before drafting an item.  They bind the item to a visible target, scene property, event, or relation and record the shortest temporal span and, when applicable, spatial region judged sufficient for a decision.  This evidence-constrained procedure prevents reliance on hidden source context or an unobserved continuation.

Candidate construction follows the evidence type.  Categorical distractors use visually similar categories; localization distractors use neighboring regions or temporal positions; action and temporal distractors use related actions, reversed order, or confusable state changes.  An auxiliary large model may flag ambiguous wording, duplicated options, or candidate--evidence mismatches, but it neither assigns the gold answer nor replaces human judgment.  The writer resolves every retained issue against the released frames and the recorded task/leaf constraint.

Each retained item has one verified correct option and four plausible distractors.  An \emph{insufficient evidence} option is admitted only when the absence of decisive visual support is itself the intended, independently verified answer; it is not a repair for corrupt or accidentally incomplete input.  After the gold option, evidence span, and task/leaf label pass verification, option positions are shuffled and the answer key is regenerated.

\subsection{Expert Annotation, Multi-round Review, and Adjudication}
\label{app:expert-review}
Three senior domain experts cover initial annotation, independent review, and adjudication.  Round 1 freezes a proposed question, five candidates, gold answer, evidence span/region, and task/leaf labels.  In Round 2, an independent reviewer inspects the released video without relying on the first annotator's answer and checks answerability, option uniqueness, evidence sufficiency, label consistency, and visual-mark consistency.  Any disagreement enters Round 3, in which an adjudicator either establishes one visually supported record or rejects the item.  Review stops only when the gold option is unique, the evidence boundary is sufficient, the task/leaf mapping is correct, and all distractors fail for documented visual reasons.  Automated VLM outputs remain auxiliary consistency signals and never replace the human decision.

\subsection{Quality Control, Label Correction, and Release Audit}
\label{app:quality-release}
Quality control operates at item, clip, split, and release levels.  Item-level checks cover decoding, answerability, evidence sufficiency, option uniqueness, task/leaf agreement, and mark consistency.  A detected label conflict triggers a joint re-check of question scope, evidence, gold option, task, and leaf: a field is corrected only when the released frames uniquely support the correction; otherwise the item is removed.  Clip- and split-level checks reject duplicates, near-duplicates, shared evidence clips, and overlapping temporal ranges across train, validation, and test.  Release-level checks validate option permutations, stable identifiers, the source manifest, and deterministic evaluation scripts.  Construction traces and adjudication notes remain separate from evaluation prompts so that they cannot act as privileged test information.

\section{Evaluation Protocol and Metrics}
\label{app:metrics}

\subsection{Video Input Protocol}
\label{app:video-input}
Every method receives the same released visual input, question, and five options for an item.  The final \bench{} evaluation uses a fixed video-decoding and frame-sampling procedure, spatial preprocessing policy, prompt template, and answer parser.  Methods operate without access to hidden construction records, gold evidence, answer keys, or test-partition annotations.  Reported comparisons use the same benchmark examples and one prediction file per run.

\subsection{Answer Parsing and Accuracy}
\label{app:answer-parsing}
The evaluator normalizes a prediction to one of \(\{A,B,C,D,E\}\).  It accepts an explicit final option label in a structured answer, or an unambiguous option label in the decoded response; responses without a uniquely parsable option are marked incorrect.  Let \(a_n\) and \(\hat a_n\) be the gold and parsed predictions for item \(n\).  Accuracy is
\begin{equation}
\mathrm{Acc}=\frac{1}{N}\sum_{n=1}^{N}\mathbb{I}[\hat a_n=a_n].
\end{equation}
The primary score uses all \(N=2{,}731\) items in \bench{}.

\subsection{Evidence Localization Metrics}
\label{app:evidence-localization}
Methods that emit evidence tags predict a temporal index \(\hat T\) and one or more spatial grid cells \(\hat R\).  Temporal Hit (T-Hit) is the fraction of items for which \(\hat T\) overlaps the annotated key frame or temporal window.  Region Hit (R-Hit) is the fraction for which \(\hat R\) overlaps the annotated target region.  These are grounding diagnostics, not substitutes for answer accuracy: a response may be correct while relying on incomplete evidence, or localized correctly but choose an incorrect option.

\subsection{Score Aggregation}
\label{app:score-aggregation}
In addition to overall accuracy, we report aggregates by capability level, capability group, task, leaf, and source.  Each aggregate is the item-weighted accuracy within its group.  Overall accuracy is always calculated directly over all items; it is never obtained by averaging group scores.  T-Hit and R-Hit use the same grouping rule when evidence annotations are available.  This protocol ensures that all reported views trace back to the same prediction file.

\subsection{Validity Audits}
\label{app:validity-audits}
The validity audit distinguishes genuine video use from answer priors by comparing text-only, random-single-frame, shuffled-frame, and full-video inputs while retaining the same checkpoint, prompt, option order, answer parser, and test examples.  Appendix~\ref{app:video-evidence-audit} reports the resulting capability-wise accuracies from the frozen prediction files.

\section{Additional Details of \method{}}
\label{app:method-details}

\subsection{Evidence Scores}
\label{app:evidence-scores}
The global saliency score is computed from the visual encoder self-attention. For layer \(l\) and head \(h\), let \(A^{l,h}\in\mathbb{R}^{L\times L}\) be the attention map normalized over key tokens. We average the attention received by token \(i\) over all query tokens, heads, and layers:
\begin{equation}
s_i^{\mathrm{sal}}
=
\frac{1}{N_l}
\sum_{l=1}^{N_l}
\frac{1}{L}
\sum_{m=1}^{L}
\frac{1}{H}
\sum_{h=1}^{H}
A_{m,i}^{l,h},
\label{eq:saliency-score}
\end{equation}
where \(N_l\) and \(H\) are the numbers of visual encoder layers and attention heads.

The question relevance and temporal change scores are
\begin{equation}
\begin{aligned}
s_i^{\mathrm{rel}}
&=
\operatorname{cos}
\left(
\mathbf{W}_v\mathbf{v}_i,\,
\mathbf{W}_q\mathbf{q}
\right),\\
s_i^{\mathrm{chg}}
&=
1-
\max_{j\in\mathcal{N}(i)}
\operatorname{cos}
\left(
\mathbf{W}_c\mathbf{v}_i,\,
\mathbf{W}_c\mathbf{v}_j
\right),
\end{aligned}
\label{eq:relevance-change-score}
\end{equation}
where \(\mathbf{W}_v\), \(\mathbf{W}_q\), and \(\mathbf{W}_c\) are learnable projections. The neighborhood \(\mathcal{N}(i)\) contains spatially adjacent tokens in the same frame and tokens at identical or nearby patch coordinates in adjacent sampled frames.

\paragraph{Time--Region Cell Prior.}
Each token has a temporal index \(\tau_i\in[T]\) and patch coordinate \(\mathbf{p}_i\). A fixed partition function \(\Gamma\) maps \(\mathbf{p}_i\) to one of \(G\) coarse spatial regions:
\begin{equation}
g_i=\Gamma(\mathbf{p}_i),
\qquad
\mathcal{C}_{\tau,g}
=
\left\{
i\in[L]
\mid
\tau_i=\tau,\ g_i=g
\right\}.
\label{eq:time-region-cell}
\end{equation}
For each cell, we estimate evidence strength from its most responsive tokens:
\begin{equation}
\begin{aligned}
m_{\tau,g}
&=
\max
\left(
1,\,
\left\lceil
\eta
\left|\mathcal{C}_{\tau,g}\right|
\right\rceil
\right),\\
c_{\tau,g}
&=
\operatorname{TopMean}_{m_{\tau,g}}
\left(
\left\{
u_i
\mid
i\in\mathcal{C}_{\tau,g}
\right\}
\right),
\end{aligned}
\label{eq:time-region-prior}
\end{equation}
where \(\eta\) is the token selection ratio. Appendix~\ref{app:eta-sensitivity} studies the impact of \(\eta\).

\subsection{Budget Allocator}
\label{app:budget-allocator}
The allocator first summarizes the score distribution as
\begin{equation}
\mathbf{h}_{\ell}
=
\left[
\operatorname{Mean}(\boldsymbol{\ell});
\operatorname{Std}(\boldsymbol{\ell});
\operatorname{Max}(\boldsymbol{\ell});
\operatorname{TopMean}(\boldsymbol{\ell})
\right],
\label{eq:score-summary}
\end{equation}
where \(\operatorname{TopMean}\) averages the top 10\% scores. It then predicts the Beta parameters for \(\gamma\):
\begin{equation}
\begin{aligned}
(a_{\gamma},b_{\gamma})
&=
\operatorname{Softplus}
\left(
\operatorname{MLP}
\left(
[\mathbf{q};\mathbf{h}_{\ell}]
\right)
\right)
+
\epsilon,\\
\gamma
&\sim
\operatorname{Beta}
(a_{\gamma},b_{\gamma}).
\end{aligned}
\label{eq:visual-budget-beta}
\end{equation}
At inference time, \(\gamma=a_{\gamma}/(a_{\gamma}+b_{\gamma})\), yielding deterministic token allocation.

\subsection{Evidence-Aware SFT}
\label{app:evidence-aware-sft}
In the first training stage, sparsification is disabled and the model receives the complete visual sequence \(\mathbf{H}\). Given the target evidence-answer string \(Y^{*}\), E-SFT optimizes
\begin{equation}
\mathcal{L}_{\mathrm{eSFT}}
=
-
\sum_{t}
\log p_{\theta}
\left(
Y^{*}_{t}
\mid
Y^{*}_{<t},\mathbf{H},Q
\right),
\label{eq:esft-loss}
\end{equation}
which teaches the model to follow the evidence format before RL.

\subsection{Cost Penalty}
\label{app:cost-penalty}
Let \(\mathcal D(E)\) be the set of available evidence dimensions (temporal and/or regional).  The breadth cost averages only those available dimensions,
\begin{equation}
\begin{aligned}
C_{\mathrm{cost}}
&=\frac{1}{|\mathcal D(E)|}\Biggl[
\mathbb I[T\in\mathcal D(E)]
\frac{|\widehat{\mathcal T}(E)|}{T}
\\
&\qquad+
\mathbb I[G\in\mathcal D(E)]
\frac{|\widehat{\mathcal G}(E)|}{G}
\Biggr].
\end{aligned}
\label{eq:app-evidence-cost}
\end{equation}
When both dimensions are available, this reduces to the one-half average used in the main formulation.

\subsection{Group Relative Policy Optimization}
\label{app:grpo}
For GRPO, we sample a group of responses for each question and compute relative advantages from their evidence-aware returns. Let \(\mathcal{R}_{k}\) be the return of the \(k\)-th response in a group. We use
\begin{equation}
\widehat{A}_{k}
=
\frac{\mathcal{R}_{k}-\operatorname{Mean}(\{\mathcal{R}_{j}\})}
{\operatorname{Std}(\{\mathcal{R}_{j}\})+\epsilon},
\label{eq:grpo-advantage}
\end{equation}
and optimize the clipped policy objective with a KL penalty to the E-SFT policy. This keeps RL updates focused on improving evidence alignment and sparse background routing without drifting from the supervised evidence format.

\subsection{Evidence Mapping, Tag Parsing, and Run Records}
\label{app:evidence-parser}
For a sampled token grid, define the annotation-availability tests
\begin{equation}
\begin{aligned}
c_{\mathrm T}(i)&=
\begin{cases}
\mathbb I[\tau_i\in\mathcal T^*],&\text{if temporal labels exist},\\
1,&\text{otherwise},
\end{cases}\\
c_{\mathrm G}(i)&=
\begin{cases}
\mathbb I[g_i\in\mathcal G^*],&\text{if regional labels exist},\\
1,&\text{otherwise}.
\end{cases}
\end{aligned}
\end{equation}
The annotation-to-token map is then
\begin{equation}
\mathcal I^*=\{i\in[L]\mid c_{\mathrm T}(i)c_{\mathrm G}(i)=1\}.
\label{eq:app-annotation-token-mapping}
\end{equation}
Thus, when only one annotation type is present, only its corresponding test restricts the target set.

The canonical output has the form
\[
\begin{aligned}
E &::= \texttt{Evidence: }\mathcal T_p\texttt{-}\mathcal G_p
      \texttt{; Answer: }[A\texttt{-}E],\\
\mathcal T_p &::= \texttt{T}d(\texttt{/T}d)^*,\\
\mathcal G_p &::= \texttt{R}d(\texttt{/R}d)^*,
\end{aligned}
\]
where \(\mathcal T_p\) and \(\mathcal G_p\) are nonempty lists of temporal and regional indices, respectively,
where \(d\) denotes a zero-padded nonnegative integer.  A deterministic parser extracts the sets, removes duplicates, and checks each index against the sampled-frame count and grid size stored in the run manifest.  A trajectory is valid only if it contains exactly one final answer label and every supplied evidence index is in range; malformed or conflicting answer labels are invalid.  For items with only temporal or only regional supervision, the unavailable field is omitted and the available field alone determines evidence validity.  T-Hit/R-Hit are nonempty-overlap diagnostics, whereas \(R_{\mathrm T}\) and \(R_{\mathrm G}\) above are annotation-coverage fractions.

The four reward weights, overlap conventions, parser version, sampled-frame indices, and grid layout are frozen in the run manifest.  The sensitivity report in Appendix~\ref{app:reward-weight} varies these weights around the default configuration while preserving the main-paper optimization protocol.

\subsection{Training Configuration, Video Preprocessing, and Reproducibility}
\label{app:training-preprocessing}
Table~\ref{tab:app-training-config} reports the numerical settings used in the experiments.  Training follows E-SFT on the complete visual sequence and then GRPO with sparse evidence focusing and background compression enabled.

\begin{table}[t]
\centering
\small
\setlength{\tabcolsep}{4pt}
\renewcommand{\arraystretch}{1.08}
\begin{tabularx}{\columnwidth}{P{0.46\columnwidth}Y}
\toprule
Setting & Value \\
\midrule
Compute & \(8\times\) NVIDIA A100, 80\,GB each \\
Adaptation / precision & LoRA / BF16 \\
Maximum sequence length & 2,048 \\
Memory control & gradient checkpointing \\
Per-device batch size & 1 \\
Gradient accumulation & 8 \\
Learning-rate schedule & cosine \\
Base learning rate & \(1\times10^{-5}\) \\
Warmup ratio / weight decay & 0.03 / 0.01 \\
LoRA rank / alpha / dropout & 32 / 64 / 0.05 \\
Time--region cell pooling ratio & \(\eta=0.10\) \\
Default visual-token budget & \(\rho=0.40\) \\
\bottomrule
\end{tabularx}
\caption{Optimization settings used in the reported experiments.}
\label{tab:app-training-config}
\end{table}

Every final run additionally freezes the decoder and version, sampled-frame count and exact indices, temporal stride, spatial resize/crop policy, backbone-native processor and version, token-grid layout, prompt template, evidence/answer parser version, random seed, checkpoint identifier, and prediction-file checksum.  The released clip, question, and options are identical across methods; a backbone-native visual processor may differ, but all such differences are recorded with the prediction file.

\section{Additional Diagnostic and Robustness Analyses}
\label{app:additional_analyses}

\subsection{Complete Benchmark Results of Existing VLMs}
\label{app:experiment-scope}
The core results include the 26-backbone in-domain comparison, progressive reward ablation, cross-dataset transfer evaluation, and qualitative evidence-selection comparison. This section provides complementary analyses covering benchmark difficulty, validity of video use, visual-token budget, inference efficiency, sparse-token integration, RL hyperparameters, and reward weights. Unless stated otherwise, experiments that use sparse selection use Token Keep 40.0.

To establish the difficulty of \bench{} before evaluating the proposed training method, we compare general-purpose, remote-sensing-specialized, and proprietary/API VLMs.  Table~\ref{tab:app-complete-existing-vlms} reports all models under the same five-choice benchmark interface, item set, option order, prompt construction, and answer parser; model-native visual processors are retained where required and documented in the run records.  Overall accuracy is computed over all 2,731 test instances, and group accuracies follow the seven capability groups defined by the taxonomy.  The comparison covers representative MiniCPM, LLaVA, Qwen, VideoLLaMA, and InternVL model families~\citep{yu2025minicpmv45,li2024llavaonevision,bai2025qwen25vl,qwen2025qwen3vl,qwen2026qwen36,zhang2025videollama3,wang2025internvl35}.

\begin{table*}[t]
\centering
\scriptsize
\setlength{\tabcolsep}{3.2pt}
\renewcommand{\arraystretch}{1.02}
\begin{tabular*}{\textwidth}{@{\extracolsep{\fill}}lcccccccc@{}}
\toprule
Model & Overall $\uparrow$ & SP $\uparrow$ & TP $\uparrow$ & AP $\uparrow$ & CER $\uparrow$ & SER $\uparrow$ & SLR $\uparrow$ & SCR $\uparrow$ \\
\midrule
\rowcolor{black!6}
\multicolumn{9}{l}{\textit{Open-source general VLMs}} \\
MiniCPM-V-4.5 & 22.78 & 42.13 & 20.27 & 23.17 & 36.36 & 18.64 & 21.77 & 14.93 \\
LLaVA-OneVision-7B & 23.62 & 44.17 & 21.53 & 23.87 & 37.83 & 19.37 & 22.54 & 15.18 \\
Qwen2.5-VL-7B-Instruct & 24.88 & 48.12 & 23.86 & 25.38 & 43.26 & 22.11 & 24.23 & 13.28 \\
VideoLLaMA3-7B & 24.94 & 45.63 & 22.37 & 25.14 & 39.26 & 20.47 & 24.23 & 16.34 \\
InternVL3.5-8B & 25.87 & 52.20 & 25.21 & 26.24 & 47.99 & 25.91 & 18.47 & 15.28 \\
Qwen3-VL-8B-Instruct & 29.74 & 59.75 & 22.66 & 31.91 & 52.35 & 20.23 & 32.13 & 16.60 \\
InternVL3.5-14B & 28.36 & 55.10 & 27.42 & 28.31 & 48.62 & 23.84 & 27.46 & 17.92 \\
LLaVA-OneVision-72B & 28.74 & 53.96 & 27.88 & 28.91 & 47.34 & 24.16 & 27.72 & 18.38 \\
InternVL3-14B & 29.12 & 55.42 & 28.06 & 29.28 & 48.91 & 24.52 & 28.34 & 18.70 \\
LLaVA-Video-72B (Qwen2) & 29.95 & 54.86 & 28.41 & 30.52 & 48.33 & 25.28 & 29.14 & 18.76 \\
GLM-4.6V-Flash & 30.48 & 57.82 & 28.74 & 30.15 & 49.13 & 24.91 & 29.82 & 18.62 \\
Qwen2.5-VL-32B & 30.92 & 57.21 & 29.34 & 31.12 & 50.06 & 26.18 & 30.11 & 19.34 \\
InternVL3.5-38B & 31.08 & 58.04 & 29.72 & 31.35 & 50.42 & 26.41 & 30.28 & 19.46 \\
Qwen2.5-VL-72B & 31.36 & 57.76 & 29.63 & 31.57 & 49.84 & 26.57 & 30.43 & 19.67 \\
InternVL3-38B & 31.54 & 58.66 & 30.11 & 31.82 & 51.06 & 26.72 & 30.74 & 19.82 \\
Kimi-VL-16B-A3B-Instruct & 32.09 & 58.44 & 30.37 & 31.63 & 51.19 & 26.82 & 31.46 & 20.45 \\
Qwen3-VL-32B & 32.74 & 60.42 & 31.26 & 33.12 & 52.34 & 27.52 & 32.28 & 20.62 \\
InternVL3-78B & 33.02 & 60.96 & 31.84 & 33.41 & 52.86 & 27.83 & 32.72 & 20.89 \\
InternVL3.5-241B-A28B-Instruct & 33.14 & 61.62 & 32.14 & 33.86 & 53.28 & 27.64 & 33.12 & 19.81 \\
Qwen3-VL-235B-A22B-Instruct & 33.62 & 61.45 & 31.84 & 34.16 & 52.77 & 27.86 & 33.48 & 21.11 \\
Gemma-4-31B & 34.41 & 62.73 & 33.57 & 34.81 & 54.22 & 28.79 & 34.17 & 21.43 \\
Qwen3.6-27B-Instruct & 35.30 & 63.71 & 33.44 & 35.26 & 54.83 & 28.49 & 35.62 & 22.78 \\
\midrule
\rowcolor{black!6}
\multicolumn{9}{l}{\textit{Open-source domain-specialized VLMs}} \\
LLaVA1.5-UAV & 28.11 & 54.83 & 28.64 & 26.37 & 45.92 & 21.84 & 24.76 & 18.63 \\
GeoChat-UAV & 28.99 & 58.21 & 33.17 & 27.24 & 50.46 & 21.37 & 24.91 & 18.42 \\
SIS-Motion-7B & 32.42 & 58.76 & 30.82 & 31.69 & 49.71 & 28.46 & 29.57 & 21.88 \\
\midrule
\rowcolor{black!6}
\multicolumn{9}{l}{\textit{Proprietary/API VLMs}} \\
Qwen3-VL-Max & 37.62 & 65.02 & 37.12 & 37.64 & 55.86 & 31.28 & 36.96 & 25.96 \\
Kimi-K2.5 & 39.27 & 66.74 & 39.05 & 39.86 & 57.68 & 33.42 & 38.74 & 27.10 \\
GPT-5 & 39.95 & 67.58 & 39.71 & 40.84 & 58.72 & 34.36 & 39.32 & 27.55 \\
Gemini-3-Pro & 40.34 & 67.91 & 40.08 & 41.23 & 59.14 & 34.88 & 39.72 & 27.89 \\
\bottomrule
\end{tabular*}
\caption{\textbf{Complete \bench{} results for existing VLMs. } Values are accuracies (\%).  SP, TP, AP, CER, SER, SLR, and SCR denote Scene Perception, Target Perception, Action Perception, Complex-Environment Reasoning, Spatiotemporal Evolution Reasoning, Spatiotemporal Localization Reasoning, and Spatiotemporal Consistency Reasoning, respectively.  Overall and group accuracies are computed from the same frozen prediction file for each model under the protocol in Appendix~\ref{app:metrics}.}
\label{tab:app-complete-existing-vlms}
\end{table*}

The strongest evaluated open-source, domain-specialized, and proprietary/API models reach 35.30\%, 32.42\%, and 40.34\% overall accuracy, respectively.  Across the evaluated systems, Scene Perception has the highest group accuracy, whereas Spatiotemporal Consistency Reasoning remains substantially lower.  This pattern is consistent with the benchmark placing greater demands on temporal and cross-frame relation reasoning than on scene recognition, although the aggregate results do not isolate which visual or architectural factors cause the gap.

\subsection{Validity Audit: Dependence on Video Evidence}
\label{app:video-evidence-audit}
To test whether performance depends on video evidence rather than only question--option priors or static appearance, we fix a Qwen3.6-27B-Instruct checkpoint fine-tuned on RSVideo-Instruct with full-video inputs and answer supervision, and change only its inference-time input.  Table~\ref{tab:app-video-evidence-audit} reports the capability-wise results under the fourinference-time input conditions. Text only removes all visual evidence; Random single frame retains static appearance and partial layout; Shuffled frames retains multi-frame content but destroys temporal order; Full video retains both content and order.  All four conditions use the same model parameters, questions, option order, prompt, parser, and test set.

\begin{table*}[t]
\centering
\scriptsize
\setlength{\tabcolsep}{3.0pt}
\renewcommand{\arraystretch}{1.05}
\begin{tabular*}{\textwidth}{@{\extracolsep{\fill}}llcccccccc@{}}
\toprule
Input Setting & Retained Evidence & Overall $\uparrow$ & SP $\uparrow$ & TP $\uparrow$ & AP $\uparrow$ & CER $\uparrow$ & SER $\uparrow$ & SLR $\uparrow$ & SCR $\uparrow$ \\
\midrule
Text only & Question and answer options & 23.49 & 28.61 & 24.27 & 22.83 & 25.94 & 23.47 & 23.29 & 20.84 \\
Random single frame & Static appearance and partial layout & 33.17 & 63.47 & 40.40 & 27.03 & 52.11 & 25.22 & 29.84 & 22.20 \\
Shuffled frames & Multi-frame content without correct order & 35.13 & 63.14 & 40.02 & 32.52 & 51.75 & 28.19 & 33.10 & 24.12 \\
Full video & Visual content and temporal order & 36.59 & 62.89 & 39.69 & 35.68 & 51.32 & 30.32 & 35.67 & 25.66 \\
\bottomrule
\end{tabular*}
\caption{Validity audit by inference-time input degradation.  All values are accuracies (\%), and higher is better ($\uparrow$).  The checkpoint and evaluation protocol remain fixed across rows; only the available visual and temporal evidence changes.}
\label{tab:app-video-evidence-audit}
\end{table*}

Text-only input obtains 23.49\%, 3.49 points above the 20.00\% chance level, indicating limited but nonzero predictive signal from the question and options.  A random frame raises Overall accuracy to 33.17\% and slightly exceeds Full video on SP, TP, and CER, indicating that static appearance suffices for many items in those groups.  Its AP, SER, SLR, and SCR scores are respectively 8.65, 5.10, 5.83, and 3.46 points below Full video.  Shuffled frames reach 35.13\%, 1.46 points below ordered Full video, whereas Full video attains the highest Overall, AP, SER, SLR, and SCR scores.  Together, these results are consistent with useful contributions from visual content and temporal order, particularly in the four temporally demanding capability groups, while static appearance remains informative for SP, TP, and CER.

\subsection{Time--Region Cell Pooling Ratio}
\label{app:eta-sensitivity}
To examine how local evidence is aggregated within each time--region cell, we vary the token selection ratio \(\eta\) while fixing the visual-token budget at \(\rho=0.40\), together with the backbone, training data, checkpoint-selection rule, preprocessing, rollout budget, and evaluator.  A smaller \(\eta\) emphasizes the most responsive tokens in each cell, whereas a larger \(\eta\) incorporates broader local context but may dilute sparse evidence with background responses.  Table~\ref{tab:app-eta-sensitivity} compares \(\eta\in\{0.05,0.10,0.20\}\) under the same Qwen3.6-27B protocol with the default configuration.

\begin{table}[t]
\centering
\small
\setlength{\tabcolsep}{4pt}
\renewcommand{\arraystretch}{1.05}
\begin{tabular*}{0.96\columnwidth}{@{\extracolsep{\fill}}lcccc@{}}
\toprule
Pooling Condition & \(\eta\) & Accuracy $\uparrow$ & TH $\uparrow$ & RH $\uparrow$ \\
\midrule
Narrow pooling  & 0.05 & 39.91 & 56.6 & 53.7 \\
Selected default & 0.10 & 40.63 & 57.4 & 54.8 \\
Broad pooling   & 0.20 & 40.14 & 56.9 & 54.2 \\
\bottomrule
\end{tabular*}
\caption{Sensitivity to the time--region cell pooling ratio under the fixed Qwen3.6-27B protocol.  \(\eta\) controls the proportion of highest-scoring tokens aggregated within each time--region cell; Accuracy, TH, and RH are percentages, where higher is better.}
\label{tab:app-eta-sensitivity}
\end{table}

The results show that the cell-level pooling ratio controls the balance between concentrated evidence and surrounding context.  Narrow pooling at \(\eta=0.05\) relies on very few highly responsive tokens and is therefore more sensitive to isolated or noisy responses.  Broad pooling at \(\eta=0.20\) incorporates more tokens from each cell but can dilute sparse target evidence with background responses.  The selected setting, \(\eta=0.10\), achieves the highest Accuracy, TH, and RH, indicating that aggregating the top 10\% of tokens provides a suitable balance between evidence concentration and local contextual support.

\subsection{Visual-token Budget and Background Context}
\label{app:budget-sensitivity}
To quantify the accuracy--evidence trade-off, we vary the retained local-evidence budget while fixing the backbone, training data, checkpoint-selection rule, preprocessing, rollout budget, and evaluator.  Table~\ref{tab:app-budget-sensitivity} compares the selected ratio \(\rho=0.40\) with lower and higher budgets under the same Qwen3.6-27B protocol.  All configuration and hyperparameter decisions are made on the RSVideo-Instruct validation split; after the configuration is fixed, \bench{} is evaluated once.

\begin{table}[t]
\centering
\small
\setlength{\tabcolsep}{4pt}
\renewcommand{\arraystretch}{1.05}
\begin{tabular*}{0.96\columnwidth}{@{\extracolsep{\fill}}lcccc@{}}
\toprule
Budget Condition & \(\rho\) & Accuracy $\uparrow$ & TH $\uparrow$ & RH $\uparrow$ \\
\midrule
Lower-budget run & 0.30 & 39.72 & 56.1 & 53.2 \\
Reported default & 0.40 & 40.63 & 57.4 & 54.8 \\
Higher-budget run & 0.50 & 40.18 & 57.0 & 54.1 \\
\bottomrule
\end{tabular*}
\caption{Budget sensitivity under the fixed Qwen3.6-27B protocol.  \(\rho\) denotes the retained visual-token budget ratio; Accuracy, TH, and RH are percentages, where higher is better.}
\label{tab:app-budget-sensitivity}
\end{table}

Among the three tested ratios, \(\rho=0.40\) gives the highest accuracy, TH, and RH, reaching 40.63\%, 57.4\%, and 54.8\%, respectively.  Reducing \(\rho\) to 0.30 lowers accuracy by 0.91 points and reduces both grounding metrics, whereas increasing \(\rho\) to 0.50 lowers accuracy by 0.45 points.  Thus, \(\rho=0.40\) gives the best observed trade-off among the tested ratios.

\subsection{Inference Efficiency}
\label{app:inference-efficiency}
To quantify the inference benefit of sparse visual-token selection, we profile Dense E-SFT and \method{} on the full \bench{} test split of 2,731 questions.  Both systems use the Qwen3.6-27B backbone and the same NVIDIA A100 80\,GB GPU, BF16 inference, batch size 1, deterministic decoding, prompt, and answer parser.  Each video is represented by 12 uniformly sampled temporal positions.  Peak memory is the maximum allocated GPU memory, and latency is the mean end-to-end time per question over the full split after 20 warm-up examples. Table~\ref{tab:app-inference-efficiency} reports the resulting peak memory,
latency, and test accuracy.

\begin{table}[t]
\centering
\scriptsize
\setlength{\tabcolsep}{3pt}
\renewcommand{\arraystretch}{1.05}
\begin{tabular*}{\columnwidth}{@{\extracolsep{\fill}}lccc@{}}
\toprule
Setting & Memory (GB) $\downarrow$ & Latency (s) $\downarrow$ & Acc. $\uparrow$ \\
\midrule
Dense E-SFT & 72.6 & 5.27 & 36.78 \\
\method{} & 67.1 & 3.79 & 40.63 \\
\bottomrule
\end{tabular*}
\caption{Inference efficiency of Qwen3.6-27B on \bench{} under a fixed single-GPU protocol.  Memory denotes peak allocated GPU memory, Latency is the mean end-to-end time per question, and Acc. is test accuracy (\%); lower is better for memory and latency, whereas higher is better for accuracy.}
\label{tab:app-inference-efficiency}
\end{table}

On \bench{}, \method{} uses 5.5\,GB less peak memory (7.6\%) and 1.48\,s less mean latency per question (28.1\%) than dense E-SFT, while improving accuracy by 3.85 points.  The smaller memory change relative to the visual-token reduction is expected because model parameters and non-visual states remain resident, whereas the shorter visual sequence directly reduces multimodal prefilling and attention computation.  Because the rows represent two complete systems, the accuracy difference is not attributed to sparsification alone.

\subsection{Sparse-token Integration}
\label{app:sparse-integration}
To identify where selection is most effective and how selected evidence is best fused, we compare five integration variants at the same 40\% token budget.  Table~\ref{tab:app-sparse-integration} holds the Qwen3.6-27B backbone and training protocol fixed while changing only selection position and fusion strategy.

\begin{table}[t]
\centering
\fontsize{6.5pt}{7.2pt}\selectfont
\setlength{\tabcolsep}{1.8pt}
\renewcommand{\arraystretch}{1.05}
\begin{tabular*}{\columnwidth}{@{\extracolsep{\fill}}P{0.30\columnwidth}C{0.24\columnwidth}C{0.22\columnwidth}c@{}}
\toprule
Variant & Selection Position & Fusion Strategy & Accuracy $\uparrow$ \\
\midrule
Pre-projector pruning & Before projector & Concatenation & 38.42 \\
Post-projector concat & After projector & Concatenation & 39.18 \\
Post-projector gated sum & After projector & Gated sum & 39.71 \\
Post-projector cross-attention & After projector & Cross-attention & 40.63 \\
LLM-layer pruning & LLM layers & Token pruning & 38.96 \\
\bottomrule
\end{tabular*}
\caption{Effect of sparse-token integration position and fusion strategy under the fixed Qwen3.6-27B protocol.  Higher accuracy is better.}
\label{tab:app-sparse-integration}
\end{table}

Post-projector cross-attention achieves the best observed accuracy, 40.63\%, exceeding pre-projector pruning, post-projector concatenation, post-projector gated summation, and LLM-layer pruning by 2.21, 1.45, 0.92, and 1.67 points, respectively.  These results support the post-projector cross-attention configuration among the tested variants.

\subsection{RL Optimization Sensitivity}
\label{app:rl-sensitivity}
To test whether the reported gain depends on a narrow RL configuration, we vary the differentiable selection temperature, GRPO group size, and KL coefficient separately.  Each control fixes all other settings.  Tables~\ref{tab:app-gumbel-temperature}, \ref{tab:app-group-size}, and \ref{tab:app-kl-coefficient} isolate exploration smoothness, group-relative estimation, and regularization against the evidence-aware SFT reference, respectively.

\begin{table}[t]
\centering
\small
\setlength{\tabcolsep}{5pt}
\renewcommand{\arraystretch}{1.05}
\begin{tabular*}{0.96\columnwidth}{@{\extracolsep{\fill}}cccc@{}}
\toprule
Temperature & Accuracy $\uparrow$ & TH $\uparrow$ & RH $\uparrow$ \\
\midrule
0.3 & 39.46 & 55.8 & 53.1 \\
0.5 & 40.12 & 56.7 & 54.0 \\
0.7 & 40.63 & 57.4 & 54.8 \\
1.0 & 40.21 & 56.9 & 54.2 \\
1.5 & 39.37 & 55.6 & 52.9 \\
\bottomrule
\end{tabular*}
\caption{\textbf{Effect of Gumbel-Sigmoid temperature.}  Accuracy, TH, and RH are reported in \%; higher values are better.}
\label{tab:app-gumbel-temperature}
\end{table}

Temperature 0.7 gives the best observed accuracy, TH, and RH.  Temperatures 0.5 and 1.0 remain within 0.51 points of its accuracy, whereas 0.3 and 1.5 are 1.17 and 1.26 points lower.  The tested settings therefore favor an intermediate temperature under this protocol.

\begin{table}[t]
\centering
\scriptsize
\setlength{\tabcolsep}{2pt}
\renewcommand{\arraystretch}{1.05}
\begin{tabular*}{\columnwidth}{@{\extracolsep{\fill}}cccccc@{}}
\toprule
Group Size \(G\) & Accuracy $\uparrow$ & TH $\uparrow$ & RH $\uparrow$ & Format Valid $\uparrow$ & Relative Cost $\downarrow$ \\
\midrule
2 & 39.82 & 56.3 & 53.6 & 98.7 & 0.50 \\
4 & 40.63 & 57.4 & 54.8 & 99.1 & 1.00 \\
6 & 40.41 & 57.1 & 54.4 & 99.0 & 1.48 \\
8 & 40.28 & 56.8 & 54.2 & 98.9 & 1.96 \\
\bottomrule
\end{tabular*}
\caption{\textbf{Effect of GRPO group size \(G\).}  Accuracy, TH, RH, and Format Valid are percentages, where higher is better; Relative Cost is normalized to the default group size \(G=4\), and lower is better.}
\label{tab:app-group-size}
\end{table}

Group size \(G=4\) gives the highest observed accuracy and format validity at the reference training cost.  \(G=2\) halves the reported relative cost but lowers accuracy by 0.81 points, whereas \(G=6\) and \(G=8\) increase relative cost without improving accuracy.  Thus, \(G=4\) provides the best observed accuracy--cost trade-off among the tested settings.

\begin{table}[t]
\centering
\scriptsize
\setlength{\tabcolsep}{3pt}
\renewcommand{\arraystretch}{1.05}
\begin{tabular*}{0.96\columnwidth}{@{\extracolsep{\fill}}cccc@{}}
\toprule
KL Coefficient & Accuracy $\uparrow$ & Format Valid $\uparrow$ & Evidence Length $\downarrow$ \\
\midrule
0.00 & 39.74 & 96.8 & 2.84 \\
0.01 & 40.21 & 98.5 & 2.61 \\
0.03 & 40.63 & 99.1 & 2.43 \\
0.05 & 40.36 & 99.2 & 2.31 \\
0.10 & 39.58 & 99.4 & 2.06 \\
\bottomrule
\end{tabular*}
\caption{\textbf{Effect of KL regularization against the evidence-aware SFT reference.}  Accuracy and Format Valid are reported in \%; higher is better ($\uparrow$).  Evidence Length denotes the average number of evidence units, for which lower indicates more compact evidence ($\downarrow$).}
\label{tab:app-kl-coefficient}
\end{table}

Without KL regularization, format validity drops to 96.8\% and accuracy reaches 39.74\%.  A coefficient of 0.03 achieves the best accuracy, 40.63\%, with 99.1\% valid outputs.  Larger coefficients produce slightly higher format validity and shorter evidence outputs but lower accuracy, showing that excessive regularization restricts evidence generation.

\subsection{Reward-weight Sensitivity}
\label{app:reward-weight}
To test whether the full reward depends on a single fragile allocation, we keep every reward component active and vary the relative weights of answer correctness, spatiotemporal evidence alignment, background compression, and evidence-breadth cost.  Table~\ref{tab:app-reward-weight} uses the same initialization, data, token budget, and RL protocol for every row.

\begin{table}[t]
\centering
\scriptsize
\setlength{\tabcolsep}{1.8pt}
\renewcommand{\arraystretch}{1.05}
\begin{tabular*}{\columnwidth}{@{\extracolsep{\fill}}ccccccc@{}}
\toprule
\(\lambda_{\mathrm{ans}}\) & \(\lambda_{\mathrm{st}}\) & \(\lambda_{\mathrm{bg}}\) & \(\lambda_{\mathrm{cost}}\) & Accuracy $\uparrow$ & TH $\uparrow$ & RH $\uparrow$ \\
\midrule
0.55 & 0.25 & 0.15 & 0.05 & 39.88 & 55.9 & 53.7 \\
0.40 & 0.40 & 0.15 & 0.05 & 40.63 & 57.4 & 54.8 \\
0.30 & 0.50 & 0.15 & 0.05 & 40.28 & 57.8 & 54.6 \\
0.40 & 0.30 & 0.25 & 0.05 & 40.07 & 56.5 & 55.2 \\
0.40 & 0.35 & 0.15 & 0.10 & 39.71 & 56.2 & 53.9 \\
\bottomrule
\end{tabular*}
\caption{\textbf{Effect of reward-weight allocation.}  \(\lambda_{\mathrm{ans}}\), \(\lambda_{\mathrm{st}}\), \(\lambda_{\mathrm{bg}}\), and \(\lambda_{\mathrm{cost}}\) denote answer correctness, spatiotemporal evidence alignment, background compression, and evidence-breadth cost weights; Accuracy, TH, and RH are reported in \%, where higher is better.}
\label{tab:app-reward-weight}
\end{table}

The \(0.40/0.40/0.15/0.05\) allocation achieves the best overall accuracy, 40.63\%.  Assigning more weight to spatiotemporal alignment increases TH from 57.4\% to 57.8\% but lowers accuracy by 0.35 points, whereas assigning more weight to background compression increases RH from 54.8\% to 55.2\% but lowers accuracy by 0.56 points.  Increasing the evidence-breadth cost weight reduces accuracy to 39.71\%.  These results show that the default allocation provides the best balance among answer correctness, evidence grounding, and compact evidence selection.
\section{Leaf-complete Qualitative Examples}
\label{app:qualitative_examples}
\label{app:qualitative-leaf-complete}
We provide one qualitative evidence card for each of the 26 active leaf capabilities in Table~\ref{tab:app-leaves}.  Consistent with the taxonomy described above, each leaf specifies a distinct observable evidence pattern rather than a new dataset source or a new evaluation metric.  Each card pairs an anonymized frame strip or public mark with the corresponding question, answer options, and visible evidence scope.  The examples illustrate what the leaf is designed to test, rather than reporting additional model results.

\small
\paragraph{Leaf 01: Scene Type.}

Leaf 01 follows the path \emph{L1 Perception} \(\rightarrow\) \emph{G1 Scene Perception} \(\rightarrow\) \emph{T1 Scene Type}. It belongs to perception because the decision rests on directly observable global scene appearance rather than an inferred event or target trajectory.

This leaf tests whether a model can integrate whole-frame context across the clip and select the scene category best supported by that evidence. Local objects may provide useful cues, but the predicted label must characterize the scene as a whole.

\begin{figure*}[tbp]
\centering
\includegraphics[width=\textwidth]{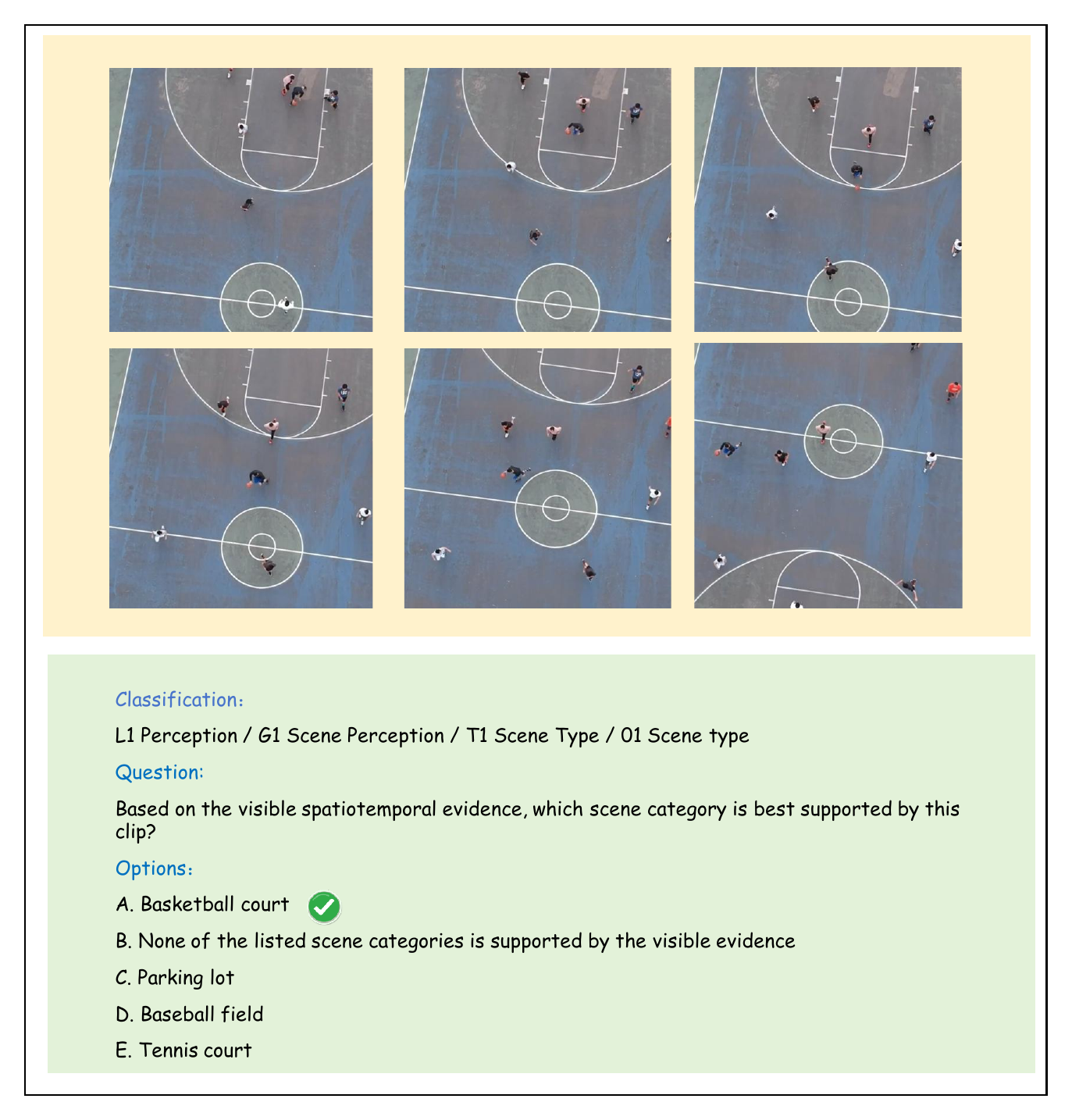}
\caption{Qualitative example for Leaf 01, scene type.}
\label{fig:app-leaf-01}
\end{figure*}

As shown in Fig.~\ref{fig:app-leaf-01}, the example asks which scene category is supported by the clip and contrasts a basketball court, parking lot, baseball field, tennis court, and an evidence-insufficient alternative.

\paragraph{Leaf 02: Viewpoint Type.}

Leaf 02 follows the path \emph{L1 Perception} \(\rightarrow\) \emph{G1 Scene Perception} \(\rightarrow\) \emph{T2 Viewpoint Type}. It is assigned to scene perception because camera elevation and orientation can be determined from the overall projection geometry of the observed scene.

This leaf assesses whether a model can distinguish viewpoints using perspective, visible object surfaces, and horizon-related cues. The judgment concerns the imaging configuration rather than the identity or action of any individual target.

\begin{figure*}[tbp]
\centering
\includegraphics[width=\textwidth]{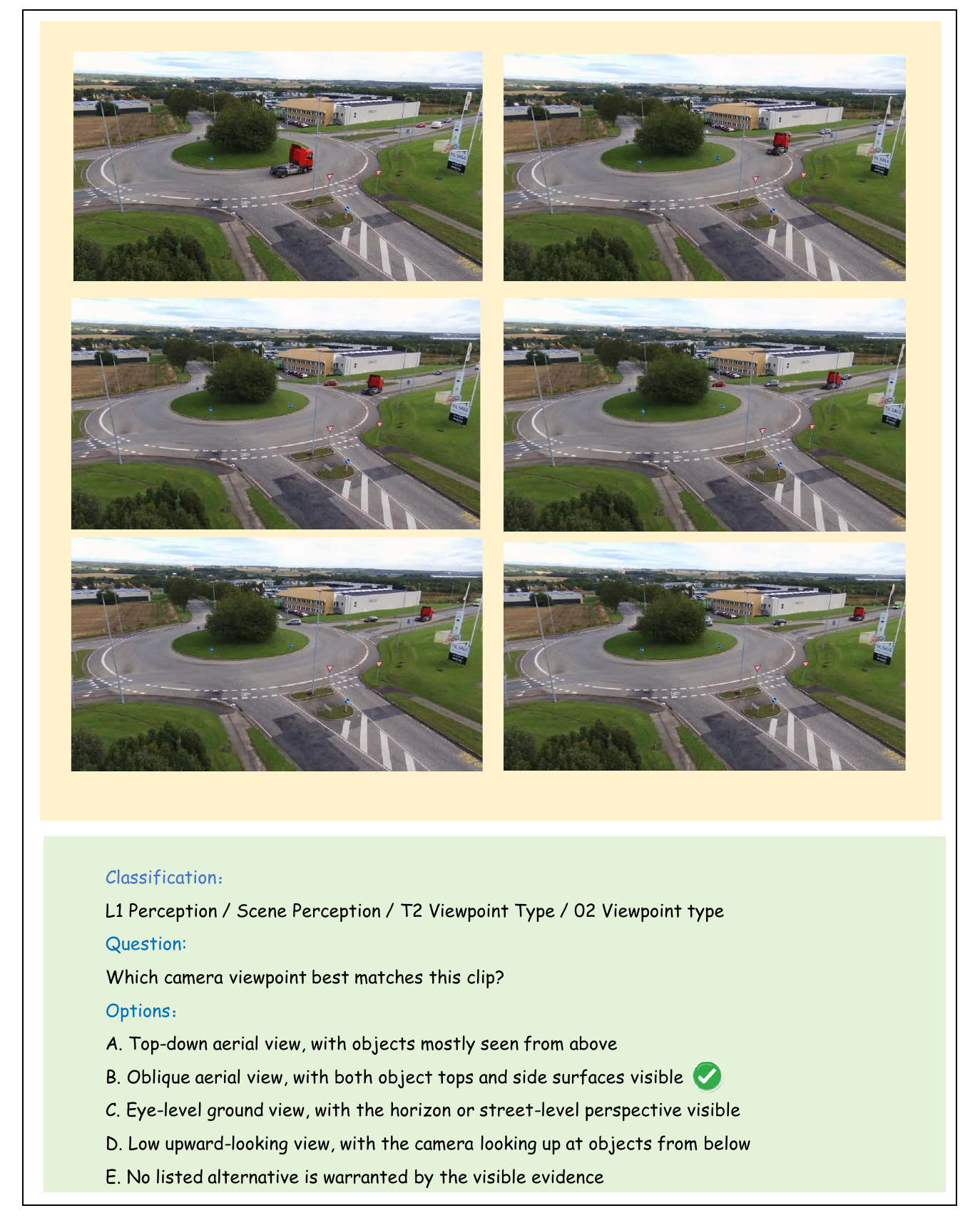}
\caption{Qualitative example for Leaf 02, viewpoint type.}
\label{fig:app-leaf-02}
\end{figure*}

As shown in Fig.~\ref{fig:app-leaf-02}, the example asks the model to choose among top-down aerial, oblique aerial, eye-level ground, low upward-looking, and unsupported viewpoints.

\paragraph{Leaf 03: Spatial Relation Between Static Land Objects.}

Leaf 03 follows the path \emph{L1 Perception} \(\rightarrow\) \emph{G1 Scene Perception} \(\rightarrow\) \emph{T3 Static Spatial Relation}. It falls under scene perception because the answer is determined by the stable arrangement of land regions relative to a specified reference object.

This leaf evaluates the grounding of directional relations between static scene components. It requires identifying both the reference region and the queried region before comparing their positions.

\begin{figure*}[tbp]
\centering
\includegraphics[width=\textwidth]{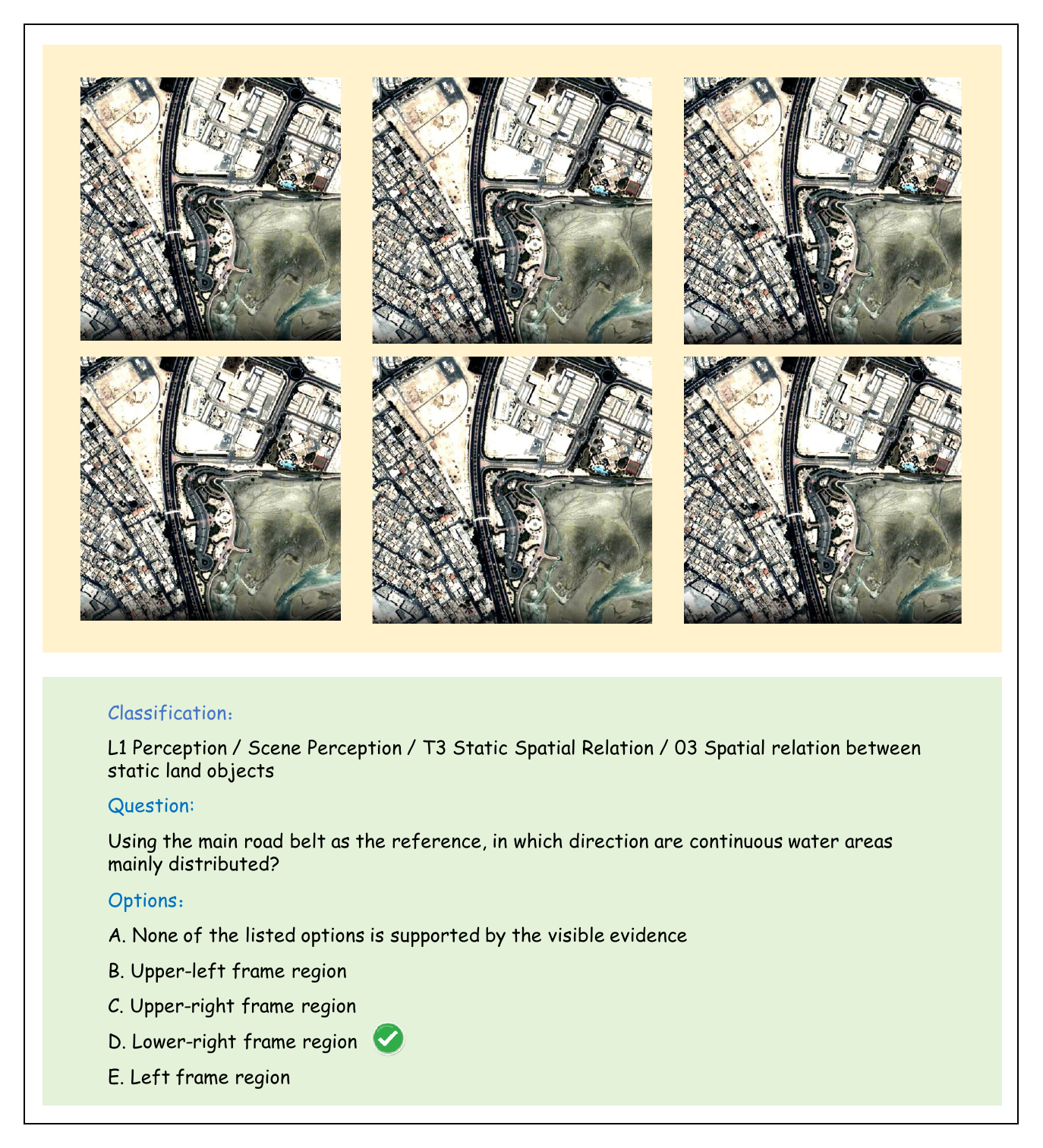}
\caption{Qualitative example for Leaf 03, spatial relation between static land objects.}
\label{fig:app-leaf-03}
\end{figure*}

As shown in Fig.~\ref{fig:app-leaf-03}, the main road belt is used as the reference, and the model must determine whether the continuous water areas lie mainly in the upper-left, upper-right, lower-right, or left part of the frame, or whether the relation is unsupported by the evidence.

\paragraph{Leaf 04: Target Identification.}

Leaf 04 follows the path \emph{L1 Perception} \(\rightarrow\) \emph{G2 Target Perception} \(\rightarrow\) \emph{T4 Target Recognition}. It is a target-perception leaf because its output is a semantic category derived from the localized appearance of a visible target.

This leaf tests fine-grained recognition of the queried object while separating target evidence from the surrounding scene context. The model must select a supported category while retaining the option to reject all listed categories when the evidence is insufficient.

\begin{figure*}[tbp]
\centering
\includegraphics[width=\textwidth]{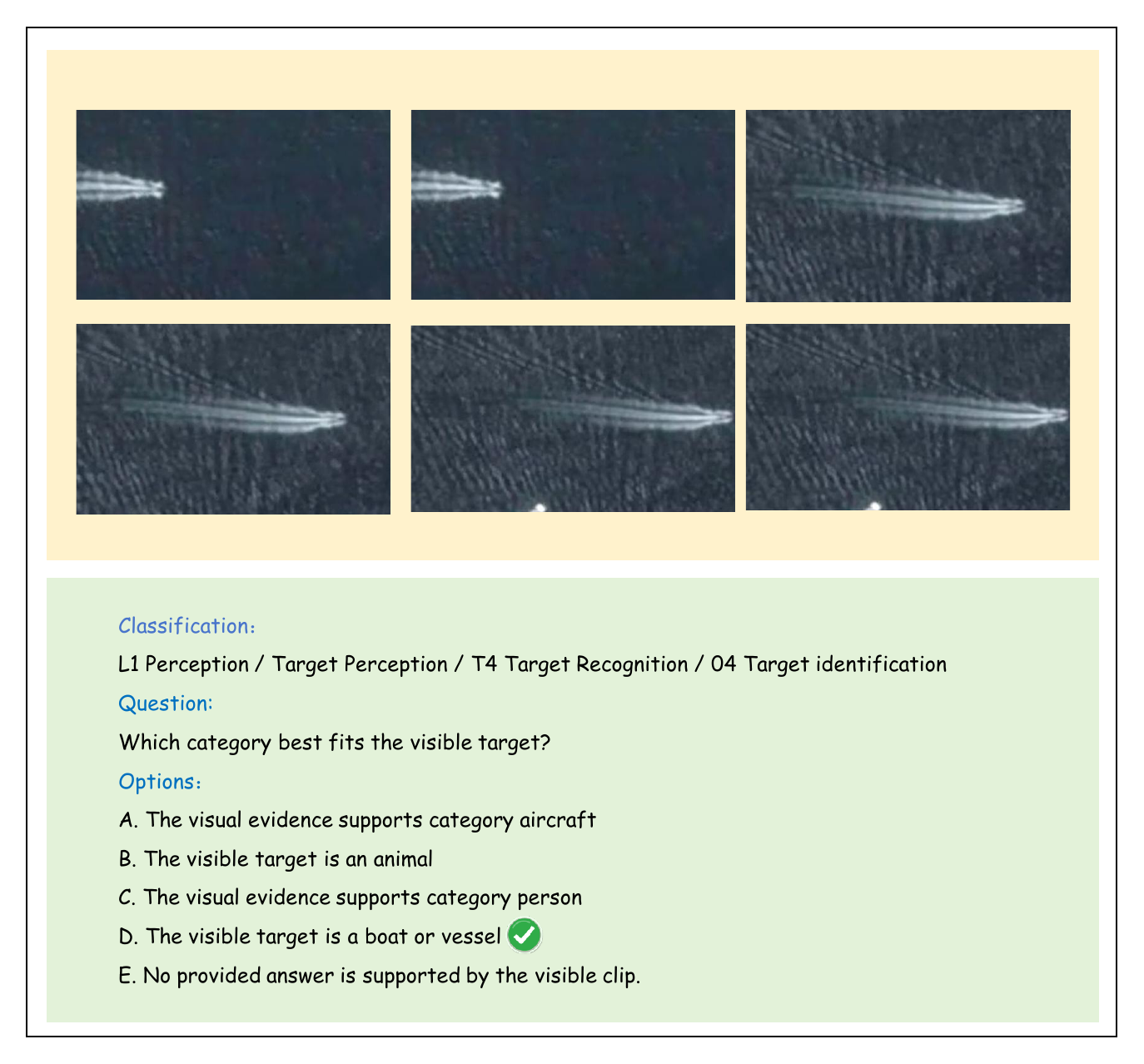}
\caption{Qualitative example for Leaf 04, target identification.}
\label{fig:app-leaf-04}
\end{figure*}

As shown in Fig.~\ref{fig:app-leaf-04}, the example asks which category fits the visible target and contrasts aircraft, person, animal, vessel, and a no-supported-answer option.

\paragraph{Leaf 05: Target Counting.}

Leaf 05 follows the path \emph{L1 Perception} \(\rightarrow\) \emph{G2 Target Perception} \(\rightarrow\) \emph{T5 Target Counting}. It belongs here because the required operation is to detect and enumerate all visible instances of a specified target class at a given time.

This leaf measures whether a model can avoid missed instances and double counting when the targets are small, crowded, or partially overlapping. The selected count must be supported by the visible evidence at the requested timestamp.

\begin{figure*}[tbp]
\centering
\includegraphics[width=\textwidth]{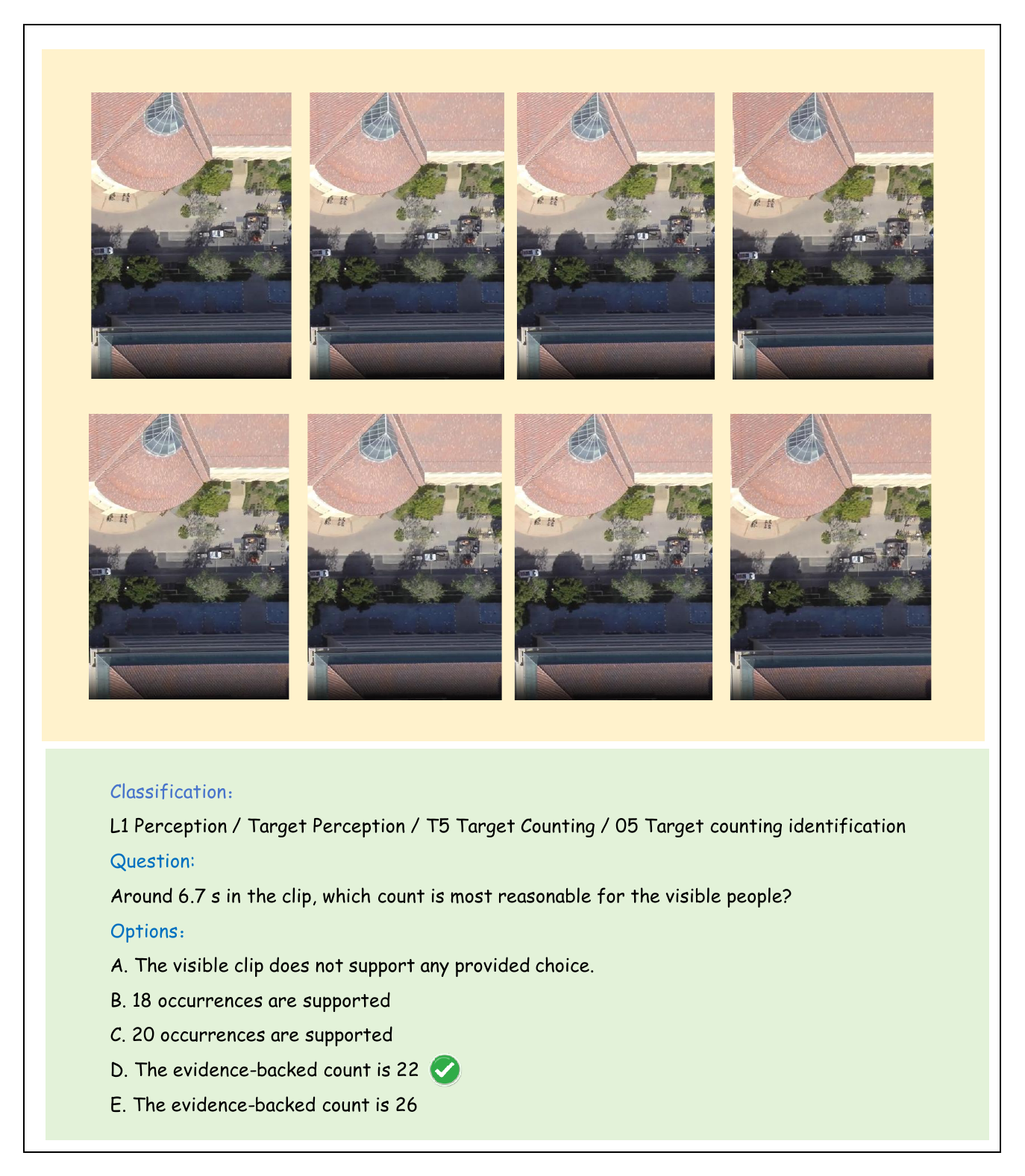}
\caption{Qualitative example for Leaf 05, target counting.}
\label{fig:app-leaf-05}
\end{figure*}

As shown in Fig.~\ref{fig:app-leaf-05}, the example asks for the number of visible people around 6.7~s and offers counts of 18, 20, 22, and 26, together with an unsupported-choice alternative.

\paragraph{Leaf 06: Reference-image Target Localization.}

Leaf 06 follows the path \emph{L1 Perception} \(\rightarrow\) \emph{G2 Target Perception} \(\rightarrow\) \emph{T6 Referring Target Localization}. It is categorized as target perception because a referring cue must first be matched to a visible target and then mapped to a spatial region.

This leaf evaluates cue-conditioned localization rather than open-ended target detection. The model must resolve the referred instance and report its position under the supplied grid convention.

\begin{figure*}[tbp]
\centering
\includegraphics[width=\textwidth]{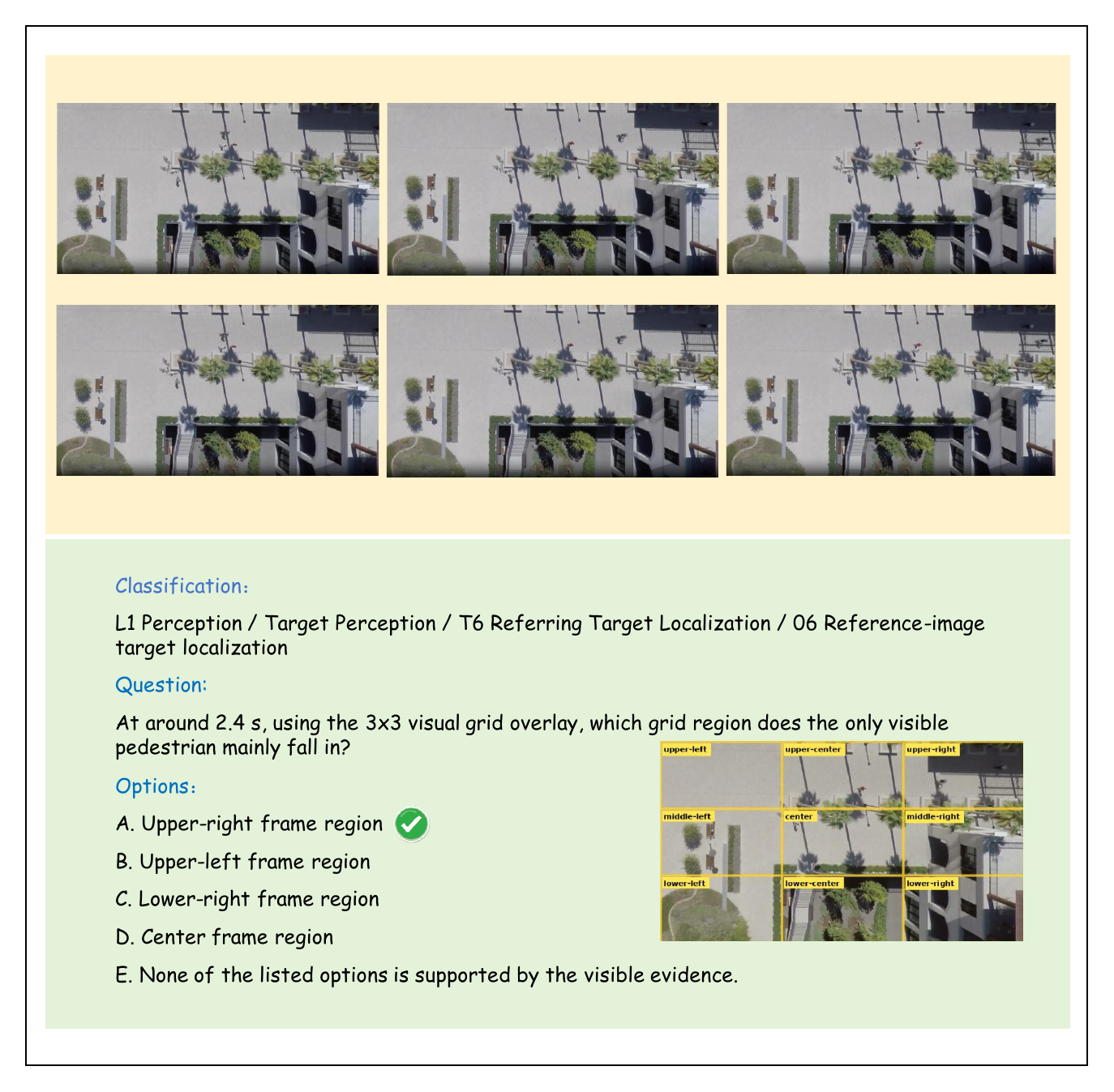}
\caption{Qualitative example for Leaf 06, reference-image target localization.}
\label{fig:app-leaf-06}
\end{figure*}

As shown in Fig.~\ref{fig:app-leaf-06}, the example asks where the only visible pedestrian lies in a \(3\times3\) grid around 2.4~s, with upper-right, upper-left, lower-right, center, and unsupported options.

\paragraph{Leaf 07: Action Recognition.}

Leaf 07 follows the path \emph{L1 Perception} \(\rightarrow\) \emph{G3 Action Perception} \(\rightarrow\) \emph{T7 Action Recognition}. It is placed under action perception because the label is supported by directly visible pose and motion evidence across frames.

This leaf tests recognition of a target's ongoing action without requiring a causal explanation or a longer event chain. Appearance, posture, and short-term motion must be combined to distinguish visually similar actions.

\begin{figure*}[tbp]
\centering
\includegraphics[width=\textwidth]{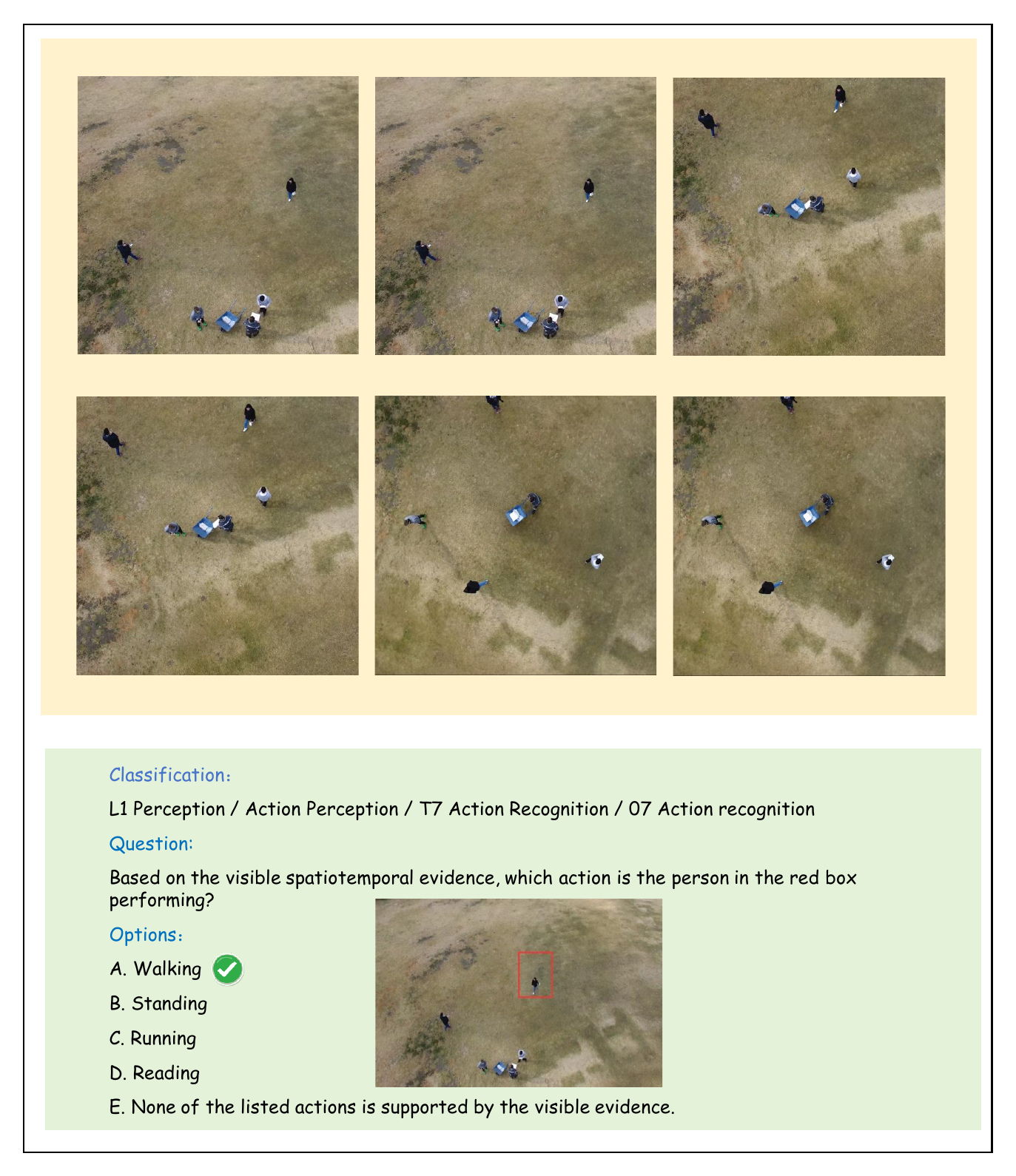}
\caption{Qualitative example for Leaf 07, action recognition.}
\label{fig:app-leaf-07}
\end{figure*}

As shown in Fig.~\ref{fig:app-leaf-07}, the example asks which activity is supported by the video and contrasts walking, standing, running, reading, and an evidence-insufficient alternative.

\paragraph{Leaf 08: Action Transition.}

Leaf 08 follows the path \emph{L1 Perception} \(\rightarrow\) \emph{G3 Action Perception} \(\rightarrow\) \emph{T8 Action Transition}. It belongs to this task because the answer is defined by a change between two directly observed action states.

This leaf assesses whether a model can segment the clip around an action transition and identify the state on each side of it. Correct recognition of both temporal segments is necessary, as recognizing either state alone is insufficient.

\begin{figure*}[tbp]
\centering
\includegraphics[width=\textwidth]{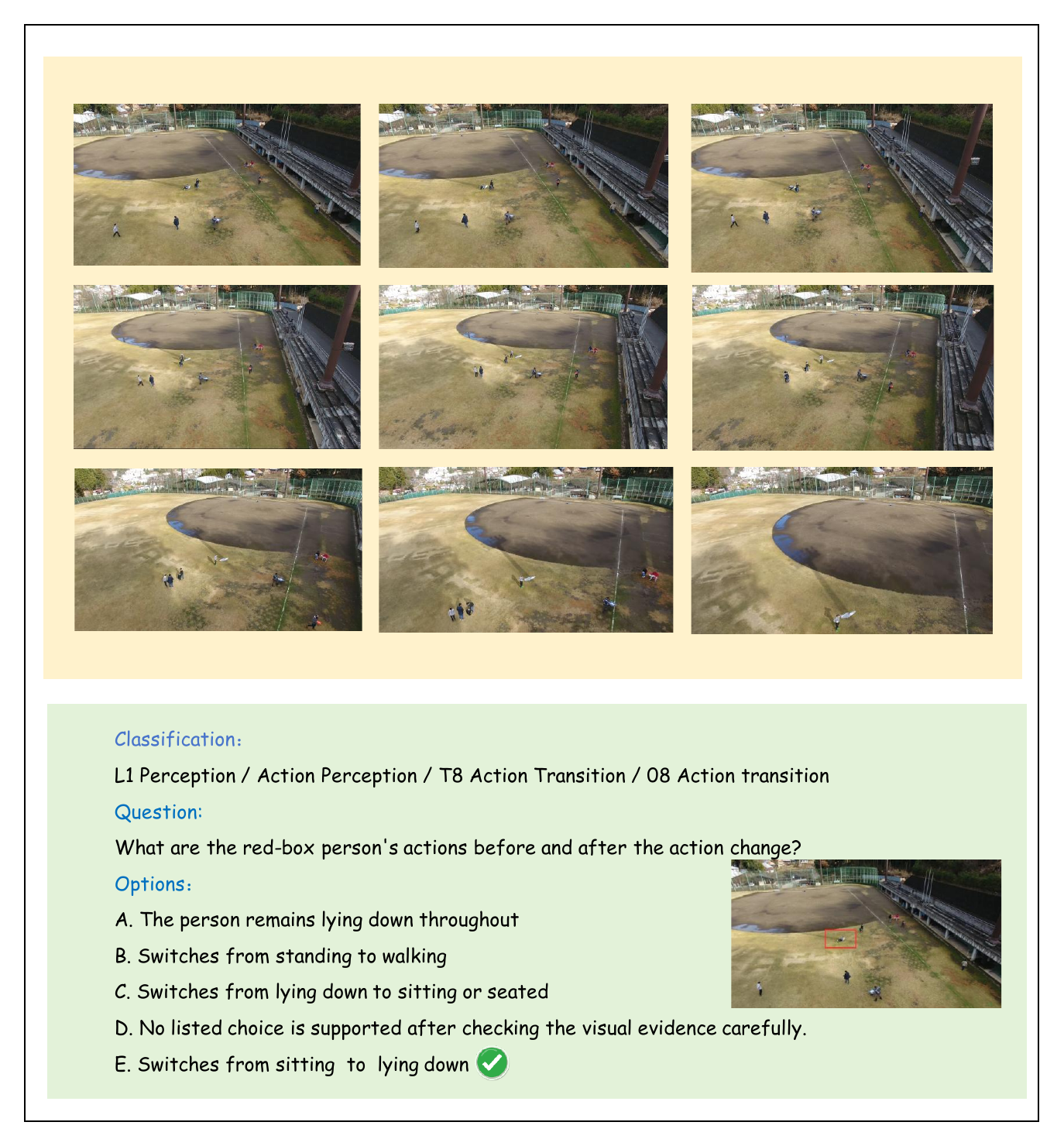}
\caption{Qualitative example for Leaf 08, action transition.}
\label{fig:app-leaf-08}
\end{figure*}

As shown in Fig.~\ref{fig:app-leaf-08}, the example tracks a red-box person and contrasts remaining lying down throughout, switching from standing to walking, switching from lying down to sitting, switching from sitting to lying down, and an unsupported alternative.

\paragraph{Leaf 09: Action Counting.}

Leaf 09 follows the path \emph{L1 Perception} \(\rightarrow\) \emph{G3 Action Perception} \(\rightarrow\) \emph{T9 Action Counting}. It is assigned to action perception because it counts boundaries between observable action states rather than counting target instances.

This leaf tests repeated temporal segmentation of a marked target's behavior. The model must distinguish genuine action-state changes from the continued execution of the same action.

\begin{figure*}[tbp]
\centering
\includegraphics[width=\textwidth]{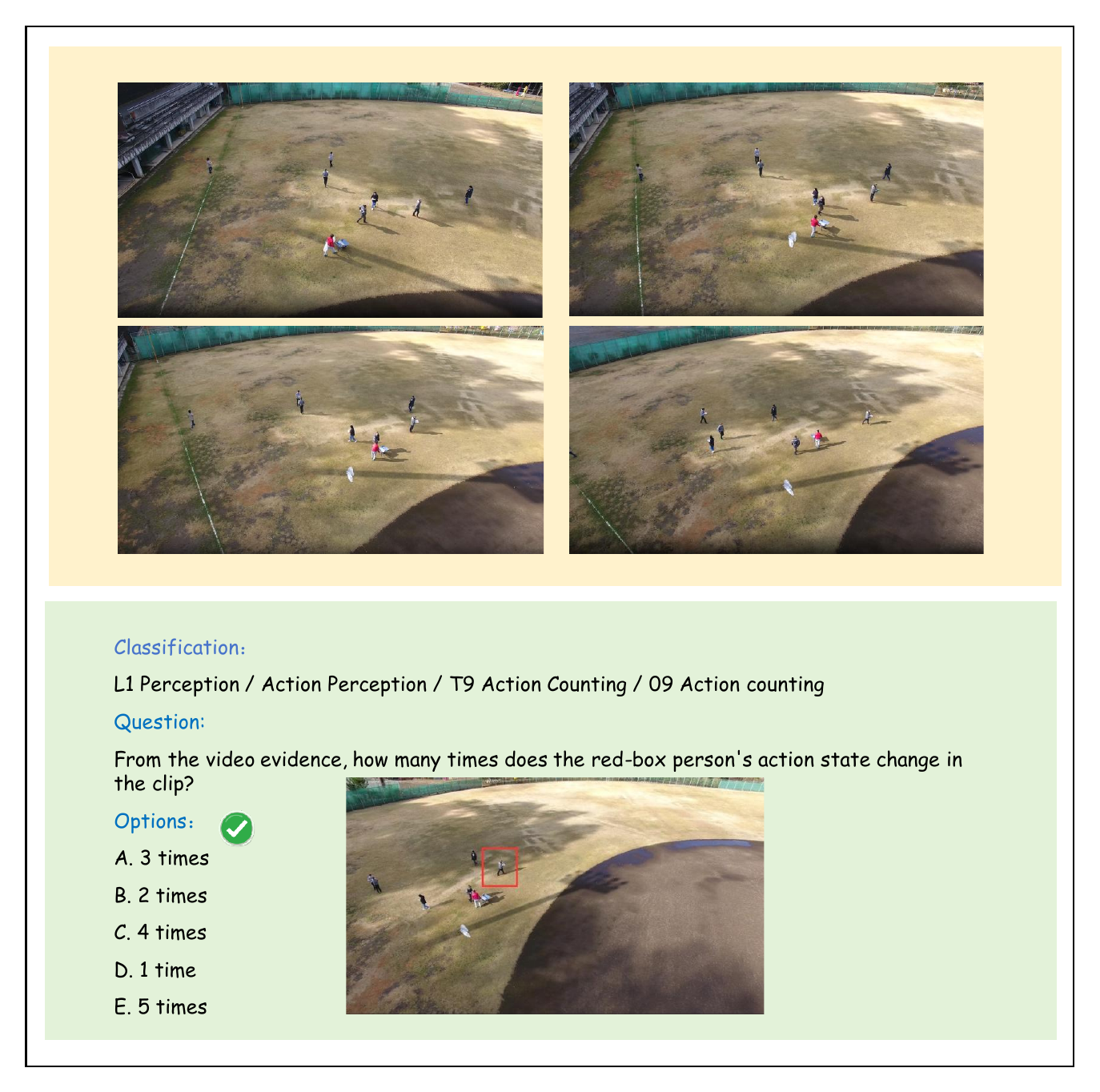}
\caption{Qualitative example for Leaf 09, action counting.}
\label{fig:app-leaf-09}
\end{figure*}

As shown in Fig.~\ref{fig:app-leaf-09}, the example asks how many times the red-box person's action state changes during the clip, with candidate counts ranging from one to five.

\paragraph{Leaf 10: Environmental Constraint Reasoning.}

Leaf 10 follows the path \emph{L2 Reasoning} \(\rightarrow\) \emph{G4 Complex-Environment Reasoning} \(\rightarrow\) \emph{T10 Event Reasoning}. It belongs to reasoning because the model must relate a target's activity to the surrounding scene structure and infer which environmental constraint best explains the observed motion.

This leaf tests event interpretation under physical layout constraints, including boundaries, passable corridors, curves, intersections, and blockages. The relevant evidence therefore combines target behavior with persistent environmental context.

\begin{figure*}[tbp]
\centering
\includegraphics[width=\textwidth]{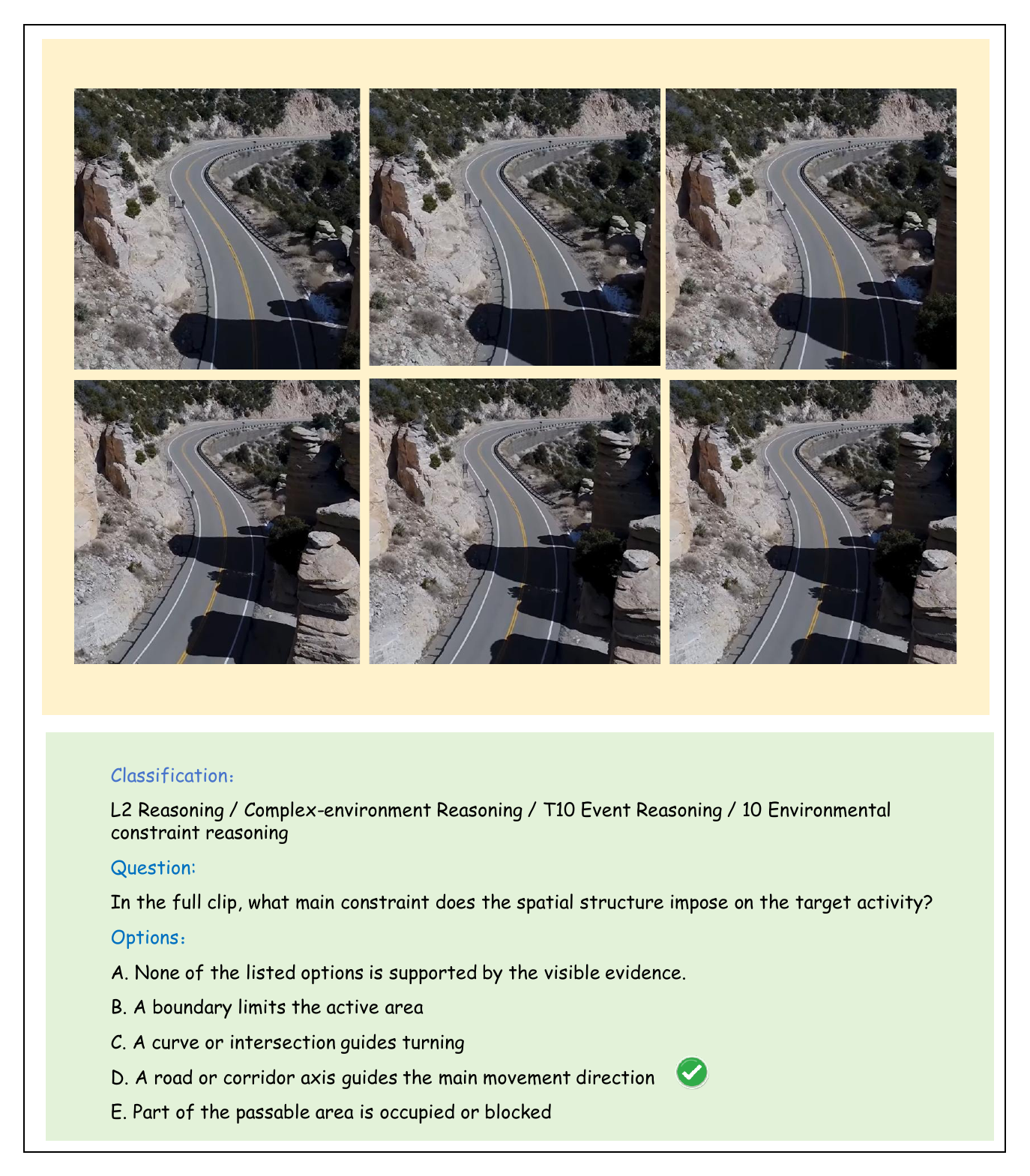}
\caption{Qualitative example for Leaf 10, environmental constraint reasoning.}
\label{fig:app-leaf-10}
\end{figure*}

As shown in Fig.~\ref{fig:app-leaf-10}, the example asks what spatial constraint governs the observed activity and contrasts a limiting boundary, a turning guide, a road or corridor axis, an occupied passable area, and insufficient evidence.

\paragraph{Leaf 11: Target State Inference.}

Leaf 11 follows the path \emph{L2 Reasoning} \(\rightarrow\) \emph{G4 Complex-Environment Reasoning} \(\rightarrow\) \emph{T10 Event Reasoning}. It is grouped with event reasoning because the target's state or role cannot be determined from appearance alone and must instead be inferred from its interaction with the surrounding event.

This leaf evaluates contextual inference about why a target appears at a marked location or what event-related state it occupies. The model must compare plausible explanations and reject those inconsistent with the visible sequence.

\begin{figure*}[tbp]
\centering
\includegraphics[width=\textwidth]{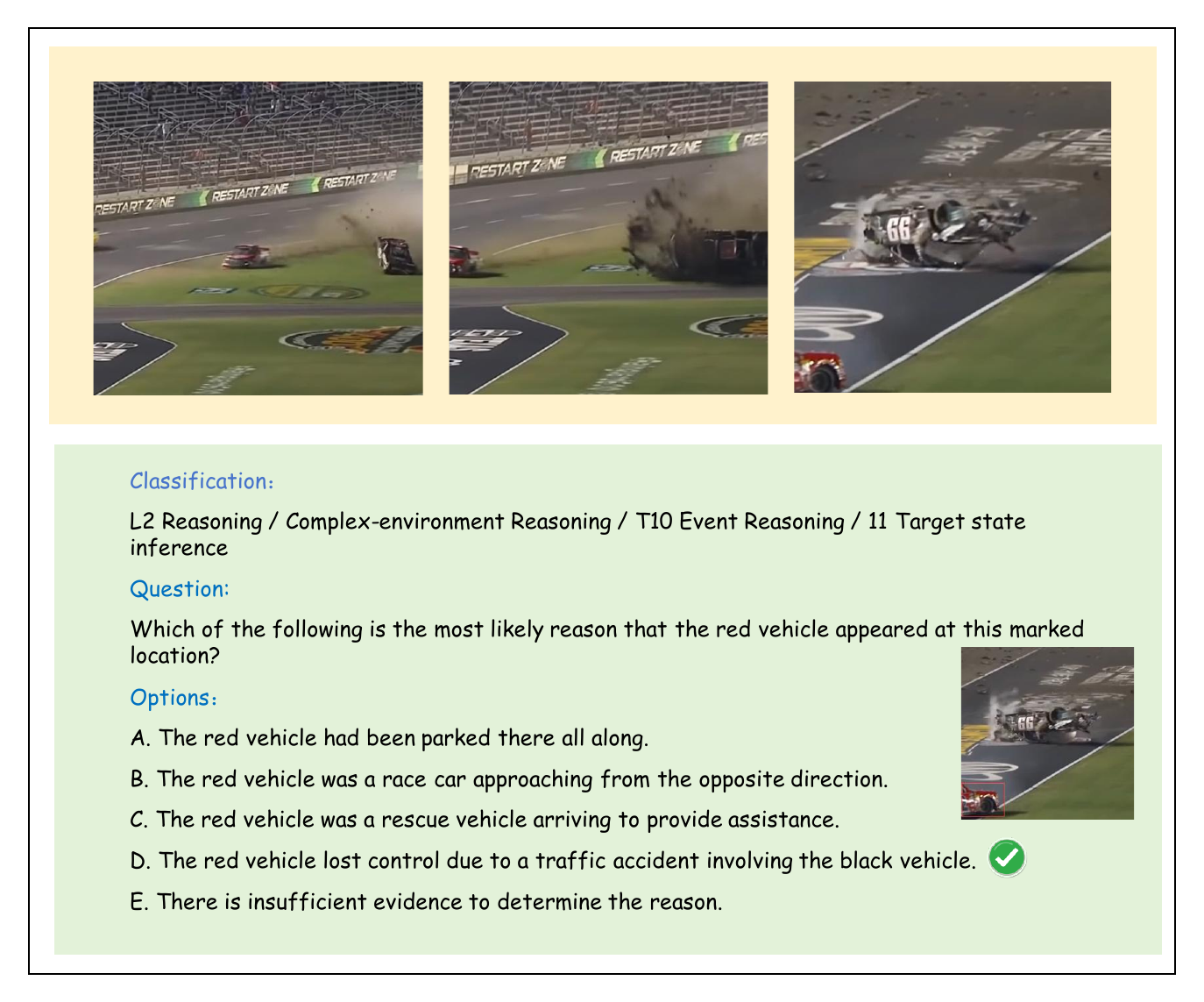}
\caption{Qualitative example for Leaf 11, target state inference.}
\label{fig:app-leaf-11}
\end{figure*}

As shown in Fig.~\ref{fig:app-leaf-11}, the example asks why a red vehicle appears at the marked location and contrasts prior parking, an opposing race car, rescue arrival, accident-related loss of control, and insufficient evidence.

\paragraph{Leaf 12: Main Visible Activity Recognition.}

Leaf 12 follows the path \emph{L2 Reasoning} \(\rightarrow\) \emph{G4 Complex-Environment Reasoning} \(\rightarrow\) \emph{T10 Event Reasoning}. It is an event-reasoning leaf because the answer summarizes the dominant activity formed by multiple targets and the scene context over the clip, rather than labeling an isolated object or action.

This leaf tests whether a model can aggregate distributed visual evidence into the principal activity represented by the sequence. Incidental objects and background structures should not outweigh the sustained activity pattern.

\begin{figure*}[tbp]
\centering
\includegraphics[width=\textwidth]{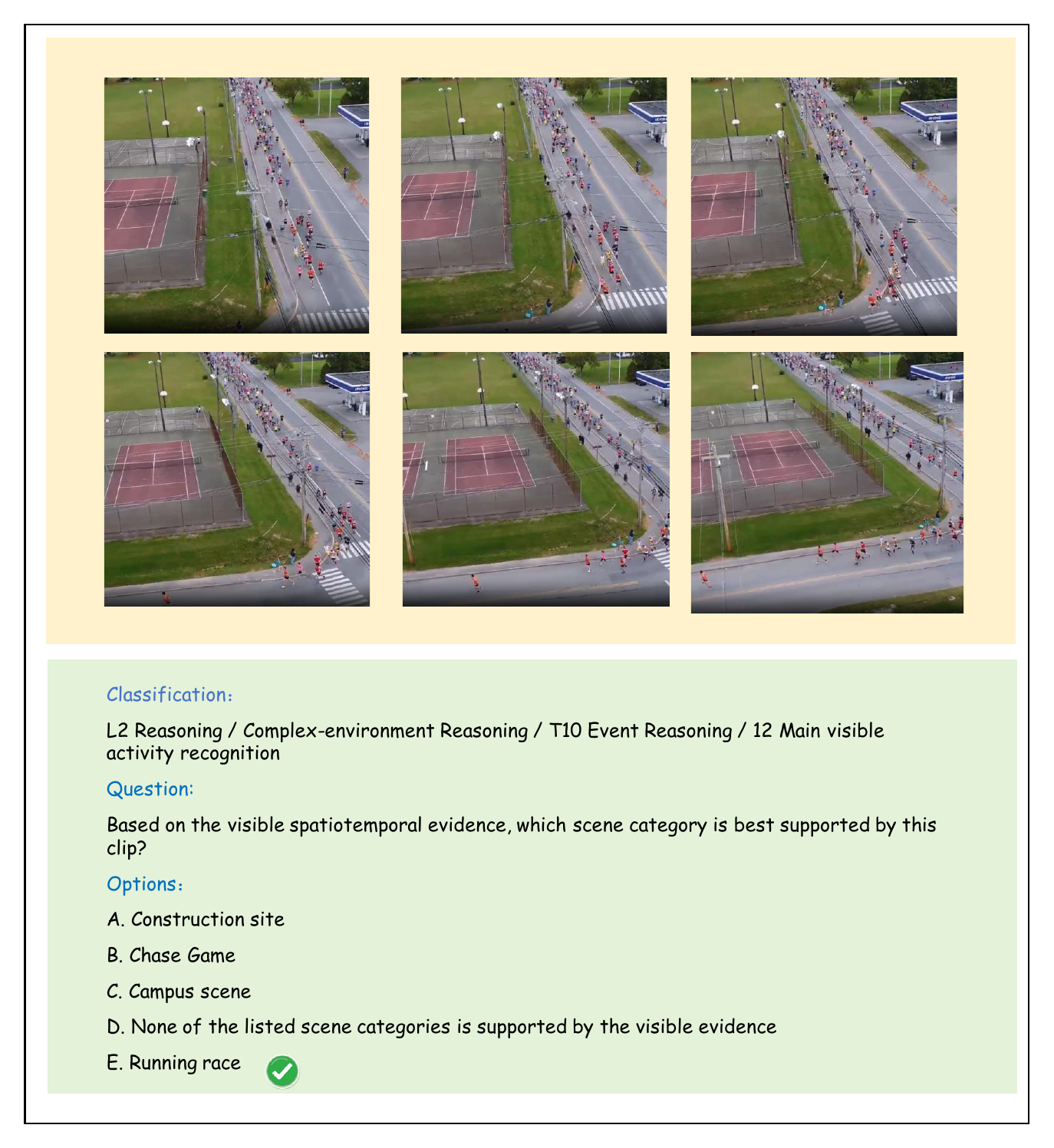}
\caption{Qualitative example for Leaf 12, main visible activity recognition.}
\label{fig:app-leaf-12}
\end{figure*}

As shown in Fig.~\ref{fig:app-leaf-12}, the example asks which category best describes the clip and contrasts a construction site, chase game, campus scene, running race, and an unsupported category.

\paragraph{Leaf 13: Disaster Evidence Monitoring.}

Leaf 13 follows the path \emph{L2 Reasoning} \(\rightarrow\) \emph{G4 Complex-Environment Reasoning} \(\rightarrow\) \emph{T11 Risk Assessment}. It belongs to risk assessment because visible environmental changes or damage cues must be interpreted as evidence of a potential hazard category.

This leaf evaluates the recognition and monitoring of disaster-related evidence while separating visually confusable risk types. The judgment is restricted to cues present in the clip and does not require unsupported claims about the cause or severity of the event.

\begin{figure*}[tbp]
\centering
\includegraphics[width=\textwidth]{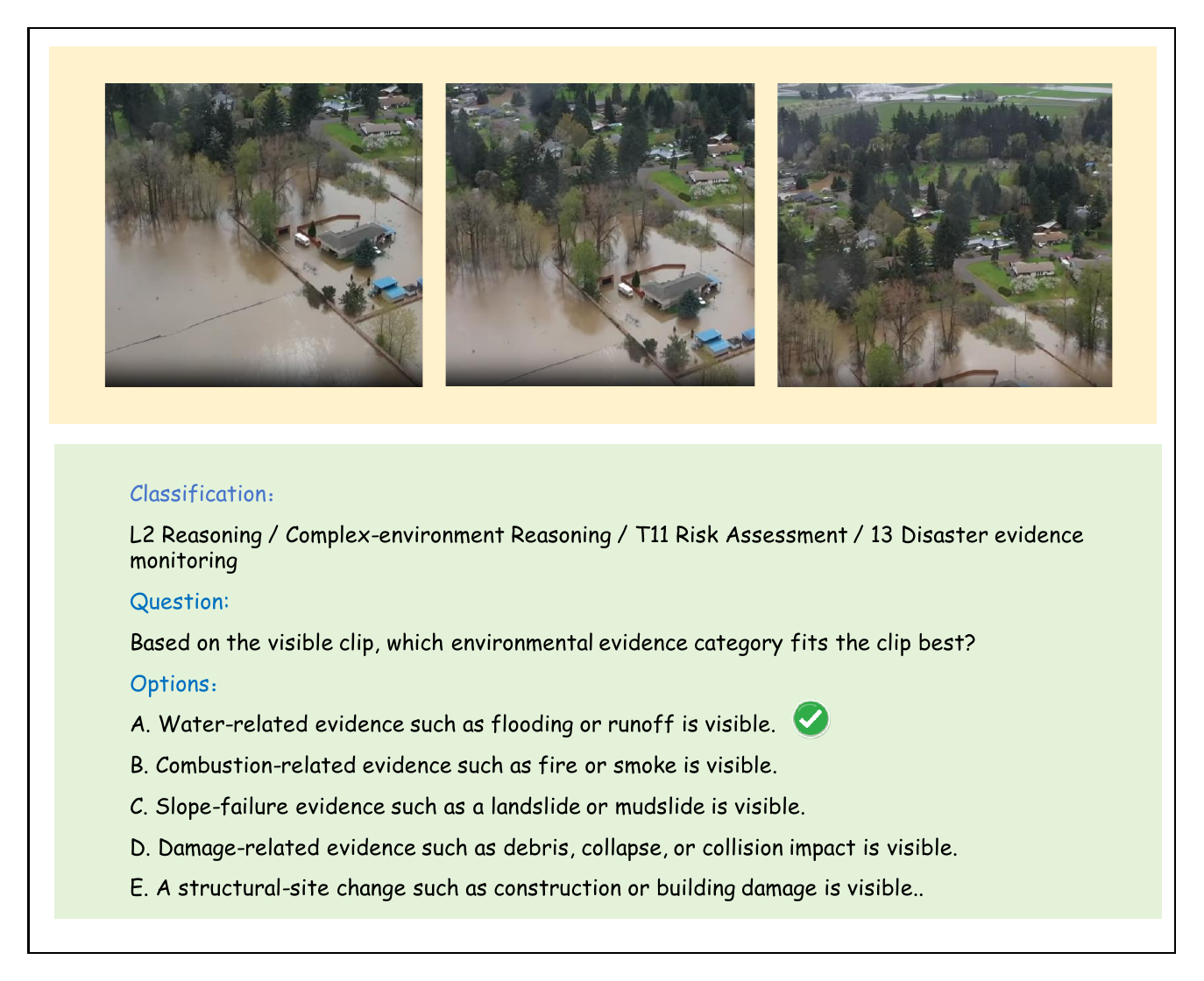}
\caption{Qualitative example for Leaf 13, disaster evidence monitoring.}
\label{fig:app-leaf-13}
\end{figure*}

As shown in Fig.~\ref{fig:app-leaf-13}, the example asks which environmental evidence category fits the clip and compares water-related, combustion-related, slope-failure, damage-related, and structural-site-change evidence.

\paragraph{Leaf 14: Abnormal Behavior.}

Leaf 14 follows the path \emph{L2 Reasoning} \(\rightarrow\) \emph{G4 Complex-Environment Reasoning} \(\rightarrow\) \emph{T11 Risk Assessment}. It is assigned to risk assessment because the model must decide whether an observed interaction constitutes dangerous behavior and support that judgment with temporal evidence.

This leaf tests the contextual detection of abnormal actions together with approximate event timing. It requires distinguishing the absence of danger from different hazardous interactions occurring at competing timestamps.

\begin{figure*}[tbp]
\centering
\includegraphics[width=\textwidth]{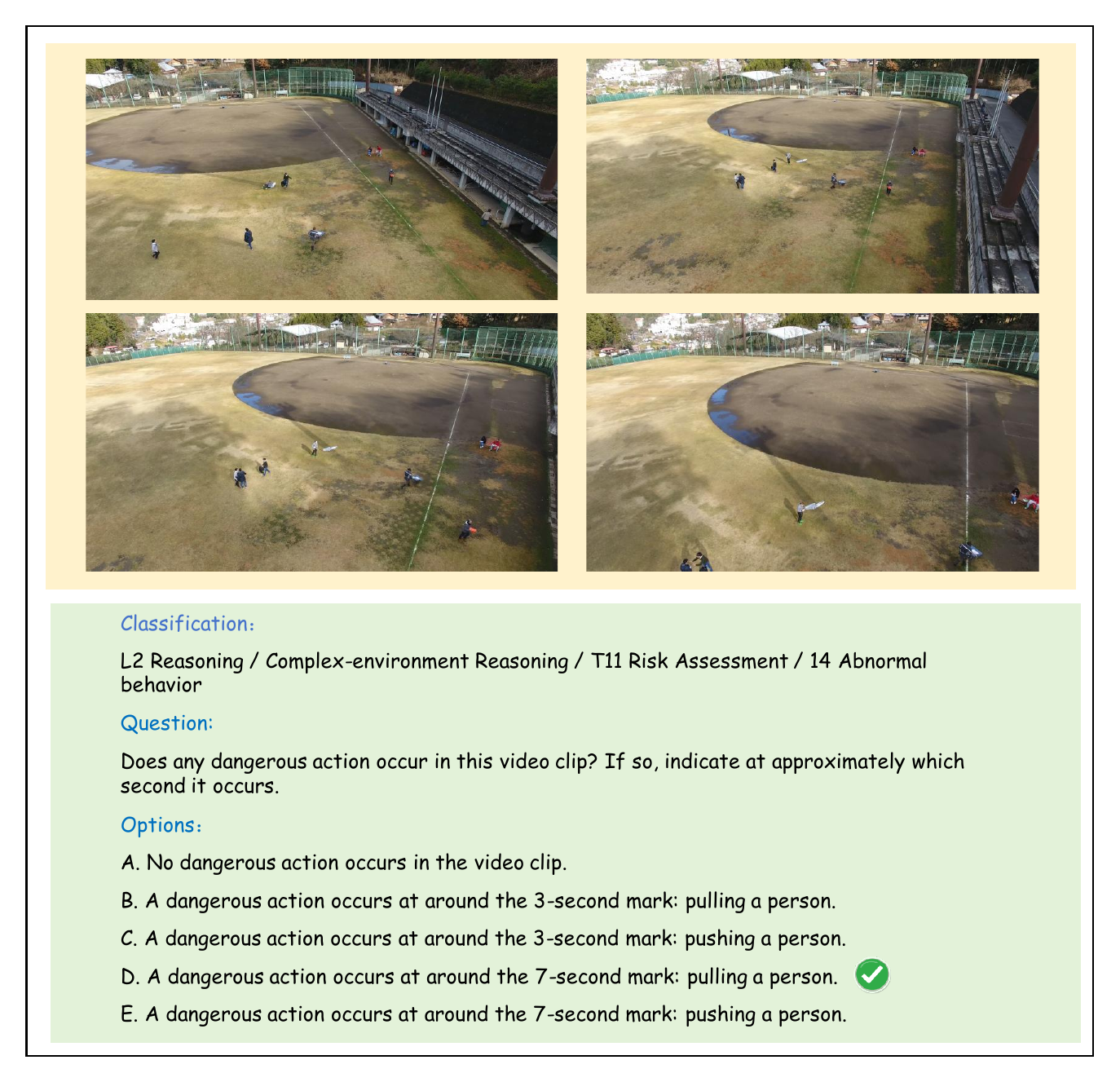}
\caption{Qualitative example for Leaf 14, abnormal behavior.}
\label{fig:app-leaf-14}
\end{figure*}

As shown in Fig.~\ref{fig:app-leaf-14}, the example asks whether pushing or pulling occurs around 3~s or 7~s, while also allowing the conclusion that no dangerous action occurs.

\paragraph{Leaf 15: Long-term Trajectory Summary for Small Moving Targets.}

Leaf 15 follows the path \emph{L2 Reasoning} \(\rightarrow\) \emph{G5 Spatiotemporal Evolution Reasoning} \(\rightarrow\) \emph{T12 Global Spatiotemporal Understanding}. It belongs to global spatiotemporal understanding because the trajectory must be integrated over the full clip despite the small size of the moving target.

This leaf tests whether a model can maintain a target track across a long temporal window and compress it into a start-to-end directional summary. Single-frame position cues are insufficient for the required description.

\begin{figure*}[tbp]
\centering
\includegraphics[width=\textwidth]{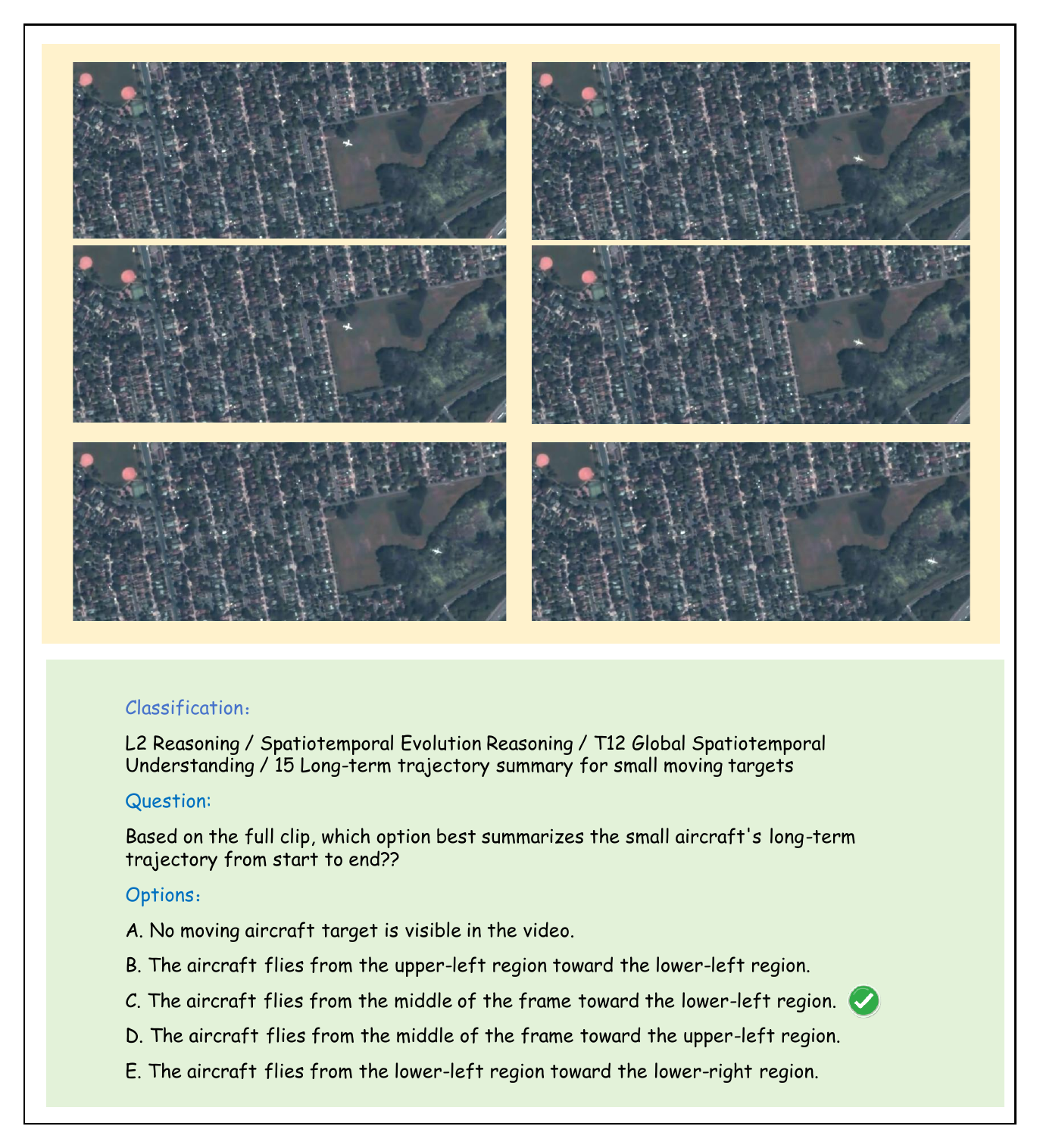}
\caption{Qualitative example for Leaf 15, long-term trajectory summary for small moving targets.}
\label{fig:app-leaf-15}
\end{figure*}

As shown in Fig.~\ref{fig:app-leaf-15}, the example asks which option best summarizes the small aircraft's long-term trajectory and contrasts motion from the upper-left to the lower-left, from the frame center to the lower-left, from the frame center to the upper-left, and from the lower-left to the lower-right, together with the possibility that no moving aircraft is visible.

\paragraph{Leaf 16: Displacement and Speed Estimation.}

Leaf 16 follows the path \emph{L2 Reasoning} \(\rightarrow\) \emph{G5 Spatiotemporal Evolution Reasoning} \(\rightarrow\) \emph{T12 Global Spatiotemporal Understanding}. It is placed here because the overall movement direction and displacement magnitude must be derived by comparing the target's positions across the observed sequence.

This leaf assesses quantitative motion reasoning at a coarse, evidence-supported resolution. The model must jointly estimate direction and displacement range rather than identify movement in only one frame.

\begin{figure*}[tbp]
\centering
\includegraphics[width=\textwidth]{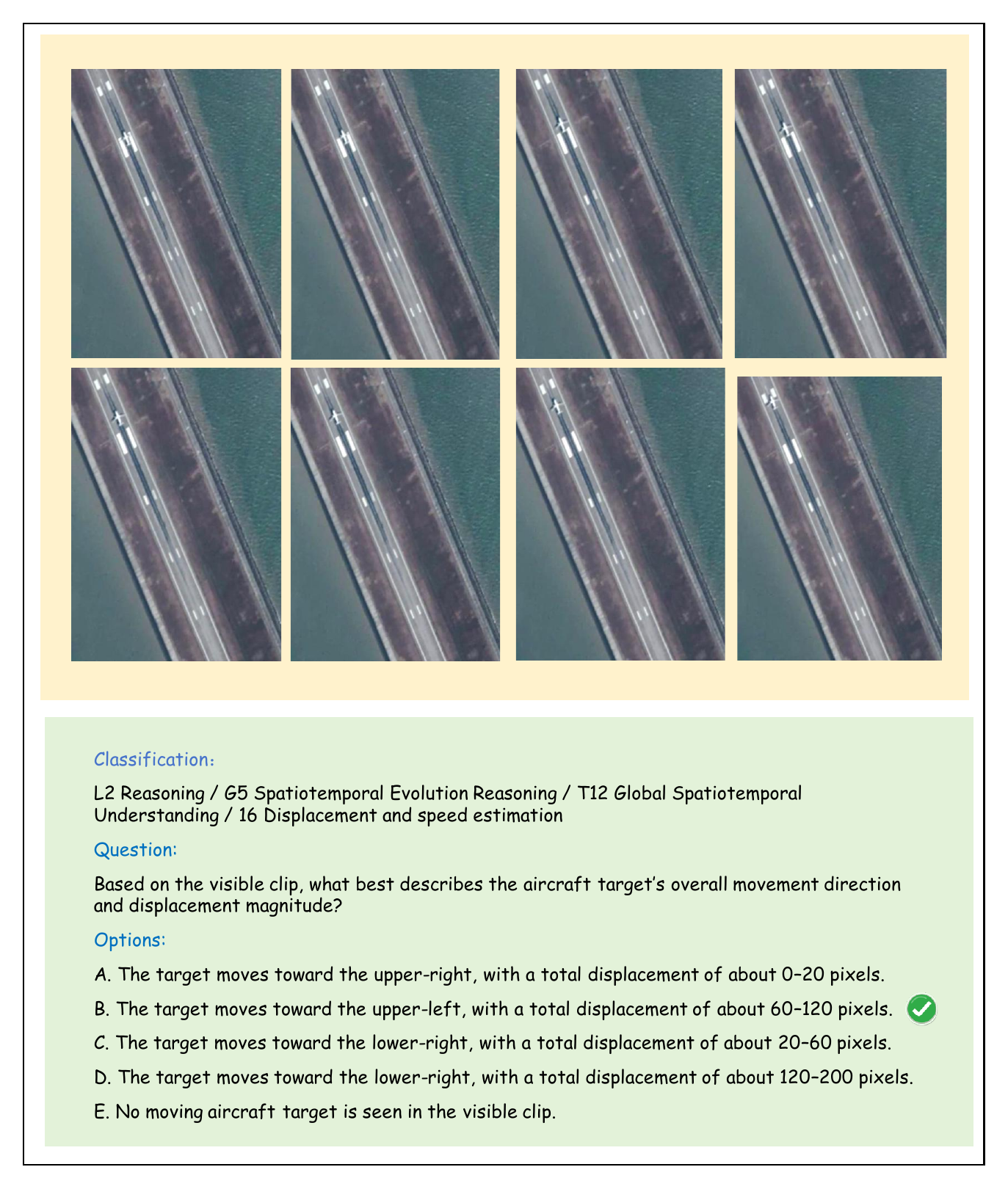}
\caption{Qualitative example for Leaf 16, displacement and speed estimation.}
\label{fig:app-leaf-16}
\end{figure*}

As shown in Fig.~\ref{fig:app-leaf-16}, the example asks for the aircraft's overall movement direction and displacement magnitude and contrasts upper-right motion over 0--20 pixels, upper-left motion over 60--120 pixels, lower-right motion over 20--60 pixels, lower-right motion over 120--200 pixels, and the possibility that no moving aircraft is visible.

\paragraph{Leaf 17: Temporal Ordering.}

Leaf 17 follows the path \emph{L2 Reasoning} \(\rightarrow\) \emph{G5 Spatiotemporal Evolution Reasoning} \(\rightarrow\) \emph{T13 Local-Temporal Understanding}. It belongs to local-temporal understanding because the required relation is the order of actions across specified neighboring intervals.

This leaf tests whether a model can preserve event chronology while recognizing the action performed in each segment. An answer containing the correct set of actions but presenting them in the wrong order should therefore be rejected.

\begin{figure*}[tbp]
\centering
\includegraphics[width=\textwidth]{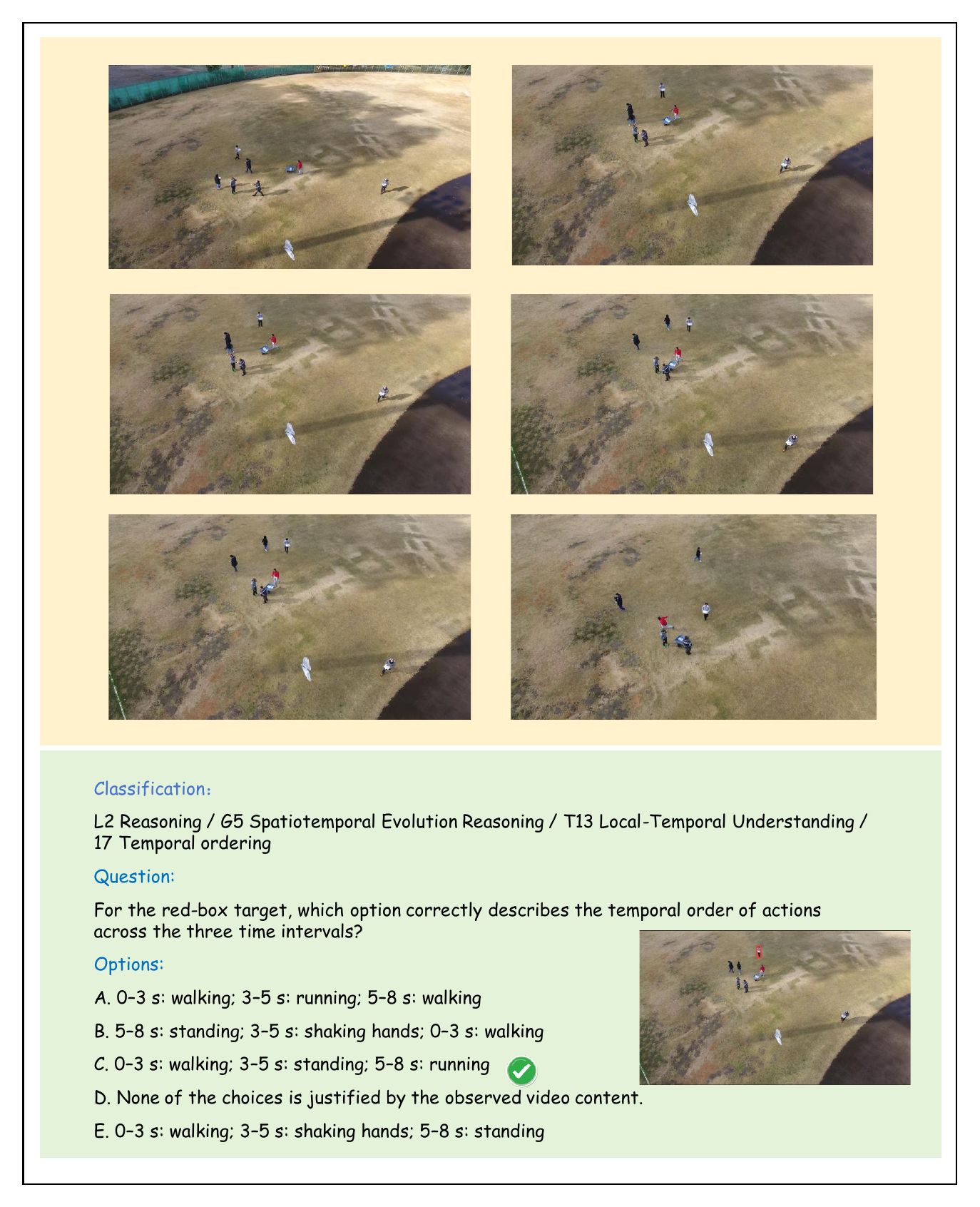}
\caption{Qualitative example for Leaf 17, temporal ordering.}
\label{fig:app-leaf-17}
\end{figure*}

As shown in Fig.~\ref{fig:app-leaf-17}, the example orders the red-box target's actions over 0--3~s, 3--5~s, and 5--8~s using combinations of walking, shaking hands, standing, and running.

\paragraph{Leaf 18: Action-duration Comparison.}

Leaf 18 follows the path \emph{L2 Reasoning} \(\rightarrow\) \emph{G5 Spatiotemporal Evolution Reasoning} \(\rightarrow\) \emph{T13 Local-Temporal Understanding}. It is assigned to this task because the answer depends on accumulating and comparing the durations of different action intervals for the same target.

This leaf evaluates temporal measurement beyond action recognition or transition detection. Repeated intervals of the same activity must be combined before selecting the activity with the longest total duration.

\begin{figure*}[tbp]
\centering
\includegraphics[width=\textwidth]{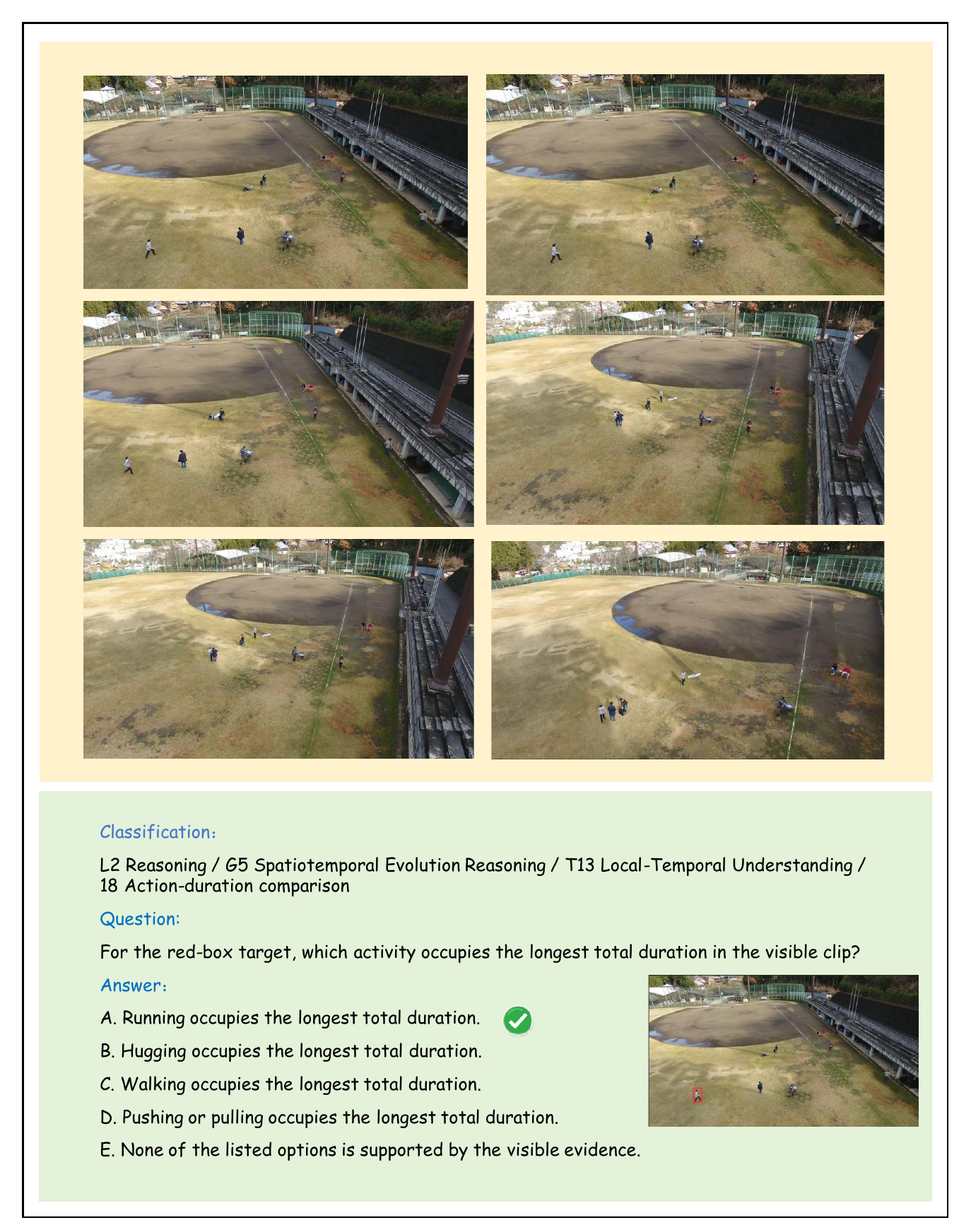}
\caption{Qualitative example for Leaf 18, action-duration comparison.}
\label{fig:app-leaf-18}
\end{figure*}

As shown in Fig.~\ref{fig:app-leaf-18}, the example asks which activity occupies the longest total duration for the red-box target, contrasting running, hugging, walking, pushing or pulling, and an unsupported alternative.

\paragraph{Leaf 19: Concurrent Visibility Interval for Multiple Targets.}

Leaf 19 follows the path \emph{L2 Reasoning} \(\rightarrow\) \emph{G5 Spatiotemporal Evolution Reasoning} \(\rightarrow\) \emph{T13 Local-Temporal Understanding}. It belongs here because the answer is obtained by intersecting the visibility intervals of multiple marked targets.

This leaf tests synchronized temporal tracking rather than independent target detection. The model must identify when all queried targets are visible simultaneously and select the closest supported interval.

\begin{figure*}[tbp]
\centering
\includegraphics[width=\textwidth]{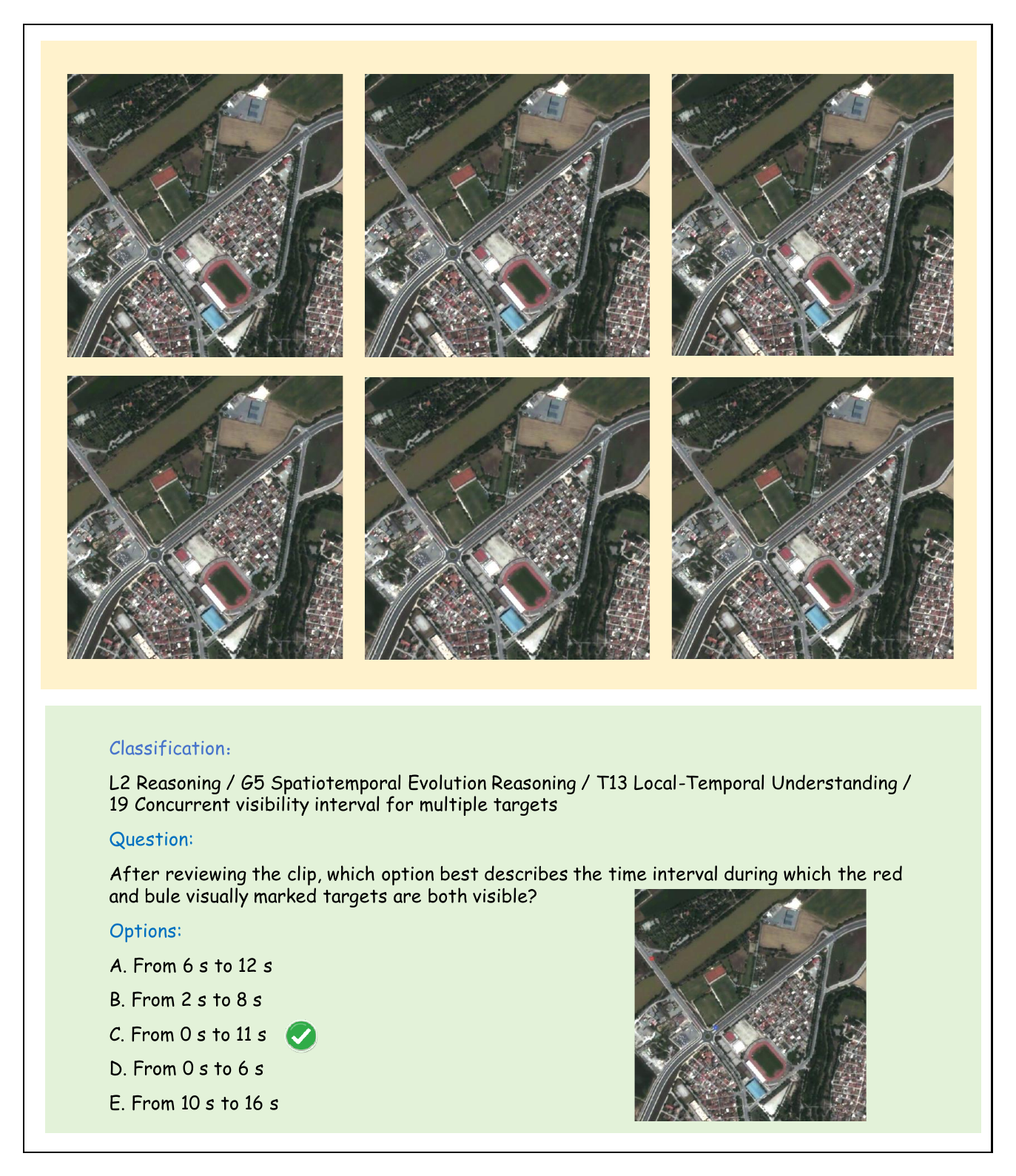}
\caption{Qualitative example for Leaf 19, concurrent visibility interval for multiple targets.}
\label{fig:app-leaf-19}
\end{figure*}

As shown in Fig.~\ref{fig:app-leaf-19}, the example asks when the red- and blue-marked targets are both visible and compares the intervals 6--12~s, 2--8~s, 0--11~s, 0--6~s, and 10--16~s.

\paragraph{Leaf 20: Single-event Spatiotemporal Localization.}

Leaf 20 follows the path \emph{L2 Reasoning} \(\rightarrow\) \emph{G6 Spatiotemporal Localization Reasoning} \(\rightarrow\) \emph{T14 Single-Event Spatiotemporal Localization}. It is assigned to this task because one event must be detected and localized to the temporal interval in which it occurs.

This leaf tests event-boundary identification for a specified target and event type. The selected interval must closely cover the observed event rather than merely contain the target somewhere in the clip.

\begin{figure*}[tbp]
\centering
\includegraphics[width=\textwidth]{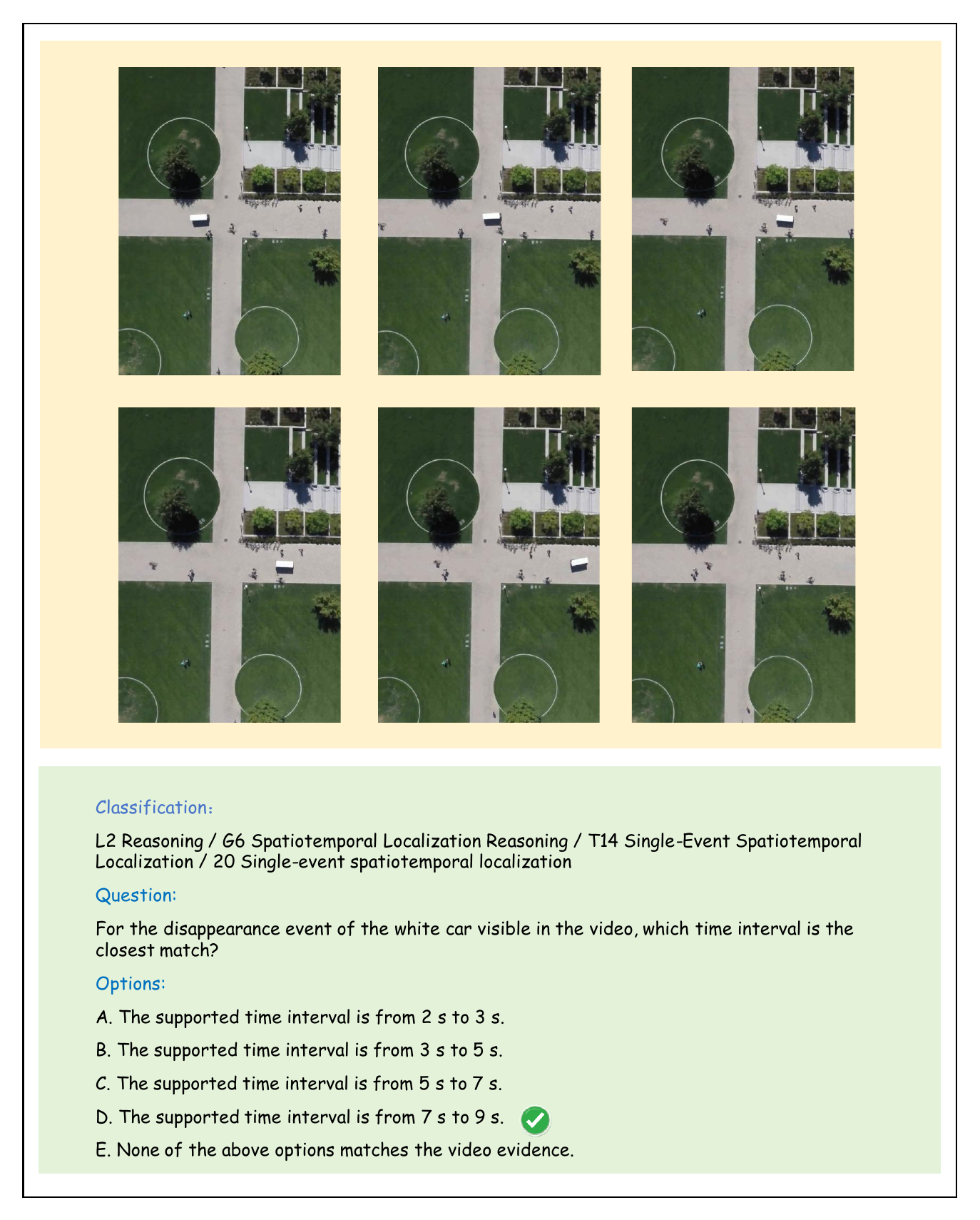}
\caption{Qualitative example for Leaf 20, single-event spatiotemporal localization.}
\label{fig:app-leaf-20}
\end{figure*}

As shown in Fig.~\ref{fig:app-leaf-20}, the example asks when a visible white car disappears and offers 2--3~s, 3--5~s, 5--7~s, and 7--9~s intervals, together with a no-match option.

\paragraph{Leaf 21: Regional Trajectory Co-occurrence.}

Leaf 21 follows the path \emph{L2 Reasoning} \(\rightarrow\) \emph{G6 Spatiotemporal Localization Reasoning} \(\rightarrow\) \emph{T15 Multi-Event Spatiotemporal Localization}. It belongs to multi-event localization because the answer jointly constrains multiple target trajectories by time and region.

This leaf evaluates whether a model can verify co-occurrence only when the queried identities, spatial region, and temporal window all agree. Observing both targets somewhere in the clip is not sufficient.

\begin{figure*}[tbp]
\centering
\includegraphics[width=0.94\textwidth]{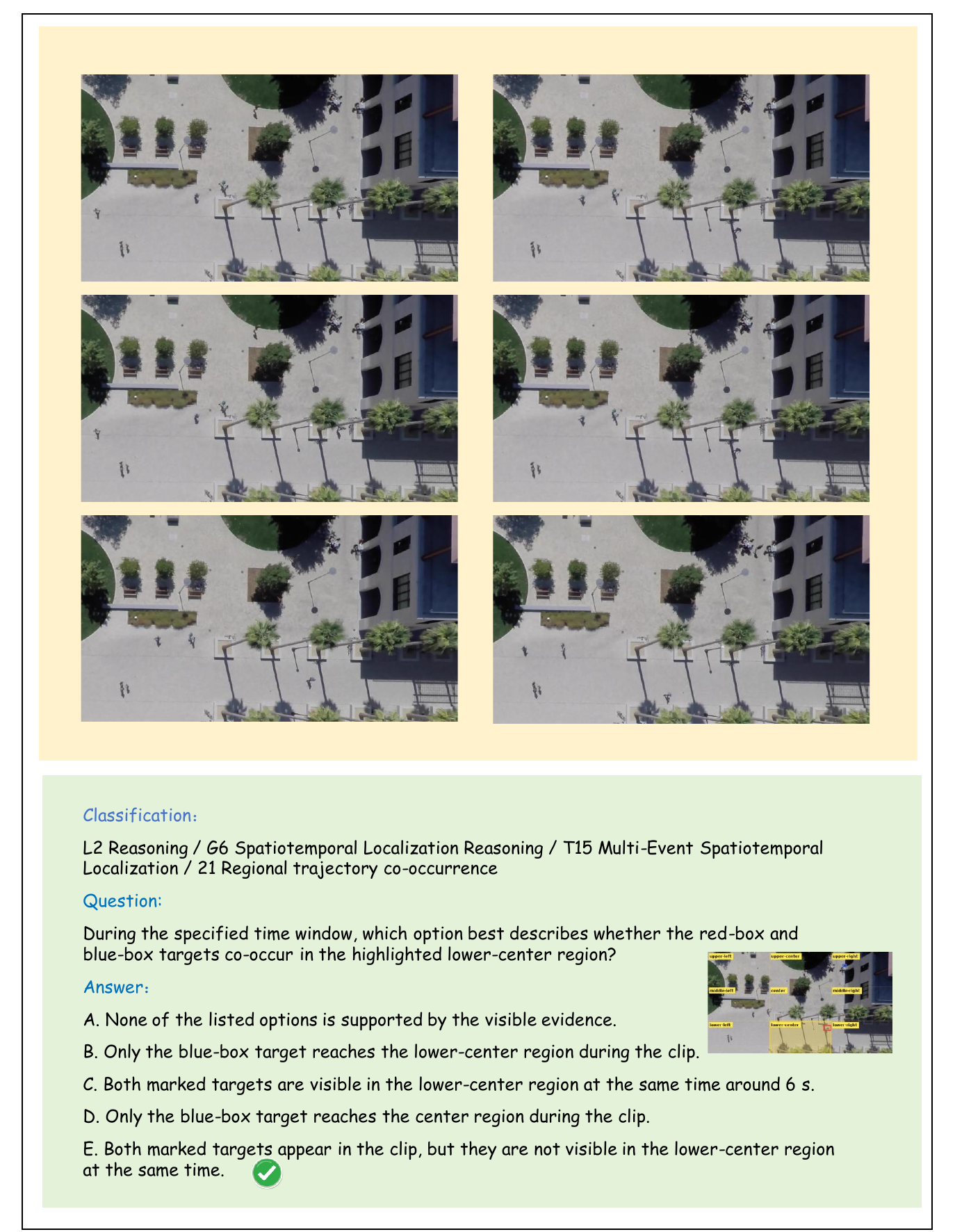}
\caption{Qualitative example for Leaf 21, regional trajectory co-occurrence.}
\label{fig:app-leaf-21}
\end{figure*}

As shown in Fig.~\ref{fig:app-leaf-21}, the example asks whether the red-box and blue-box targets are simultaneously visible in the highlighted lower-center region around 6~s and contrasts joint, single-target, and nonconcurrent alternatives.

\paragraph{Leaf 22: Relation-constrained Candidate Localization.}

Leaf 22 follows the path \emph{L2 Reasoning} \(\rightarrow\) \emph{G7 Spatiotemporal Consistency Reasoning} \(\rightarrow\) \emph{T16 Spatial Relation Consistency}. It is placed under spatial relation consistency because a candidate is valid only if its attributes and its relation to the referenced scene region are jointly satisfied.

This leaf tests localization after relational and semantic filtering. The model must first identify candidates that satisfy the stated description and then map the surviving candidate to the requested grid area.

\begin{figure*}[tbp]
\centering
\includegraphics[width=\textwidth]{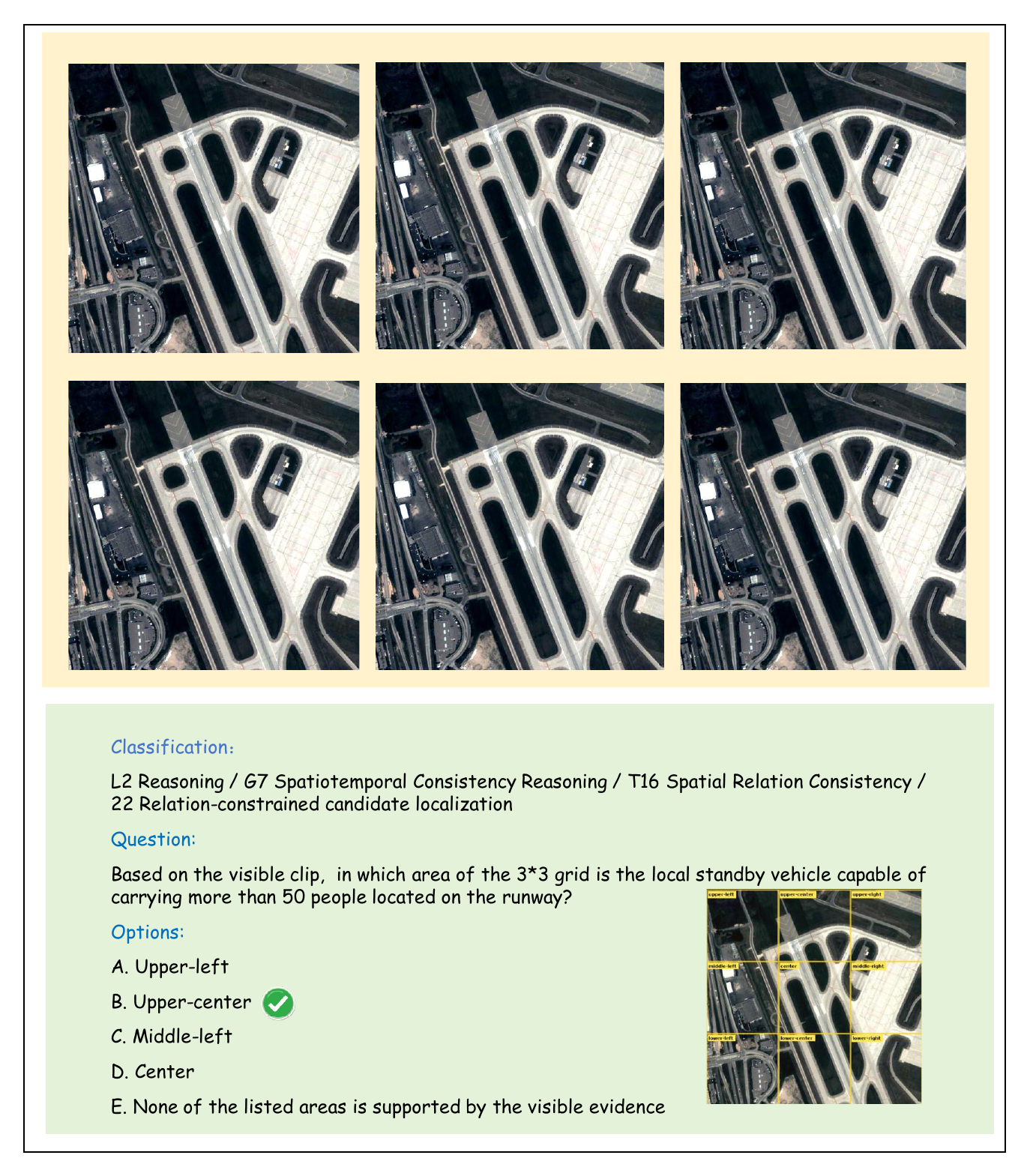}
\caption{Qualitative example for Leaf 22, relation-constrained candidate localization.}
\label{fig:app-leaf-22}
\end{figure*}

As shown in Fig.~\ref{fig:app-leaf-22}, the example asks where a local standby vehicle capable of carrying more than 50 people is located on the runway and contrasts the upper-left, upper-center, middle-left, and center regions of a \(3\times3\) grid, together with an unsupported alternative.

\paragraph{Leaf 23: Reasoned Counting of Small Moving Targets.}

Leaf 23 follows the path \emph{L2 Reasoning} \(\rightarrow\) \emph{G7 Spatiotemporal Consistency Reasoning} \(\rightarrow\) \emph{T16 Spatial Relation Consistency}. It belongs to this task because targets contribute to the count only when their positions satisfy a stated spatial-context relation.

This leaf evaluates constrained counting of small targets rather than unfiltered enumeration. The model must apply the spatial predicate consistently to every candidate before producing the total.

\begin{figure*}[tbp]
\centering
\includegraphics[width=\textwidth]{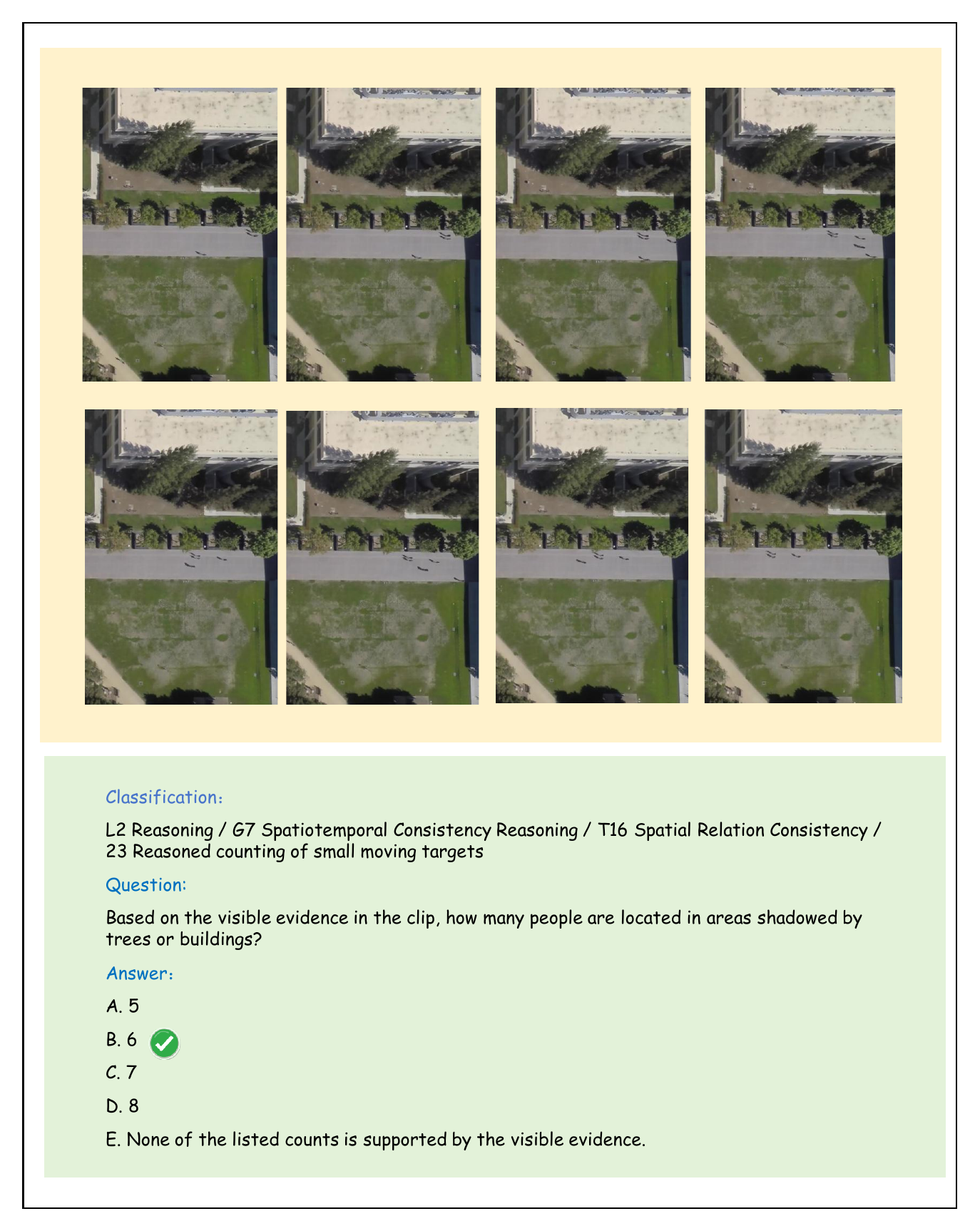}
\caption{Qualitative example for Leaf 23, reasoned counting of small moving targets.}
\label{fig:app-leaf-23}
\end{figure*}

As shown in Fig.~\ref{fig:app-leaf-23}, the example asks how many people are located in areas shadowed by trees or buildings and contrasts counts of five through eight with an unsupported-count alternative.

\paragraph{Leaf 24: Cross-window Constraint Satisfiability.}

Leaf 24 follows the path \emph{L2 Reasoning} \(\rightarrow\) \emph{G7 Spatiotemporal Consistency Reasoning} \(\rightarrow\) \emph{T17 Target Consistency}. It is assigned to target consistency because observations from separate temporal windows must be attributable to one continuous target before the complete constraint set can be satisfied.

This leaf tests whether identity and trajectory evidence remain jointly consistent across multiple temporal windows. Partial satisfaction by different targets must not be mistaken for a single valid track.

\begin{figure*}[tbp]
\centering
\includegraphics[width=0.94\textwidth]{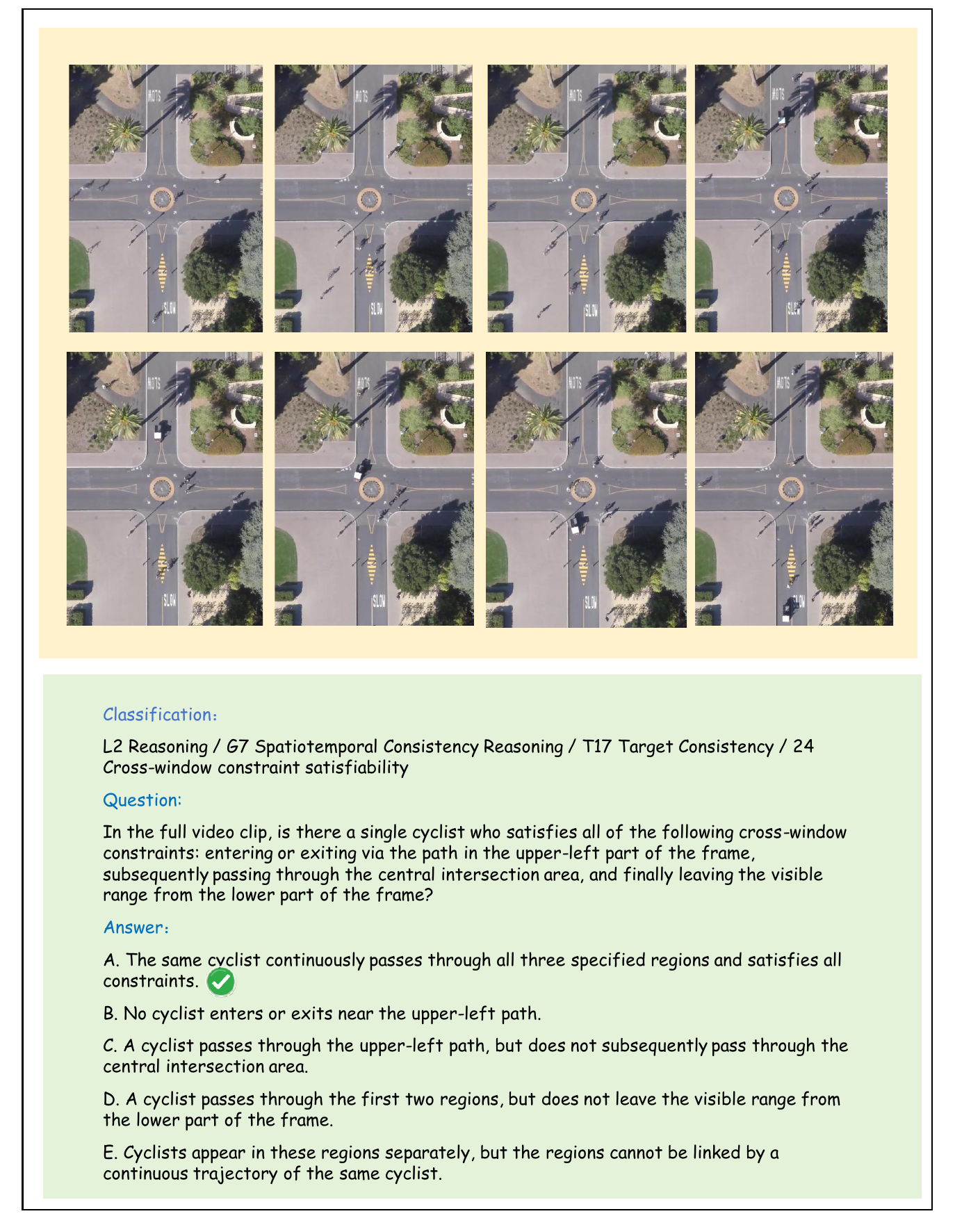}
\caption{Qualitative example for Leaf 24, cross-window constraint satisfiability.}
\label{fig:app-leaf-24}
\end{figure*}

As shown in Fig.~\ref{fig:app-leaf-24}, the example asks whether one cyclist can be linked from the upper-left path through the central intersection to a lower-frame exit and contrasts full continuity with several partial or disconnected cases.

\paragraph{Leaf 25: Target Re-identification after Occlusion.}

Leaf 25 follows the path \emph{L2 Reasoning} \(\rightarrow\) \emph{G7 Spatiotemporal Consistency Reasoning} \(\rightarrow\) \emph{T17 Target Consistency}. It belongs here because an earlier observation and a later candidate must be judged as the same or different target across an interruption in visibility.

This leaf evaluates re-identification using motion-compatible and appearance-compatible evidence around an occlusion or temporal gap. The judgment should reflect plausible scale, displacement, and movement-direction continuity rather than relying on the box colors alone.

\begin{figure*}[tbp]
\centering
\includegraphics[width=0.94\textwidth]{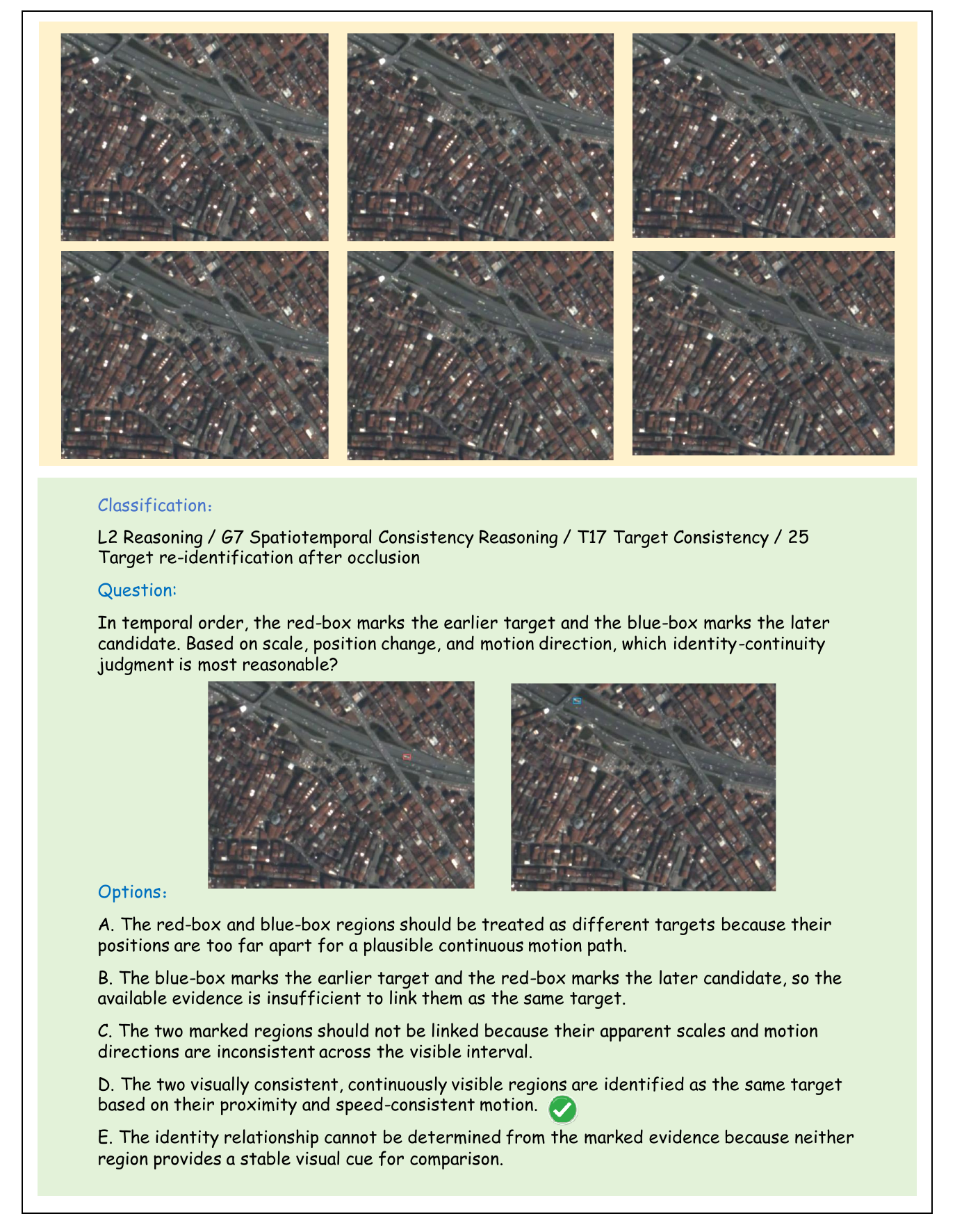}
\caption{Qualitative example for Leaf 25, target re-identification after occlusion.}
\label{fig:app-leaf-25}
\end{figure*}

As shown in Fig.~\ref{fig:app-leaf-25}, the example marks the earlier target in red and the later candidate in blue and asks which identity-continuity judgment is most consistent with their scale, position change, and motion direction.

\paragraph{Leaf 26: Minimal Constraint Relaxation.}

Leaf 26 follows the path \emph{L2 Reasoning} \(\rightarrow\) \emph{G7 Spatiotemporal Consistency Reasoning} \(\rightarrow\) \emph{T17 Target Consistency}. It is a target-consistency leaf because the model must preserve a single-target interpretation while determining which one of several cross-time constraints prevents the complete constraint set from being satisfied.

This leaf tests the minimal relaxation of an unsatisfied constraint set. The preferred answer changes only the constraint necessary to obtain a visually supported trajectory while retaining as much of the original specification as possible.

\begin{figure*}[tbp]
\centering
\includegraphics[width=0.94\textwidth]{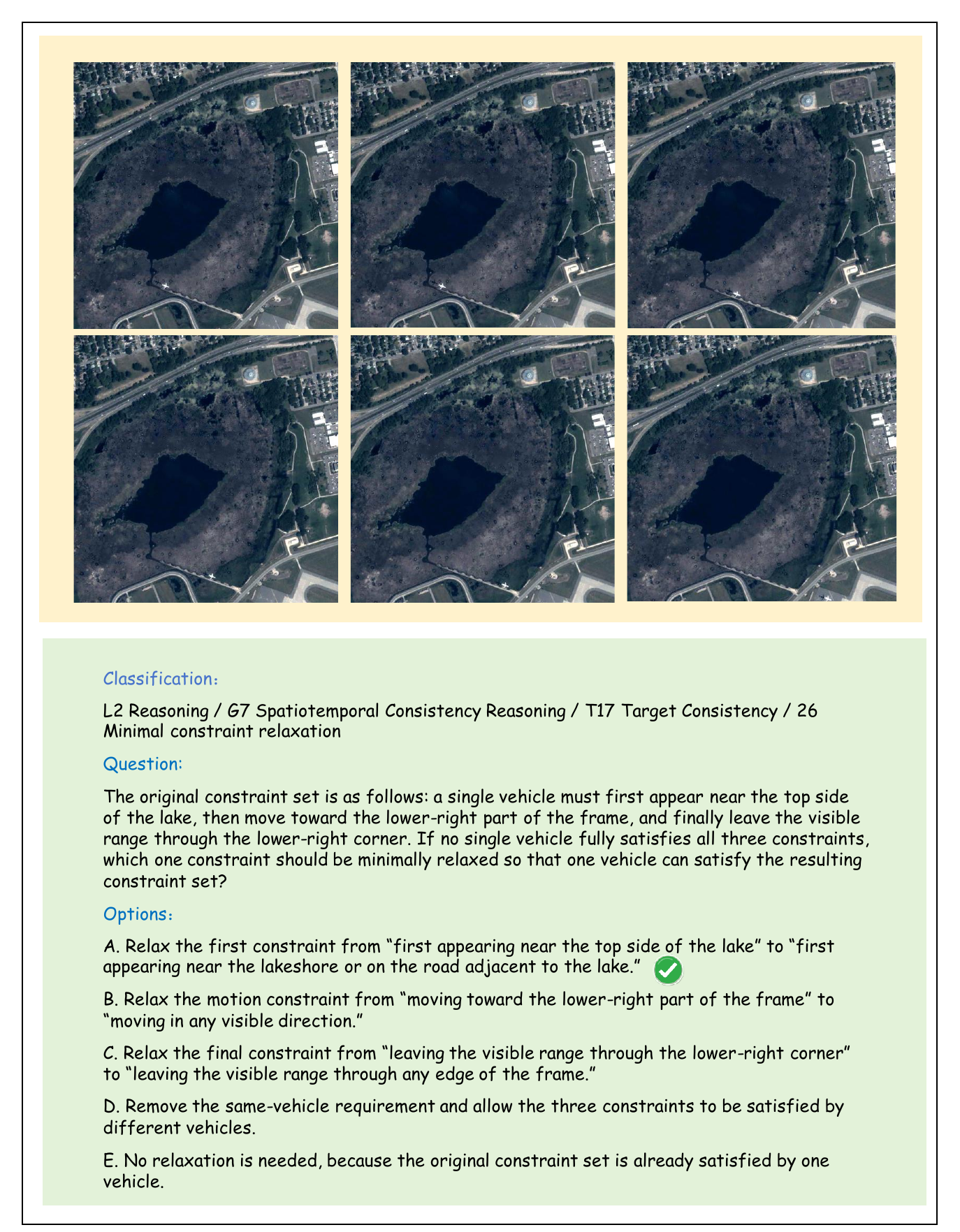}
\caption{Qualitative example for Leaf 26, minimal constraint relaxation.}
\label{fig:app-leaf-26}
\end{figure*}

As shown in Fig.~\ref{fig:app-leaf-26}, the example constrains one vehicle to appear near the top side of a lake, move toward the lower-right, and exit at the lower-right corner, and then asks whether to relax the starting-position, motion-direction, exit-location, or same-vehicle constraint, or make no relaxation.

\normalsize

\clearpage
\onecolumn
\twocolumn
\raggedbottom
\section{Datasheets}
\label{app:datasheets}

\subsection{Motivation}
\label{app:datasheets-motivation}

\begin{enumerate}

    \item \textit{``For what purpose was the dataset created?''}

    \textcolor{BurntOrange}{\textbf{A:}}
\rsvdata{} is designed to evaluate whether vision--language models can understand continuous remote-sensing videos. Existing remote-sensing benchmarks mainly focus on static images or long-interval multi-temporal observations, while general video benchmarks do not reproduce overhead viewpoints, small targets, repetitive backgrounds, and scene-constrained spatial relations. \rsvdata{} fills this evaluation gap by providing a unified five-choice video-question-answering benchmark for perception and reasoning over the included remote-sensing video sources. It supports controlled evaluation on \bench{} and method development on RSVideo-Instruct, with a taxonomy covering L1 perception and L2 reasoning over seven capability groups, 17 tasks, and 26 operational leaf capabilities.

    \item \textit{``Who created the dataset (\textit{e.g.}, which team, research group) and on behalf of which entity?''}

    \textcolor{BurntOrange}{\textbf{A:}} The dataset is created and maintained by the following authors:
    \begin{itemize}
      \item Hongjie Zhou, Shiqin Wang, Haoyang Chen, Haonan Guo, Di Wang, Juhua Liu, Fu Lin, and Yong Luo
    \end{itemize}

    \item \textit{``Who funded the creation of the dataset?''}

    \textcolor{BurntOrange}{\textbf{A:}}
The funding source is not specified in the current manuscript.
\end{enumerate}

\subsection{Composition}
\label{app:datasheets-composition}
Most of the questions in this section are intended to provide dataset consumers with the information they need to make informed decisions about using the dataset for their chosen tasks. Some of the questions are designed to elicit information about compliance with the EU's General Data Protection Regulation (GDPR) or comparable regulations in other jurisdictions. Questions that apply only to datasets that relate to people are grouped together at the end of the section. We recommend taking a broad interpretation of whether a dataset relates to people. For example, any dataset containing text that was written by people relates to people.

\begin{enumerate}

    \item \textit{``What do the instances that comprise our datasets represent (\textit{e.g.}, documents, photos, people, countries)?''}

    \textcolor{BurntOrange}{\textbf{A:}} Each instance represents a question-specific remote-sensing video evidence state. A released record contains a video/evidence-clip identifier, one question, five answer options, the gold option, split membership, source provenance, and taxonomy labels. When needed, an instance also includes a public visual mark or temporal evidence annotation.

    \item \textit{``How many instances are there in total (of each type, if appropriate)?''}

    \textcolor{BurntOrange}{\textbf{A:}} \rsvdata{} contains 10,773 five-choice question--answer instances associated with 4,629 audited evidence clips from eight public remote-sensing video sources. The dataset covers 17.02 hours of continuous video and 1,473,150 decoded frames. It includes 5,651 training items, 2,391 validation items, and 2,731 fixed test items.

    \item \textit{``Does the dataset contain all possible instances or is it a sample (not necessarily random) of instances from a larger set?''}

    \textcolor{BurntOrange}{\textbf{A:}} \rsvdata{} is a curated sample from eight public sources. It does not enumerate every remote-sensing platform, geographic region, weather condition, target class, scene type, or real-world event that may occur in remote-sensing videos.

    \item \textit{``Is there a label or target associated with each instance?''}

    \textcolor{BurntOrange}{\textbf{A:}} Yes. Each instance has one canonical gold answer among five options, together with capability-level, group, task, and leaf labels. Records also include source and split identifiers; when applicable, they include temporal windows, spatial regions, or rendered marks that identify the observable evidence used during annotation.

    \item \textit{``Is any information missing from individual instances?''}

    \textcolor{BurntOrange}{\textbf{A:}} No information required by the official five-choice evaluation interface is intentionally missing. Construction traces, adjudication notes, and train-only evidence supervision are not exposed to validation or test-time models by design.

    \item \textit{``Are relationships between individual instances made explicit (\textit{e.g.}, users' movie ratings, social network links)?''}

    \textcolor{BurntOrange}{\textbf{A:}} Yes. Stable source, video, segment, split, task, and leaf identifiers make shared provenance and taxonomy relationships explicit. The dataset does not include social-network or user-level relationship data.

    \item \textit{``Are there recommended data splits (\textit{e.g.}, training, development/validation, testing)?''}

    \textcolor{BurntOrange}{\textbf{A:}} Yes. RSVideo-Instruct contains 5,651 training items and 2,391 validation items. \bench{} contains 2,731 fixed test items and is reserved for final evaluation.

    \item \textit{``Is the dataset self-contained, or does it link to or otherwise rely on external resources (\textit{e.g.}, websites, tweets, other datasets)?''}

    \textcolor{BurntOrange}{\textbf{A:}} The release is license-aware and hybrid. It directly releases derived annotations, questions, answer options, taxonomy records, split identifiers, marks, evaluation code, and provenance metadata. Visual clips are redistributed only when permitted by the original source terms; otherwise the release provides source identifiers, temporal boundaries, checksums, and retrieval or reconstruction instructions.

    \item \textit{``Does the dataset contain data that might be considered confidential (\textit{e.g.}, data that is protected by legal privilege or by doctor--patient confidentiality, data that includes the content of individuals' non-public communications)?''}

    \textcolor{BurntOrange}{\textbf{A:}} To the best of our knowledge, \rsvdata{} does not contain confidential records, privileged communications, medical records, financial records, or legal records. The visual evidence is derived from publicly available remote-sensing video sources.

    \item \textit{``Does the dataset contain data that, if viewed directly, might be offensive, insulting, threatening, or might otherwise cause anxiety?''}

    \textcolor{BurntOrange}{\textbf{A:}} No such content is intentionally included. Some public aerial or satellite videos may show traffic, pedestrians, vehicles, emergency-related evidence, or disaster-related scene changes at remote-sensing scale. The dataset does not introduce identity labels, biometric attributes, or person-specific private information.

\end{enumerate}

\subsection{Collection Process}
\label{app:datasheets-collection}
In addition to the goals outlined in the previous section, the questions in this section are designed to elicit information that may help researchers and practitioners create alternative datasets with similar characteristics. Again, questions that apply only to datasets that relate to people are grouped together at the end of the section.

\begin{enumerate}

    \item \textit{``How was the data associated with each instance acquired?''}

    \textcolor{BurntOrange}{\textbf{A:}}
The visual inputs come from eight existing public UAV, aerial, overhead, and satellite video resources. We construct the questions, options, gold answers, public marks, taxonomy assignments, and audit records for \rsvdata{}, while preserving source attribution through source and segment identifiers.

    \item \textit{``What mechanisms or procedures were used to collect the data (\textit{e.g.}, hardware apparatuses or sensors, manual human curation, software programs, software APIs)?''}

    \textcolor{BurntOrange}{\textbf{A:}}
\rsvdata{} is constructed through source registration and decoding, clip screening, task and leaf binding, spatiotemporal evidence construction, question and answer drafting, candidate construction, joint evidence--label correction, independent review, expert adjudication, split isolation, and release audit. Three senior domain experts perform initial annotation, independent review, and adjudication. Auxiliary large-model checks serve only to flag possible wording or consistency issues; human experts determine the final labels and retain only items whose answer can be verified from the released visual evidence.

    \item \textit{``If the dataset is a sample from a larger set, what was the sampling strategy (\textit{e.g.}, deterministic, probabilistic with specific sampling probabilities)?''}

    \textcolor{BurntOrange}{\textbf{A:}} We use task-driven curation rather than probabilistic sampling. Candidate segments are retained when they contain sufficient visual evidence for one of the 17 tasks and 26 operational leaves, have usable temporal continuity, and support a unique answer under the fixed five-choice interface. Ambiguous, corrupted, duplicate, temporally overlapping, or private-metadata-dependent candidates are rejected.
\end{enumerate}

\subsection{Preprocessing, Cleaning, and Labeling}
\label{app:datasheets-preprocessing}
The questions in this section are intended to provide dataset consumers with the information they need to determine whether the raw data has been processed in ways that are compatible with their chosen tasks. For example, text that has been converted into a ``bag-of-words'' is not suitable for tasks involving word order.

\begin{enumerate}

    \item \textit{``Was any preprocessing/cleaning/labeling of the data done (\textit{e.g.}, discretization or bucketing, tokenization, part-of-speech tagging, SIFT feature extraction, removal of instances, processing of missing values)?''}

    \textcolor{BurntOrange}{\textbf{A:}}
Yes. Public videos are decoded, screened, segmented into evidence clips, and linked to stable source and segment identifiers. Questions and answer options are normalized to a fixed five-choice interface. Optional visual marks are rendered when a target or region cannot be communicated reliably through text alone. Records are checked for decodability, temporal continuity, answerability, option uniqueness, evidence sufficiency, mark consistency, split isolation, task labels, and leaf labels. Items that fail these checks are corrected when the released frames uniquely support the correction; otherwise they are removed.

    \item \textit{``Was the `raw' data saved in addition to the preprocessed/cleaned/labeled data (\textit{e.g.}, to support unanticipated future uses)?''}

    \textcolor{BurntOrange}{\textbf{A:}} The underlying source videos remain associated with their original public releases and source terms. \rsvdata{} distinguishes third-party visual content from derived benchmark records. Redistributable clips are packaged directly when permitted; restricted clips are represented by provenance metadata, temporal boundaries, checksums, and reconstruction instructions. The newly constructed questions, options, marks, taxonomy assignments, splits, and audit records are maintained as derived dataset annotations.

    \item \textit{``Is the software that was used to preprocess/clean/label the data available?''}

    \textcolor{BurntOrange}{\textbf{A:}} The release package will contain the preprocessing, manifest, and evaluation resources required to reproduce the benchmark interface.
\end{enumerate}

\subsection{Uses}
\label{app:datasheets-uses}
The questions in this section encourage dataset creators to distinguish supported and unsupported uses. Explicit use boundaries help dataset consumers make informed decisions and avoid potential risks or harms.

\begin{enumerate}

    \item \textit{``Has the dataset been used for any tasks already?''}

    \textcolor{BurntOrange}{\textbf{A:}}
Yes. In this paper, \rsvdata{} is used to benchmark remote-sensing video understanding in vision--language models, analyze capability-specific failures, and train, validate, and evaluate \method{}, our evidence-aware spatiotemporal focusing method.

    \item \textit{``Is there a repository that links to any or all papers or systems that use the dataset?''}

    \textcolor{BurntOrange}{\textbf{A:}} The release repository will serve as the index for benchmark resources and reported papers or systems that use \rsvdata{}.

    \item \textit{``What (other) tasks could the dataset be used for?''}

    \textcolor{BurntOrange}{\textbf{A:}}
Beyond the evaluations in this paper, \rsvdata{} can support academic benchmarking of remote-sensing video perception and reasoning, controlled error analysis, video-input adaptation, spatiotemporal evidence localization, small-target temporal tracking, action and event reasoning, and method development using the training and validation splits.

    \item \textit{``Is there anything about the composition of the dataset or the way it was collected and preprocessed/cleaned/labeled that might impact future uses?''}

    \textcolor{BurntOrange}{\textbf{A:}} Yes. \rsvdata{} is English-only, uses a fixed five-choice format, inherits coverage limitations from eight public sources, and has naturally imbalanced task, leaf, source, platform, scene, and target distributions. It serves as a controlled benchmark rather than a complete census of remote-sensing video phenomena.

    \item \textit{``Which tasks fall outside the intended use of the dataset?''}

    \textcolor{BurntOrange}{\textbf{A:}} The intended use excludes privacy-invasive monitoring, biometric identification, operational surveillance, targeting, autonomous control, disaster-response certification, and other safety-critical decision-making without appropriate authorization, independent validation, and human oversight.
\end{enumerate}

\subsection{Distribution}
\label{app:datasheets-distribution}
The following answers document the distribution procedure used for internal or third-party access.

\begin{enumerate}

    \item \textit{``Is the dataset distributed to third parties outside of the creating entity (\textit{e.g.}, company, institution, organization)?''}

    \textcolor{BurntOrange}{\textbf{A:}} The dataset is not yet publicly distributed. Public research access will be provided through the official release channel and will remain subject to the licenses and terms of the underlying source datasets.

    \item \textit{``How is the dataset distributed (\textit{e.g.}, tarball on website, API, GitHub)?''}

    \textcolor{BurntOrange}{\textbf{A:}} The release will use a license-aware hybrid distribution. It will include derived annotations, data splits, public marks, evaluation code, provenance records, legally redistributable clips, and retrieval or reconstruction metadata for visual content that cannot be repackaged.

    \item \textit{``When is the dataset distributed?''}

    \textcolor{BurntOrange}{\textbf{A:}} The distribution point will be the paper's official publication date.

    \item \textit{``Is the dataset distributed under a copyright or other intellectual property (IP) license, and/or under applicable terms of use (ToU)?''}

    \textcolor{BurntOrange}{\textbf{A:}} Yes. The release manifest specifies the license for the annotations, taxonomy records, marks, split manifests, and evaluation resources created for \rsvdata{}. Third-party visual content remains subject to the licenses and terms of its original sources.

    \item \textit{``Have any third parties imposed IP-based or other restrictions on the data associated with the instances?''}

    \textcolor{BurntOrange}{\textbf{A:}} Yes. The eight public source datasets retain their respective licenses, access conditions, attribution requirements, and redistribution restrictions. \rsvdata{} preserves source attribution and uses metadata/reconstruction release modes when visual redistribution is not verified.

    \item \textit{``Do any export controls or other regulatory restrictions apply to the dataset or to individual instances?''}

    \textcolor{BurntOrange}{\textbf{A:}} We are not aware of export-control restrictions specific to the derived benchmark annotations. Users are responsible for complying with applicable laws, regulations, source licenses, and institutional policies in their jurisdictions.
\end{enumerate}

\subsection{Maintenance}
\label{app:datasheets-maintenance}
The following answers document dataset maintenance and communicate the maintenance procedure to dataset consumers.

\begin{enumerate}

    \item \textit{``Who supports, hosts, and maintains the dataset?''}

    \textcolor{BurntOrange}{\textbf{A:}} The authors of this work are the designated dataset maintainers and release hosts.

    \item \textit{``How can the owner/curator/manager of the dataset be contacted (\textit{e.g.}, email address)?''}

    \textcolor{BurntOrange}{\textbf{A:}} The dataset maintainers can be contacted at \texttt{d\_wang@whu.edu.cn} and \texttt{luoyong@whu.edu.cn}.

    \item \textit{``Is there an erratum?''}

    \textcolor{BurntOrange}{\textbf{A:}} There is no erratum for the initial submission. The official release and later benchmark versions document known issues and approved corrections, if any.

    \item \textit{``How is the dataset updated (\textit{e.g.}, to correct labeling errors, add new instances, delete instances)?''}

    \textcolor{BurntOrange}{\textbf{A:}} Dataset updates correct labeling errors, revise invalid items, improve release records, or extend the benchmark. Each update carries a version identifier and change record.

    \item \textit{``How are older versions of the dataset supported, hosted, and maintained?''}

    \textcolor{BurntOrange}{\textbf{A:}} Retained version identifiers and change records trace reported results to the benchmark version used; the release manifest records the hosting policy for older packages.

    \item \textit{``If others want to extend/augment/build on/contribute to the dataset, is there a mechanism for them to do so?''}

    \textcolor{BurntOrange}{\textbf{A:}} The official release channel provides a mechanism for reporting errors and proposing corrections or extensions.
\end{enumerate}
\end{document}